\documentclass[a4paper]{cas-dc} 

\usepackage[numbers,compress,longnamesfirst]{natbib}
\usepackage{float}
\usepackage{graphicx} 
\usepackage{tabularx} 
\usepackage{booktabs}
\usepackage{multirow}
\usepackage{xcolor}
\usepackage{subfig}

\usepackage{calc}
\usepackage{caption}
\usepackage{makecell} 
\usepackage{array}
\usepackage[figuresright]{rotating}
\def\tsc#1{\csdef{#1}{\textsc{\lowercase{#1}}\xspace}}

\tsc{WGM}
\tsc{QE}

\newcolumntype{C}{>{\centering\arraybackslash}X}

\begin{document}
	\begin{sloppypar} 
		\let\printorcid\relax
		\let\WriteBookmarks\relax
		\def\floatpagepagefraction{1}
		\def\textpagefraction{.001}

		\shortauthors{F. Li}  
		\shorttitle{GSPan for Arbitrary-Scale Pansharpening}

		\title [mode = title]{GSPan: A Continuous Gaussian Primitive Representation for Arbitrary-Scale Pansharpening}

		\author[1]{Fangyi Li}
		\author[1]{Xiaoyuan Yang}
		\author[1]{Yixiao Li}
		\author[1]{Zongyang Sui}

		\author[2]{Kangqing Shen}
		\cormark[1]                      
		\ead{shenkq@mail.tsinghua.edu.cn}

		\author[3]{Gemine Vivone}

		\cortext[1]{Corresponding author.}

		\affiliation[1]{organization={Beihang University},
			addressline={School of Mathematical Sciences}, 
			city={Beijing},	
			postcode={102206}, 
			country={China}}
		
		\affiliation[2]{organization={Tsinghua University},
			addressline={Department of Automation}, 
			city={Beijing},
			postcode={100084}, 
			country={China}}

        \affiliation[3]{organization={National Research Council - Institute of Methodologies for Environmental Analysis, CNR-IMAA},
            city={Tito},
            postcode={85050}, 
            country={Italy}}

		\begin{abstract}
			Pansharpening aims to generate high-resolution multispectral (HRMS) images by preserving spectral content from low-resolution multispectral (LRMS) observations while incorporating spatial details guided by a panchromatic (PAN) image. Although deep learning has advanced this task, most existing methods formulate pansharpening as fixed-grid HRMS prediction, which limits scale adaptation. To address this limitation, we propose GSPan, a framework that introduces 2D Gaussian Splatting (GS) into pansharpening. Instead of directly predicting HRMS pixels on a predefined grid, GSPan represents band-wise residual details as continuous and learnable 2D Gaussian primitives. To estimate these primitives from complementary PAN and MS observations, we design a Dual-Stream Hierarchical Interaction (DSHI) architecture with a Spatial-Spectral Interactive Attention (SSIA) module for cross-stream feature interaction. The predicted primitives are rendered as a residual detail field and injected into the upsampled MS image. This continuous representation allows GSPan to render fused images on arbitrary target sampling grids without scale-specific retraining. It further enables a Scale-Decoupled Asymmetric Inference (SDAI) strategy, which estimates primitives at a reduced resolution and renders the fused image at the target resolution for efficient large-scene pansharpening. Experiments on QuickBird, GaoFen-2, WorldView-3, and the large-scene WorldView-3-4K dataset show that GSPan delivers state-of-the-art fusion performance across both cropped-patch and large-scene evaluations. Moreover, SDAI markedly accelerates large-scene inference by decoupling primitive estimation from target-grid rendering, achieving a favorable trade-off between computational efficiency and fusion quality. These results demonstrate the potential of continuous Gaussian residual representations as a flexible and scale-decoupled alternative to fixed-grid HRMS prediction, enabling more efficient large-scene pansharpening.

		\end{abstract}

		
		\maketitle

		\section{Introduction}
		
        Pansharpening is a fundamental problem in remote sensing image fusion, aiming to overcome the intrinsic trade-off between spatial and spectral resolutions in satellite imaging. Panchromatic (PAN) sensors provide high-spatial-resolution observations with limited spectral information, whereas multispectral (MS) sensors capture richer spectral responses at a lower spatial resolution. Accordingly, pansharpening seeks to generate high-resolution multispectral (HRMS) images by preserving the spectral content of low-resolution multispectral (LRMS) observations while incorporating fine spatial details guided by the PAN image \cite{vivone2020new,li2022deep,ZHANG2021323}.  
        Historically, this domain has been explored through three major paradigms \cite{Vivone2014Comparison}: component substitution (CS), multiresolution analysis (MRA), and variational optimization (VO)  . 
		These methods rely on physical models or statistical assumptions, generally achieve good spectral or spatial fidelity, and do not require large training datasets. However, their performance degrade in large-scene or complex scenes due to limitations in modeling intricate nonlinear interactions between PAN and MS images.
        	\begin{figure}
			\centering
			\includegraphics[width=\linewidth]{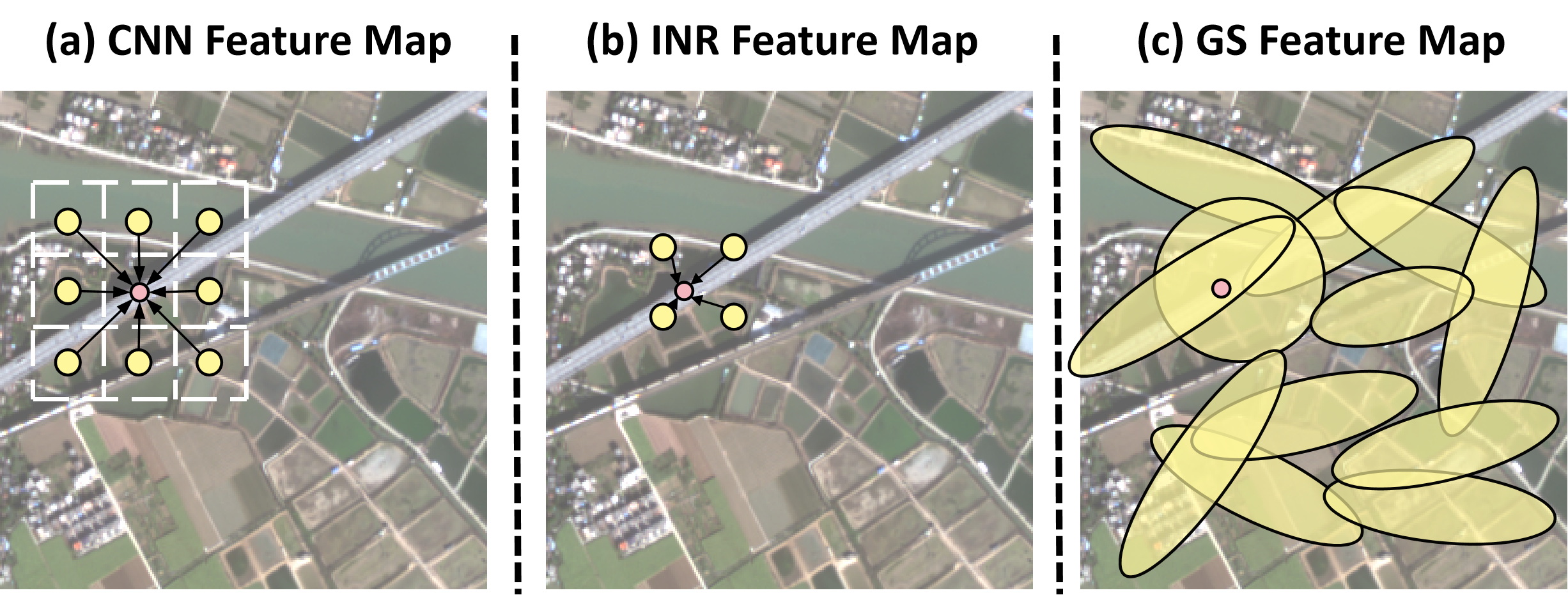}  
			\caption{Conceptual comparison of different representation paradigms. (a) CNN Feature Map: Discrete grid-based representation, illustrated by a CNN feature map as a representative example: features are organized on a regular pixel lattice, making reconstruction dependent on predefined output grids. (b) INR Feature Map: representation is achieved via coordinate-based queries on a latent grid. (c) GS Feature Map (Ours): explicit Gaussian primitives adaptively align with the natural geometric structures.	}
			\label{fig:1}
		\end{figure}
		
		Deep learning (DL) has significantly propelled the field of pansharpening by establishing complex nonlinear mappings between PAN and MS images. A diverse range of architectures has been explored, from early convolutional neural network (CNN)-based frameworks \cite{masi2016pansharpening, Yang2017PanNet, deng2020detail} and generative adversarial network (GAN)-based models \cite{liu2020psgan,pangan_Ma2020} to Transformer-driven architectures \cite{zhou2022panformer, lu2025msan}. More recently, diffusion-based approaches \cite{meng2023pandiff} and state-space models \cite{he2025pan} have further enhanced fusion performance. Despite their success, most existing learning-based methods still formulate pansharpening as a grid-wise reconstruction problem at a predefined output resolution. Although their architectures differ, most existing DL pansharpening methods ultimately predict HRMS images on a predefined discrete grid, often with a fixed scale factor determined by the training or evaluation protocol. Fig.~\ref{fig:1}(a) uses a CNN feature map as a representative example of such grid-based representations, where features are organized on a regular pixel lattice. This formulation does not naturally support arbitrary-scale rendering, since changing the output scale usually requires modifying the upsampling strategy, decoder design, or training setting, offering limited flexibility for real-world scenarios.

		To accommodate the diverse resolution requirements in practical scenarios, pansharpening at arbitrary resolutions has emerged as a novel research focus \cite{9638577,10974400,Ting2025Hyperspectral}. A prominent paradigm in this direction is the implicit neural representation (INR), which models images as continuous functions of coordinates \cite{chen2021learning}. As depicted in Fig.~\ref{fig:1}(b), INRs reconstruct high-resolution textures by querying latent codes at continuous sub-pixel locations, a mechanism that has been successfully applied to various tasks including  arbitrary-scale image super-resolution and image fusion \cite{cao2023ciaosr, lee2022local,liang2024fourier}. However, the coordinate-based query mechanism of INRs typically relies on point-wise multi-layer perceptron (MLP) evaluations, which can lead to increased computational overhead as the target resolution grows. Such characteristics present practical challenges in terms of inference efficiency, particularly when dealing with large scene satellite images prevalent in remote sensing applications.
		
		As a more efficient and geometrically aware alternative, gaussian splatting (GS) \cite{kerbl20233d, chen2024survey, chen2025generalized} has shown strong potential. GS represents details through a collection of learnable 2D Gaussian primitives, each characterized by explicit attributes including location, scaling, orientation, residual coefficient, and spectral coefficient vector. As illustrated in Fig. 1(c), these overlapping primitives form a continuous geometric field, allowing for precise querying at any sub-pixel coordinate. By replacing conventional point-wise interpolation with explicit geometric representation, Gaussian primitives can better preserve structural continuity by explicitly parameterizing spatial extent and orientation. Despite its important impact on 3D reconstruction and super-resolution, the integration of GS into pansharpening paradigm remains an open and promising frontier.
		
		Building upon the explicit geometric modeling advantages of GS, we propose GSPan, the first framework that adapts Gaussian Splatting to pansharpening task. The core of GSPan lies in representing residual detail field as a continuous field of learnable 2D Gaussian primitives. Specifically, we design a Dual-Stream Hierarchical Interaction (DSHI) architecture to generate refined Gaussian primitives: the panchromatic stream provides high-frequency structural guidance, while the multispectral stream supplies  the spectral context. These two complementary streams are integrated via a Spatial-Spectral Interactive Attention (SSIA) module, facilitating mutual feature interaction to estimate precise Gaussian attributes. By adding the rendered Gaussian residuals to the upsampled MS image, GSPan reconstructs sharp spatial details while preserving  spectral fidelity.
 
        The continuous nature of the proposed Gaussian representation further enables arbitrary-scale rendering. Once the Gaussian primitive attributes are estimated from a given PAN-MS input pair, the residual detail field can be rendered on different target sampling grids by adjusting the sampling density in the continuous coordinate space. In this way, GSPan can produce HRMS outputs at user-specified spatial scales without modifying the network architecture or retraining the model. This arbitrary-scale rendering property naturally leads to a Scale-Decoupled Asymmetric Inference (SDAI) strategy for large-scene processing. In standard inference, Gaussian primitive attributes are estimated from the full-resolution PAN-MS inputs and then rendered on the corresponding full-resolution grid, which better exploits fine PAN-guided structures but incurs a high computational cost. In contrast, SDAI estimates the Gaussian primitive attributes at a reduced resolution and renders the fused image on the desired high-resolution grid using the same continuous Gaussian representation. In this way, SDAI decouples the resolution used for primitive attribute estimation from the final rendering resolution. This inference mode substantially improves efficiency, while sacrificing part of the spatial fidelity because fine structures available only in the full-resolution PAN image are not fully used during primitive estimation. Therefore, standard GSPan is more suitable for quality-oriented full-resolution reconstruction, whereas SDAI provides an efficiency-oriented option for large-scene satellite image fusion.
		
		The main contributions of this work are summarized as follows:
		\begin{enumerate}
			\item   
            
            We propose GSPan, which, to the best of our knowledge, represents the first attempt to introduce Gaussian Splatting into the pansharpening task.
			\item Leveraging the continuous representation of Gaussian primitives, GSPan supports flexible rendering at arbitrary pansharpening scales. This scale-continuous rendering capability further enables the proposed Scale-Decoupled Asymmetric Inference (SDAI) strategy, which performs primitive estimation at a reduced resolution and renders the fused image at the target resolution for efficient large-scene inference.	
			\item We design a Dual-Stream Hierarchical Interaction (DSHI) block along with a {Spatial-Spectral Interactive Attention (SSIA)} module to synergistically couple panchromatic structural guidance with multispectral content, providing precise guidance for primitive attribute estimation.
			
			\item We construct a specialized large-scene dataset WV3-4K  to evaluate fusion quality and inference efficiency in $4096 \times 4096$ scenarios. Extensive experiments on QB, GF2, WV3 and WV3-4K datasets demonstrate that GSPan achieves state-of-the-art performance under standard inference, while SDAI provides an efficient solution for large-scene satellite image fusion.
		\end{enumerate}
		
		\section{Related Work}

		\subsection{Deep-Learning-Based Pansharpening at Fixed Resolutions}
		Over the past decade, deep learning has significantly advanced pansharpening. Early representative methods, primarily based on Convolutional Neural Networks (CNNs) such as APNN \cite{scarpa2018target} and TFNet \cite{tfnet_Liu2020}, established the end-to-end nonlinear mapping paradigm. Subsequently, researchers focused on enhancing the network's ability to model complex spatial textures through feature-level multi-scale modeling and global context reasoning. For instance, LPPN  \cite{JIN2022158} utilizes Laplacian pyramids to decompose spatial details across different frequency bands. MSAN \cite{lu2025msan} and MARNet  \cite{pereira2026multi} employ multi-receptive-field attention mechanisms to aggregate hierarchical features. Model-driven and hybrid learning methods have been explored to integrate model-based priors with deep networks. For instance, CSLP~\cite{CHEN2025103002} incorporates compressed sensing into a neural framework to enhance feature fidelity. To further push the boundaries of spatial-spectral modeling, state-space models like PanMamba~\cite{he2025pan} and $S^3$Mamba~\cite{11478251} leverage Mamba's linear complexity to capture long-range dependencies. Similarly, diffusion-based methods such as PanDiff~\cite{meng2023pandiff} introduce generative denoising processes into pansharpening, while low-rank diffusion models~\cite{RUI2024102325} have also been explored in related hyperspectral pansharpening tasks. Beyond backbone design, several studies focus on data efficiency and scale generalization. ZS-Pan~\cite{CAO2024102001} and Zero-Sharpen~\cite{WANG2024102003} investigate zero-shot or image-specific pansharpening schemes, while continual-learning-guided training~\cite{SHEN202345} and DocsNet~\cite{shen2022docsnet} address robustness across reduced- and full-resolution settings.
		
		Despite their sophisticated feature extraction modules, their output formulation is usually tied to a predefined discrete grid or fixed scale factor. This lacks the flexibility to generate arbitrary-resolution products beyond the specific scale used during training.
		
		\subsection{Arbitrary-Scale Pansharpening via continuous representation}
		To meet the emerging demand for flexible spatial reconstruction in diverse remote sensing applications, the concept of Arbitrary-Scale Pansharpening  has recently gained attention. This paradigm was formally established by He et al. \cite{9638577}. While their study primarily addressed hyperspectral data, the underlying principles are directly applicable to the Arbitrary-Scale Multispectral (ASMS) pansharpening task explored in this work. He et al. presents two fundamental challenges that traditional fixed-scale models are not designed to handle. First, since target resolutions in practical scenarios can involve a wide range of potential integer or non-integer scales, it is practically infeasible to train a specific model for every candidate resolution. Second, the model must guarantee spectral fidelity at any arbitrary spatial resolution, especially when the target scale deviates significantly from the training scale. To tackle these issues, ARHS-CNN \cite{9638577} introduced a two-step relay optimization that decouples feature enhancement from spatial rescaling. By first fusing features at the PAN resolution and subsequently tuning them to target scales via a rescaling subnetwork, this framework enables a versatile paradigm for multi-scale fusion. 

        Recent efforts have moved beyond traditional discrete convolutional frameworks toward continuous representation paradigms. Implicit Neural Representations (INRs) have emerged as a powerful approach for continuous signal modeling, representing images as coordinate-based functions, $f(\mathbf{x})$, where signal values are queried via Multi-Layer Perceptrons (MLPs). This paradigm was pioneered by the Local Implicit Image Function (LIIF) \cite{chen2021learning}, and has since evolved through various architectural enhancements—such as frequency-domain encoding and attention mechanisms—to better capture fine textures \cite{cao2023ciaosr, lee2022local,liang2024fourier}. In the domain of pansharpening, INR-based models provide an effective framework for arbitrary-scale fusion \cite{yang2026g}. 
        Despite their flexibility, the coordinate-based "point-query" mechanism of most INRs typically relies on exhaustive MLP evaluations for every sub-pixel location, which lead to substantial computational overhead when rendering large  satellite images. This characteristic have motivated the quest for a more efficient explicit representation to achieve a superior balance between efficiency and fidelity.
		
		Originally developed for 3D radiance fields \cite{kerbl20233d, chen2024survey}, Gaussian Splatting (GS) has been successfully adapted to 2D image tasks, achieving ultra-fast representation and high-fidelity compression \cite{zhang2024gaussianimage}. The application of GS in arbitrary-scale super-resolution \cite{peng2025pixel,chen2025generalized,hu2025gaussiansr} highlights its advantages over the INR paradigm. Unlike the implicit coordinate-to-pixel mapping of INRs, GS-based frameworks represent image through explicit Gaussian primitives, allowing adaptively stretch and align with complex textures and directional edges. Furthermore, GS replaces the repetitive, point-wise MLP queries of INRs with a single-pass primitive field construction and efficient differentiable rasterization. In this work, we propose GSPan to exploit these explicit geometric modeling for more efficient and precise satellite image fusion.
		
		\begin{figure*}[htbp]
			\centering
			\includegraphics[width=\textwidth]{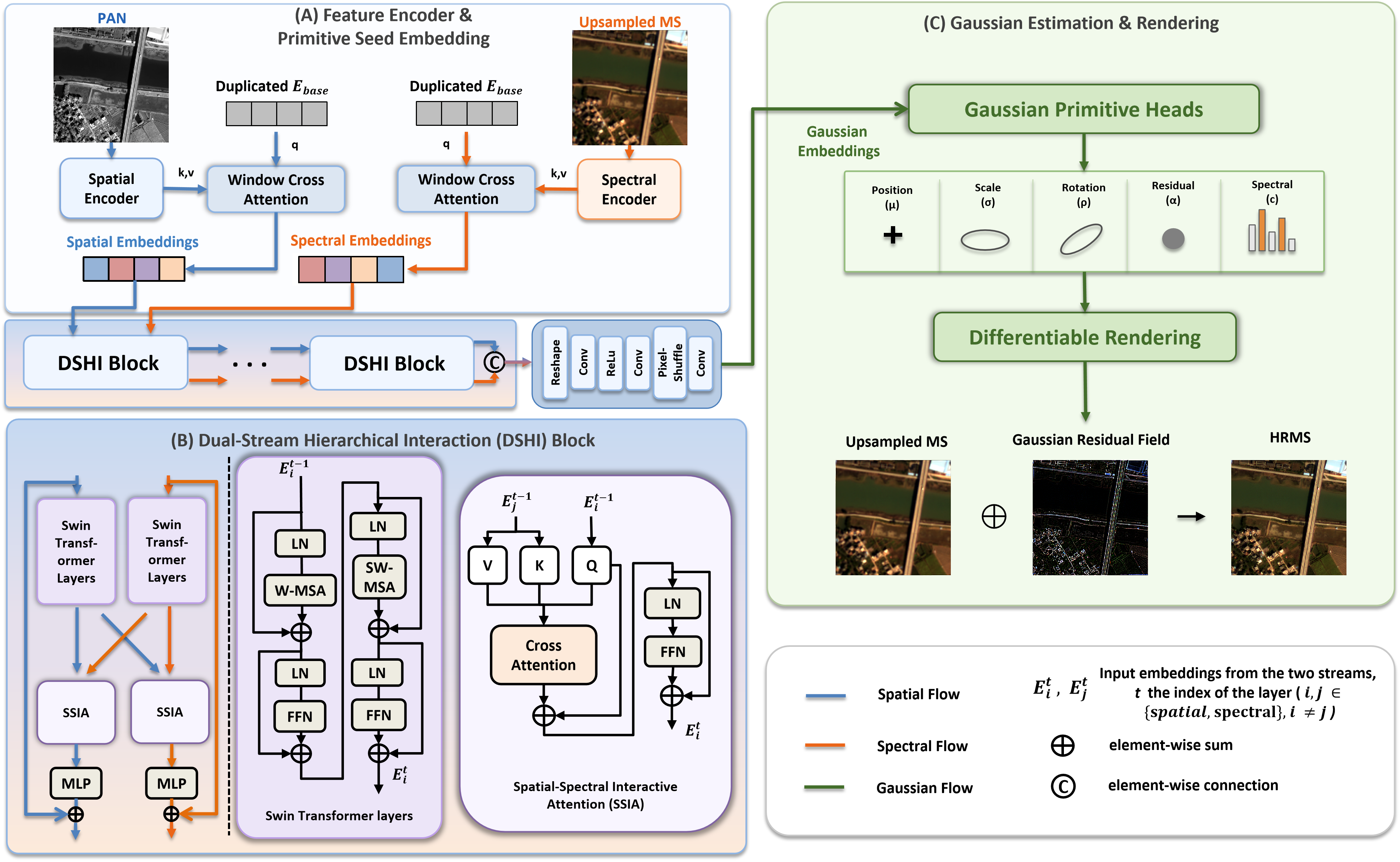}  
			\caption{Overall architecture of the proposed GSPan framework. The pipeline consists of three primary stages: (A) Feature Encoder and Primitive Seed Embedding, which anchors spatial and spectral features into raw embeddings via window cross-attention. (B) Dual-Stream Hierarchical Interaction (DSHI) Block, where Swin Transformer and Spatial-Spectral Interactive Attention (SSIA) modules facilitate intra-stream evolution and inter-stream coupling. W-MSA and SW-MSA are multi-head self attention modules with regular and shifted windowing configurations, respectively. (C) Gaussian Estimation and Rendering, where Gaussian primitives are predicted to synthesize the high-resolution residual field for final image reconstruction.}
			
			\label{fig:overviewv2}
		\end{figure*}
		\section{Methodology} \label{sec:method}
		In this section, we present the proposed GSPan,  a dual-stream spatial-spectral interaction framework for arbitrary-scale pansharpening.
		
		\subsection{Continuous Residual Representation with 2D Gaussian Splatting}
		
        The proposed framework represents the residual detail field of a pansharpened image as a collection of continuous 2D Gaussian primitives. Specifically, we represent the scene using $N$ learnable 2D Gaussians $\mathcal{G} = \{G_1, G_2, \dots, G_N\}$, where the $ i $-th  primitive is defined by five attributes: center position $\boldsymbol{\mu}_i = \{\mu_{x,i}, \mu_{y,i}\}$, scaling factors $\boldsymbol{\sigma}_i = \{\sigma_{x,i}, \sigma_{y,i}\}$, correlation factor $\rho_i \in (-1, 1)$, residual coefficient $\alpha_i \in (-1,1)$, and a spectral coefficient vector $ \mathbf{c}_i \in \mathbb{R}^{C}$ with  $C$ being the number of multispectral bands. For any sub-pixel coordinate $\mathbf{x} = (x, y)$ in the continuous 2D space, the contribution of a single Gaussian $G_i$ is formulated as:
        \begin{equation}
        	G_i(\mathbf{x}) = a_i \cdot \mathbf{c}_i \cdot \exp\left( -\frac{1}{2} (\mathbf{x} - \boldsymbol{\mu}_i)^\top \Sigma_i^{-1} (\mathbf{x} - \boldsymbol{\mu}_i) \right),
        \end{equation}
        where $\Sigma_i$ is the covariance matrix that determines the spatial extent and orientation of the Gaussian, defined as:
        \begin{equation}
        	\Sigma_i = \begin{bmatrix} \sigma_{x,i}^2 & \rho_i \sigma_{x,i} \sigma_{y,i} \\ \rho_i \sigma_{x,i} \sigma_{y,i} & \sigma_{y,i}^2 \end{bmatrix}.
        \end{equation}
        
        Here, $ a_i $ controls the overall intensity of the residual contribution, while $ \mathbf{c}_i = [c_i^{(1)}, c_i^{(2)}, \dots, c_i^{(C)}]^\top $ modulates the contribution across spectral bands independently. This formulation allows each Gaussian primitive to adaptively adjust its spatial extent, orientation, and spectral profile.
        \subsection{Network Architecture}
        \label{subsec:overview}
        The overall architecture of the proposed GSPan framework is illustrated in Fig.~\ref{fig:overviewv2}.
        To learn the primitive-level representation, our network takes the PAN image $\mathbf{I}_{PAN} \in \mathbb{R}^{1 \times H \times W}$ and the upsampled MS image $\mathbf{I}_{\widetilde{MS}} \in \mathbb{R}^{C \times H \times W}$ as source inputs, where $H$ and $W$ denote the height and width of the image respectively.  Initially, respective spatial and spectral feature maps are first extracted via encoders and then projected into a set of learnable spatial and spectral embeddings (Fig.~\ref{fig:overviewv2}(a)). These embeddings are iteratively refined through stacked Dual-Stream Hierarchical Interaction (DSHI) blocks (Fig.~\ref{fig:overviewv2}(b)) to produce $N$ $d$-dimensional Gaussian Embeddings $E_{GS} \in \mathbb{R}^{N \times d}$. Finally, Gaussian Primitive Heads decode these embeddings into explicit Gaussian attributes $\{\boldsymbol{\mu}, \boldsymbol{\sigma}, \rho, \alpha, \mathbf{c}\}$ in a canonical normalized coordinate space $\Omega \in [-1, 1]^2$, enabling them to be rendered into a residual map through an efficient differentiable rasterizer (Fig.~\ref{fig:overviewv2}(c)).
        
        The final fused image $ I_{\text{HRMS}} \in \mathbb{R}^{C \times H \times W} $ is reconstructed using a residual synthesis strategy, where the rendered Gaussian field serves as the detail increment:
        \begin{equation}
        	\mathbf{I}_{HRMS}(x, y, c) = \mathbf{I}_{\widetilde{MS}}(x, y, c) + \sum_{i=1}^{N} G_i(x, y, c),
        \end{equation}
        where $ G_i(x, y, c) $ denotes the $ c $-th component of the vector-valued output from the $ i $-th Gaussian primitive evaluated at spatial coordinate $ (x, y) $.
        
        By encouraging the Gaussians to model the sparse residual map rather than the entire image, this strategy helps preserve spectral consistency and reduces the Gaussian primitive density.

		\subsubsection{Primitive Seed Embedding}
		The Window Cross Attention (WCA) module serves as the critical interface that projected pixel-level spatial and spectral features into learnable spatial and spectral embeddings. A fundamental challenge in primitive-based representation is determining the optimal number of Gaussians $N$. To ensure that the model maintains sufficient representation capacity across variable input sizes, we design $N$ to be adaptively coupled with the input size, specifically setting $N = m\times H \times W$, where $m$ is a predefined density hyperparameter that controls the quantity of Gaussian primitives. This ensures that even for large scenes, a high density of Gaussian primitives is maintained to reconstruct fine-grained textures.
		
		In practical implementation, we partition the feature maps of size $H \times W$ into non-overlapping windows of size $k \times k$. The content within each local window is modeled by a learnable base embedding $E_{base} \in \mathbb{R}^{k^2 \times d}$, where $d$ denotes the embedding dimension. The complete raw embeddings $E$ is then constructed by duplicating $E_{base}$ for $\frac{H}{k} \times \frac{W}{k} $ times to cover the entire spatial domain. Within each window, we perform cross-attention where the base embedding acts as the Query ($Q$), while the extracted image features serve as the Key ($K$) and Value ($V$). The operation is formulated as:
		\begin{equation}
			\text{Attention}(Q, K, V) = \text{SoftMax} \left( \frac{QK^T}{\sqrt{d}} + B \right) V,
		\end{equation}
		where $B$ represents the learnable relative position bias. Through this window-based cross-attention strategy, WCA transforms pixel-level features into a structured set of spatial and spectral embeddings, providing a foundation for the subsequent hierarchical refinement.
		
		\subsubsection{Dual-Stream Hierarchical Interaction Block}
		The refinement phase is centered around stacked Dual-Stream Hierarchical Interaction (DSHI) Blocks. These blocks refine source-specific embeddings while enabling cross-stream information flow between the spatial and spectral embeddings. Structurally, each DSHI block comprises Swin Transformer layers for intra-stream representation learning and an SSIA module for inter-stream coupling.

		\textbf{Intra-stream Context Evolution.} Initially, the spatial and spectral embeddings are processed by independent Swin Transformer layers \cite{liu2021swin} to capture long-range dependencies and global contextual information. By utilizing the shifted window mechanism, the model enables Gaussian seeds to perceive structural continuity beyond local windows.

		\textbf{Spatial-Spectral Interactive Attention (SSIA).} To allow one stream $i$ to adaptively retrieve and integrate complementary features from the other stream $j$, we propose the SSIA module as the key fusion mechanism. As illustrated in Fig.~\ref{fig:overviewv2}(B), SSIA facilitates a deep exchange of information by treating one stream as the query and the other as the reference. For instance, to inject rich spatial details into the spectral stream, the spectral embedding $E_{spec}$ acts as the Query ($Q_i$), while the spatial embedding $E_{spat}$ provides the Key ($K_j$) and Value ($V_j$). The cross-attention operation is mathematically formulated as:
		\begin{equation}
			\begin{aligned} 
				Q_i = E_{i} W_Q, \quad K_j = E_{j} W_K, \quad V_j = E_{j} W_V, \\
				\operatorname{MHCA}(E_i, E_j) = \operatorname{SoftMax}\left(\frac{Q_i K_j^T}{\sqrt{d}}+ B \right ) V_j,
			\end{aligned}
		\end{equation}
		where $E_i, E_j \in \{E_{spat}, E_{spec}\}$ denote the input embeddings from the two streams, and $W_Q, W_K, W_V$ are learnable projection matrices. Following the multi-head cross-attention (MHCA) layer, a residual addition, Layer Normalization (LN), and a Feed-Forward Network (FFN) are applied to further refine the fused representation:
		\begin{equation}
    E_i^{\text{out}} = \operatorname{FFN} \bigl( \operatorname{LN} ( \operatorname{MHCA}(E_i, E_j) + E_i ) \bigr) + E_i^{\text{res}},
\end{equation}	
		where $E_i^{res}$ denotes the residual bypass from the input. In dual-stream architecture, SSIA is performed symmetrically. This symmetric interaction encourages the final Gaussian embeddings to incorporate both geometric and spectral information.
		
		\subsubsection{Gaussian Estimation and Rendering}
		\label{subsec:reconstruction}

		The final stage of GSPan transforms the refined representations into explicit Gaussian primitives. The refined spatial and spectral embeddings are fused by a convolution block. To achieve the target primitive density, the fused Gaussian embeddings $E_{gaussian}$ is subsequently upsampled to $E_{gaussian}^{\uparrow} \in \mathbb{R}^{mk^2 \times d}$ through a sub-pixel convolution layer, ensuring that the number of Gaussian seeds is expanded by a factor of $m$. This high-capacity $E_{gaussian}^{\uparrow}$ then serves as the input for the Gaussian Primitive Heads, which consists of MLP branches to predict the geometric and spectral attributes $\{\mu, \sigma, \rho, \alpha, c\}$ for the $N$ primitives in a canonical normalized coordinate space.
		
		A key advantage of this primitive-based representation is its scale-continuous rendering capability. The predicted Gaussian field exists in a continuous coordinate space, allowing the Gaussian primitive attribute estimation process to remain independent of the final rendering scale. To render the fused result at an arbitrary target resolution scale $H_s \times W_s$, we modulate the sampling density by adjusting the sampling interval within the continuous coordinate space. 

		This mechanism, which will be further illustrate in Sec.\ref{asr}, enables flexible reconstruction without the need for model retraining.
		
		To synthesize the high-resolution residual map, we utilize an efficient differentiable 2D rasterization rendering method \cite{chen2025generalized}. In contrast to the original implementation, we have incorporated engineering enhancements to support parallel batch rendering, allowing for simultaneous processing across large image patches to drastically accelerate inference. By restricting the rendering to a localized range $r$ around each Gaussian center, the computational complexity is effectively maintained at $\mathcal{O}(r^2 H_s W_s N)$.

		\section{Experiment}
		
		\subsection{Datasets}
		\begin{table*}[htbp]
			\centering

			\setlength{\tabcolsep}{3pt}       
			\caption{Illustration of datasets.}
			    \begin{tabular}{c|c|cccc}
    \toprule
    Item  & Sub-item & QB    & GF2   & WV3   & WV3-4K \\
    \midrule
    \multirow{3}[2]{*}{Training sets} & Number of samples  & 17,139 & 19,809 & 9,714 & 9,895 \\
          &  MS shape & 16 $\times$  16 $\times$ 4 & 16 $\times$ 16 $\times$ 4 & 16 $\times$ 16 $\times$ 8 & 16 $\times$ 16 $\times$ 8 \\
          &  PAN shape & 64 $\times$ 64 $\times$ 1 & 64 $\times$ 64 $\times$ 1 & 64 $\times$ 64 $\times$ 1 & 64 $\times$ 64 $\times$ 1 \\
    \midrule
    \multicolumn{1}{c|}{\multirow{3}[2]{*}{Reduced testing sets}} & Number of samples  & 20    & 20    & 20    & 4 \\
          &  MS shape & 64 $\times$ 64 $\times$ 4 & 64 $\times$ 64 $\times$ 4 & 64 $\times$ 64 $\times$ 8 & 256 $\times$ 256 $\times$ 8 \\
          &  PAN  & 256 $\times$ 256 $\times$ 1 & 256 $\times$ 256 $\times$ 1 & 256 $\times$ 256 $\times$ 1 & 1024 $\times$ 1024 $\times$ 1 \\
    \midrule
    \multicolumn{1}{c|}{\multirow{3}[2]{*}{Full testing sets}} & Number of samples  & 20    & 20    & 20    & 4 \\
          &  {MS} shape & 128 $\times$ 128 $\times$ 4 & 128 $\times$ 128 $\times$ 4 & 128 $\times$ 128 $\times$ 8 & 1024 $\times$ 1024 $\times$ 8 \\
          &  PAN shape & 512 $\times$ 512 $\times$ 1 & 512 $\times$ 512 $\times$ 1 & 512 $\times$ 512 $\times$ 1 & 4096 $\times$ 4096 $\times$ 1 \\
    \midrule
    Bit Depth & -     & 11    & 10    & 11    & 11 \\
    \bottomrule
    \end{tabular}%
			\label{tab:dataset}%
		\end{table*}

		To verify the effectiveness of the proposed GSPan, experiments are conducted on two data sources: the public PanCollection \cite{deng2022vivone} and our self-built WV3-4K dataset. PanCollection aggregates satellite data from QuickBird (QB), GaoFen-2 (GF2) and WorldView-3 (WV3). For large-scene inference validation, we further construct the WV3-4K dataset based on 8-band WorldView-3 satellite data. Following Wald’s protocol  \cite{wald1997fusion}, reduced-resolution data are generated from the original full-resolution scenes. Specifically, the details of our datasets are presented in Table~\ref{tab:dataset}. The training set contains 9,895 $64 \times 64$ PAN patch pairs, and four $4096 \times 4096$ full-size PAN scenes serve for test. The details of our datasets are presented in Table~\ref{tab:dataset}. 
		
		Compared with the PanCollection used in this work, where reduced-resolution (RR) and full-resolution (FR) test samples are not strictly paired from the same spatial regions, the proposed WV3-4K test set provides spatially matched RR-FR scenes. The RR samples have a PAN size of $1024 \times 1024$, while the corresponding FR scenes reach $4096 \times 4096$ (PAN). This paired design allows us to evaluate not only the full-resolution fusion quality, but also the ability of GSPan to infer Gaussian primitives at a lower resolution and render them to a higher-resolution output under the proposed Scale-Decoupled Asymmetric Inference strategy.

		\begin{table*}[htbp]
			\centering
			\footnotesize  
			\renewcommand{\arraystretch}{0.9}  
			\caption{Quantitative comparison on the Pancollection datasets. The best results are highlighted in red. The second best results are highlighted in blue. $\uparrow$ indicates that the larger the value, the better the performance, and $\downarrow$ indicates that the smaller the value, the better the performance.}
			\begin{tabular}{cccccccc}
			\toprule
			\multirow{2}{*}{\textbf{Dataset}} & \multirow{2}{*}{\textbf{Method}} & \multicolumn{3}{c}{\textbf{Reference Metrics}} & \multicolumn{3}{c}{\textbf{Non-reference Metrics}} \\
			\cmidrule(lr){3-5} \cmidrule(lr){6-8}
			 
			      &       & SAM $\downarrow$   & ERGAS $\downarrow$ & $Q2^n\uparrow$  & $D_\lambda\downarrow$ & $D_s\downarrow$  & HQNR $\uparrow$ \\
			\midrule
			\multirow{11}[2]{*}{QB} & AWLP  & 8.2935±1.8497 & 8.3823±0.9594 & 0.7979±0.0880 & 0.0392±0.0101 & 0.1211±0.0236 & 0.8446±0.0307 \\
			& TV    & 7.6422±1.5925 & 8.6806±0.9108 & 0.7742±0.0903 & 0.0564±0.0153 & 0.0527±0.0052 & 0.8938±0.0149 \\
			& BDSD  & 8.5243±1.9965 & 7.7426±0.6308 & 0.8236±0.0962 & 0.1907±0.0267 & 0.1162±0.0346 & 0.7158±0.0452 \\
			& PNN   & 5.7045±0.8974 & 5.3349±0.3015 & 0.8930±0.1054 & 0.0457±0.0164 & 0.0501±0.0312 & 0.9070±0.0433 \\
			& FusionNet & 4.8920±0.8092 & 4.1573±0.2492 & 0.9208±0.0994 & 0.0550±0.0165 & 0.0246±0.0269 & 0.9222±0.0396 \\
			& LAGConv & 4.7207±0.7631 & \textcolor[rgb]{ 1,  0,  0}{3.8398±0.3075} & 0.9292±0.0936 & 0.0661±0.0108 & 0.0623±0.0202 & 0.8756±0.0133 \\
			& PSGAN & 4.9338±0.8063 & 4.0294±0.2863 & \textcolor[rgb]{ 0,  0,  1}{0.9292±0.0870} & \textcolor[rgb]{ 0,  0,  1}{0.0289±0.0104} & 0.0478±0.0142 & 0.9247±0.0113 \\
			& MSAN  & \textcolor[rgb]{ 0,  0,  1}{4.6787±0.7731} & 3.8670±0.2814 & 0.9267±0.0954 & 0.0385±0.0120 & 0.0302±0.0166 & \textcolor[rgb]{ 0,  0,  1}{0.9326±0.0256} \\
			& PanMamba & 6.3613±1.1749 & 6.5144±0.5475 & 0.9013±0.0746 & 0.0477±0.0122 & \textcolor[rgb]{ 0,  0,  1}{0.0246±0.0216} & 0.9291±0.0308 \\
			& MARNet & 5.2920±0.7456 & 4.4376±0.3360 & 0.9168±0.1048 & 0.0983±0.0197 & 0.1095±0.0235 & 0.8034±0.0350 \\
			& GSPan    & \textcolor[rgb]{ 1,  0,  0}{4.5566±0.7531} & \textcolor[rgb]{ 0,  0,  1}{3.8593±0.2670} & \textcolor[rgb]{ 1,  0,  0}{0.9317±0.0874} & \textcolor[rgb]{ 1,  0,  0}{0.0282±0.0078} & \textcolor[rgb]{ 1,  0,  0}{0.0201±0.0133} & \textcolor[rgb]{ 1,  0,  0}{0.9524±0.0178} \\
			\midrule
			\multirow{11}[2]{*}{GF2} & AWLP  & 1.9461±0.4823 & 1.6753±0.3342 & 0.8687±0.0288 & 0.0338±0.0153 & 0.1263±0.0293 & 0.8444±0.0377 \\
			& TV    & 1.6687±0.3083 & 1.6269±0.3445 & 0.9076±0.0272 & 0.0509±0.0411 & 0.0846±0.0166 & 0.8688±0.0400 \\
			& BDSD  & 1.8009±0.3143 & 1.7291±0.3863 & 0.8859±0.0311 & 0.0694±0.0213 & 0.1142±0.0210 & 0.8244±0.0256 \\
			& PNN   & 1.2024±0.2047 & 1.2605±0.1884 & 0.9382±0.0214 & 0.0145±0.0055 & 0.0266±0.0115 & 0.9594±0.0157 \\
			& FusionNet & 1.0064±0.1962 & 1.0704±0.2137 & 0.9580±0.0111 & 0.0261±0.0063 & 0.0480±0.0117 & 0.9272±0.0160 \\
			& LAGConv & 0.8926±0.1448 & \textcolor[rgb]{ 0,  0,  1}{0.8155±0.1102} & 0.9726±0.0110 & 0.0177±0.0050 & 0.0164±0.0107 & 0.9662±0.0142 \\
			& PSGAN & \textcolor[rgb]{ 0,  0,  1}{0.8332±0.1608} & 0.7522±0.1384 & \textcolor[rgb]{ 0,  0,  1}{0.9761±0.0109} & \textcolor[rgb]{ 0,  0,  1}{0.0117±0.0035} & 0.0203±0.0062 & \textcolor[rgb]{ 0,  0,  1}{0.9683±0.0086} \\
			& MSAN  & 0.8529±0.1448 & 0.7858±0.1169 & 0.9737±0.0132 & 0.0153±0.0088 & \textcolor[rgb]{ 0,  0,  1}{0.0163±0.0068} & 0.9687±0.0130 \\
			& PanMamba & 1.4601±0.2602 & 1.7170±0.3085 & 0.9337±0.0151 & 0.0225±0.0058 & 0.0213±0.0087 & 0.9567±0.0119 \\
			& MARNet & 0.9185±0.1445 & 0.8996±0.1571 & 0.9702±0.0152 & 0.0174±0.0097 & 0.0165±0.0069 & 0.9664±0.0140 \\
			& GSPan & \textcolor[rgb]{ 1,  0,  0}{0.7027±0.1294} & \textcolor[rgb]{ 1,  0,  0}{0.6470±0.1176} & \textcolor[rgb]{ 1,  0,  0}{0.9820±0.0087} & \textcolor[rgb]{ 1,  0,  0}{0.0111±0.0031} & \textcolor[rgb]{ 1,  0,  0}{0.0162±0.0063} & \textcolor[rgb]{ 1,  0,  0}{0.9729±0.0083} \\
			\midrule
			\multirow{11}[2]{*}{WV3} & AWLP  & 5.2545±1.4044 & 4.8476±1.3298 & 0.8006±0.0959 & 0.0151±0.0065 & 0.0451±0.0252 & 0.9406±0.0306 \\
			& TV    & 5.5975±1.6839 & 4.7675±1.2667 & 0.7916±0.1155 & 0.0234±0.0061 & 0.0393±0.0227 & 0.9384±0.0269 \\
			& BDSD  & 6.0381±1.7072 & 5.0489±1.6048 & 0.7861±0.1193 & 0.1138±0.0174 & 0.0903±0.0189 & 0.8060±0.0171 \\
			& PNN   & 4.2827±0.7839 & 3.1521±0.7638 & 0.8554±0.1021 & 0.0263±0.0108 & 0.0344±0.0183 & 0.9404±0.0275 \\
			& FusionNet & 3.3692±0.6340 & 2.4950±0.5813 & 0.8896±0.0954 & 0.0195±0.0087 & \textcolor[rgb]{ 1,  0,  0}{0.0274±0.0158} & \textcolor[rgb]{ 0,  0,  1}{0.9537±0.0221} \\
			& LAGConv & 3.2724±0.5495 & \textcolor[rgb]{ 0,  0,  1}{2.3797±0.5423} & 0.8901±0.0986 & 0.0205±0.0086 & 0.0323±0.0138 & 0.9479±0.0202 \\
			& PSGAN & \textcolor[rgb]{ 0,  0,  1}{3.2393±0.5830} & 2.3468±0.4727 & 0.8952±0.0950 & 0.0190±0.0086 & \textcolor[rgb]{ 0,  0,  1}{0.0281±0.0098} & 0.9535±0.0174 \\
			& MSAN  & 3.0496±0.5384 & 2.2590±0.4787 & \textcolor[rgb]{ 0,  0,  1}{0.8973±0.0933} & \textcolor[rgb]{ 0,  0,  1}{0.0177±0.0077} & 0.0371±0.0147 & 0.9460±0.0212 \\
			& PanMamba & 5.5342±1.6583 & 4.9042±1.1529 & 0.8500±0.0925 & 0.0312±0.0141 & 0.0322±0.0249 & 0.9379±0.0360 \\
			& MARNet & 3.7290±0.6196 & 2.7701±0.8265 & 0.8657±0.1124 & 0.0277±0.0130 & 0.0479±0.0198 & 0.9260±0.0309 \\
			& GSPan & \textcolor[rgb]{ 1,  0,  0}{2.9104±0.5313} & \textcolor[rgb]{ 1,  0,  0}{2.1701±0.4561} & \textcolor[rgb]{ 1,  0,  0}{0.9016±0.0939} & \textcolor[rgb]{ 1,  0,  0}{0.0140±0.0064} & 0.0289±0.0132 & \textcolor[rgb]{ 1,  0,  0}{0.9575±0.0174} \\
			\bottomrule
		\end{tabular}%
			\label{tab:Pancollection}%
		\end{table*}%
		
		Quantitative evaluation is performed using both reduced-resolution (RR) and full-resolution (FR) metrics. For the RR assessment, we use three reference evaluation metrics including SAM \cite{alparone2004global}, ERGAS \cite{alparone2007comparison}, and $Q2^n$ \cite{garzelli2009hypercomplex}. For the FR experiments in real-world scenes, we use three non-reference evaluation metrics including the HQNR \cite{9779258} index, along with its spectral distortion index $D_\lambda$ and the spatial distortion index $D_s$ to evaluate the fusion quality without a reference image.

		\subsection{Implementation details}
        The GSPan framework is implemented in PyTorch and trained on one NVIDIA Tesla V100 (32GB) GPU. We utilize the Adam optimizer with an initial learning rate of $2 \times 10^{-4}$, which is regulated by a 5-epoch linear warmup followed by a cosine annealing schedule decreasing to $1 \times 10^{-7}$. The model is trained for 300 epochs on WV3 and GF2, 180 epochs on QB with batch size 24 and optimized via the pixel-wise $\mathcal{L}_1$ loss between fused HRMS outputs and the corresponding ground truth. Regarding the architectural configuration, the local processing window is set to $8 \times 8$ with 64 Gaussian seeds. To facilitate deep feature interaction and primitive refinement, the model incorporates 4 DSHI blocks (each containing 4 layers). The density $m$ is configured as 4, which means that each pixel is represented by 4 Gaussian primitives on average.
		
		\subsection{Comparison with Representative Pansharpening Methods}

		We compare GSPan against a diverse set of representative pansharpening methods, which are categorized into two groups: traditional methods and deep learning (DL) methods.
		
		\begin{itemize}
			\item \textbf{Traditional Methods}: This category includes classical benchmarks based on different physical and mathematical paradigms, such as the component substitution (CS) method BDSD \cite{garzelli2007optimal}, the multi-resolution analysis (MRA) method AWLP \cite{otazu2005introduction}, and the variational optimization (VO) based approach TV \cite{palsson2013new}.
			
			\item \textbf{Deep Learning Methods}: To provide a comprehensive evaluation, we incorporate a broad spectrum of representative deep-learning methods. These include representative CNN-based models ({PNN} \cite{masi2016pansharpening} and {FusionNet} \cite{deng2020detail}, {LAGConv} \cite{jin2022lagconv}), the generative adversarial network ({PSGAN} \cite{liu2020psgan}), the attention-driven frameworks ({MSAN} \cite{lu2025msan}, and {MARNet} \cite{pereira2026multi}), and the state-space model (SSM) based approach ({PanMamba} \cite{he2025pan}).
		\end{itemize}

		\subsubsection{Evaluation on reduced-resolution scenes}

        The quantitative results on reduced-resolution scenes are summarized in Table 2. Overall, GSPan achieves the best or highly competitive performance across the three datasets. On the QB dataset, GSPan obtains the best SAM and $Q2^n$ scores, while its ERGAS ranks second, slightly behind LAGConv. On the GF2 and WV3 dataset, GSPan consistently achieves the best results in all three reduced-resolution metrics, including SAM, ERGAS, and $Q2^n$. These results indicate that the proposed continuous Gaussian residual representation is effective under the Wald-protocol evaluation, providing  high spectral fidelity and accurate spatial detail reconstruction.  The consistent improvements across different sensors suggest that GSPan does not depend on dataset-specific favorable bias, but achieves strong generalization on reduced-resolution pansharpening test configurations.

\begin{figure*}[htbp]
			\centering

			\captionsetup[subfloat]{labelformat=empty, skip=0pt}
			\setlength{\tabcolsep}{1pt}

			\newlength{\myw}
			\setlength{\myw}{0.14\textwidth}
			
			\begin{tabular}{ccccccl}
			
				\includegraphics[width=\myw]{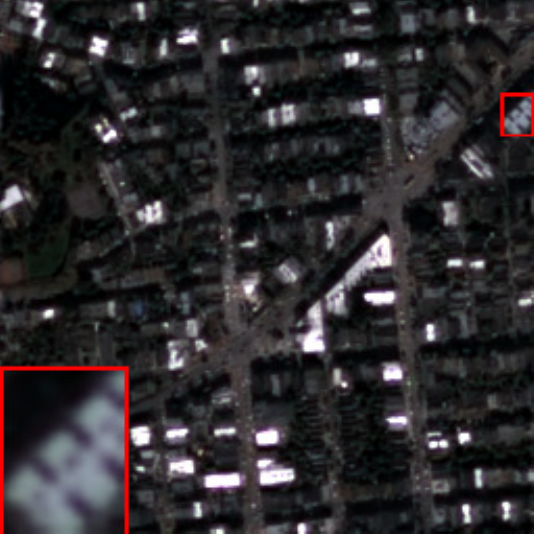} &
				\includegraphics[width=\myw]{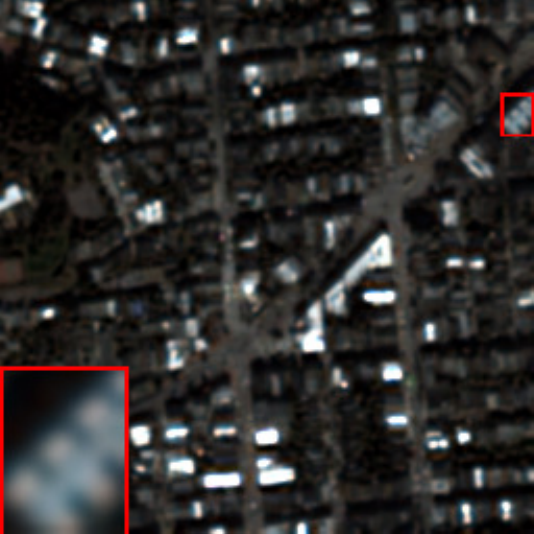} &
				\includegraphics[width=\myw]{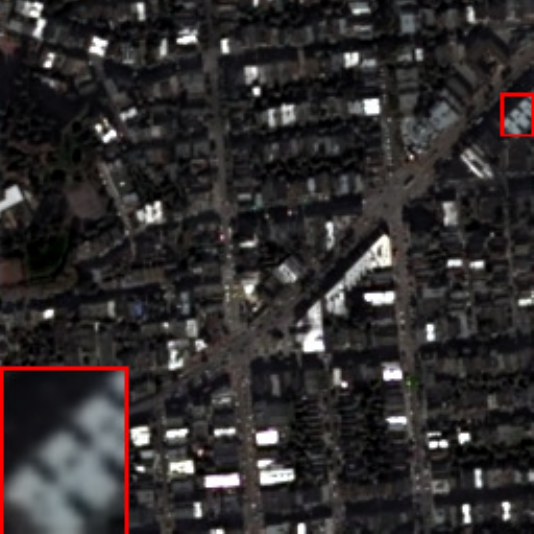} &
				\includegraphics[width=\myw]{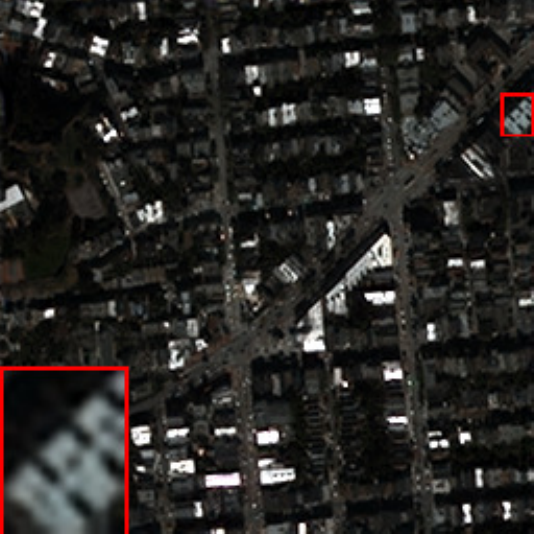} &
				\includegraphics[width=\myw]{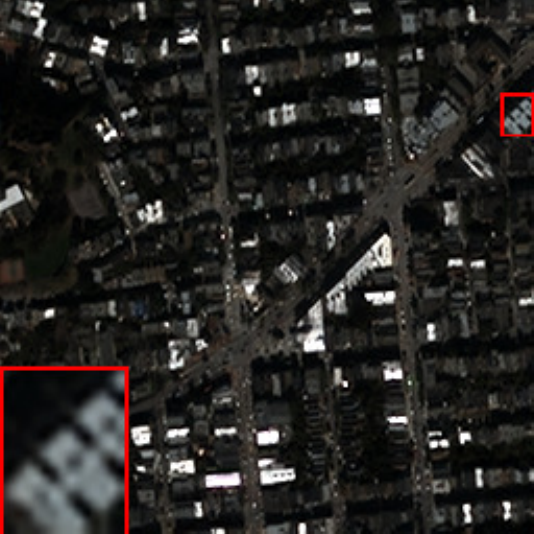} &
				\includegraphics[width=\myw]{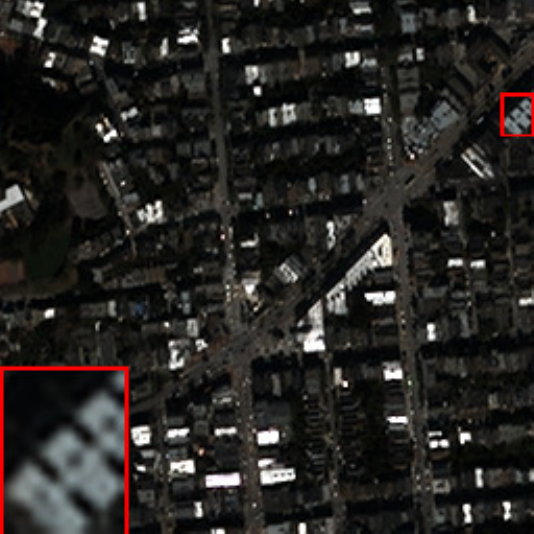} & 
				\multirow{6}{*}[\myw+0.5pt]{\includegraphics[height=\dimexpr 4\myw + 33pt\relax, trim=10 0 0 0, clip]{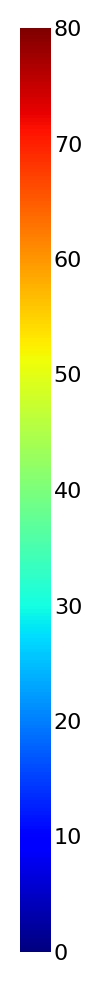}} \\[-1pt]
				
				\includegraphics[width=\myw]{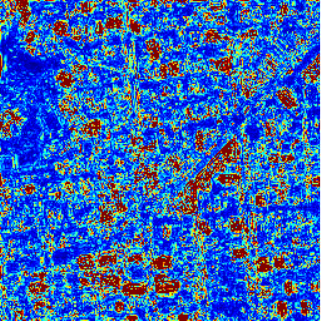} &
				\includegraphics[width=\myw]{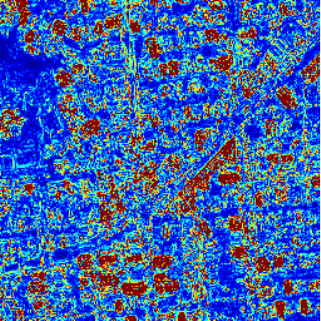} &
				\includegraphics[width=\myw]{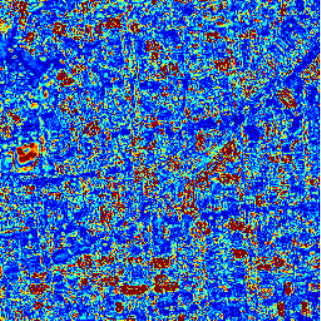} &
				\includegraphics[width=\myw]{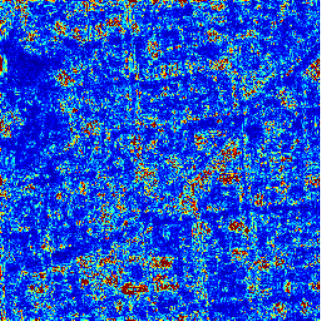} &
				\includegraphics[width=\myw]{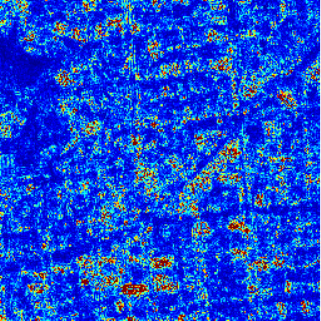} &
				\includegraphics[width=\myw]{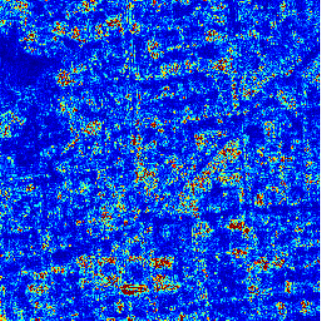} & \\[-2pt]
				
				\small AWLP & \small TV & \small BDSD & \small PNN & \small FusionNet & \small LAGConv & \\[-2pt]
				
				\includegraphics[width=\myw]{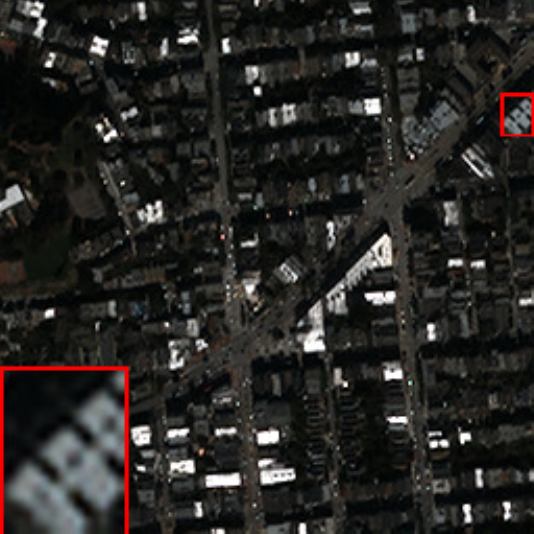} &
				\includegraphics[width=\myw]{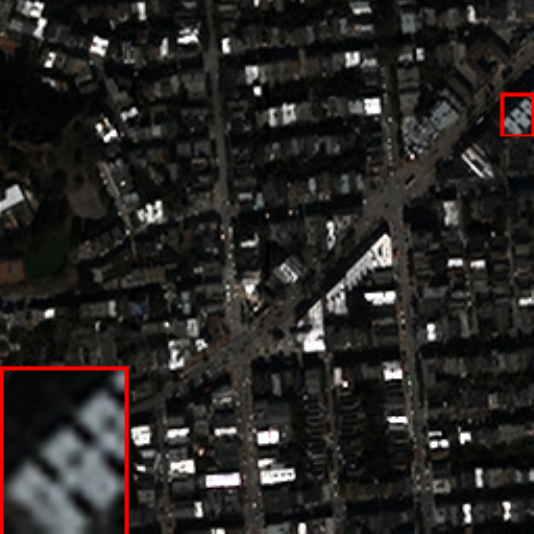} &
				\includegraphics[width=\myw]{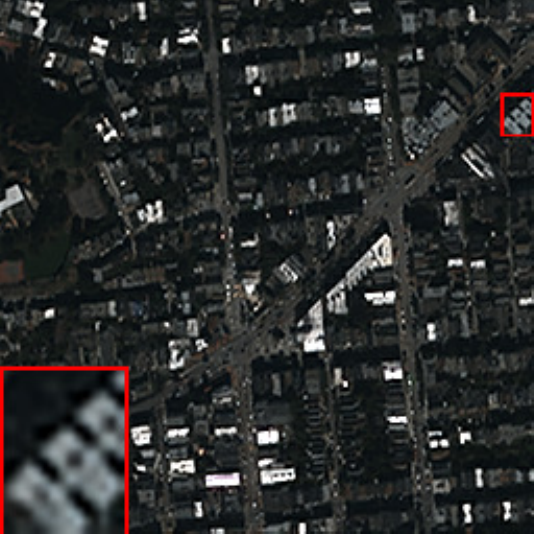} &
				\includegraphics[width=\myw]{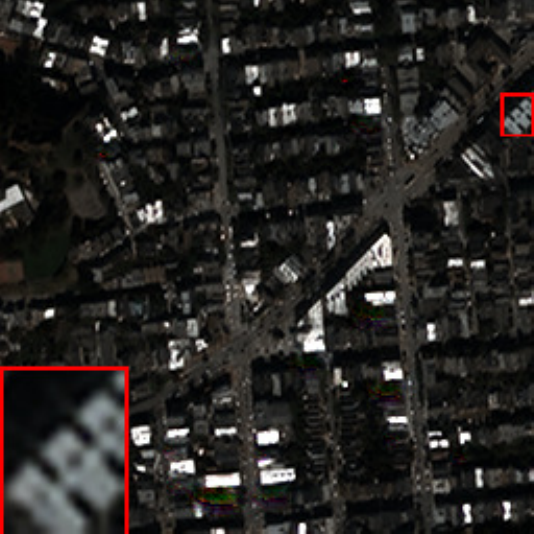} &
				\includegraphics[width=\myw]{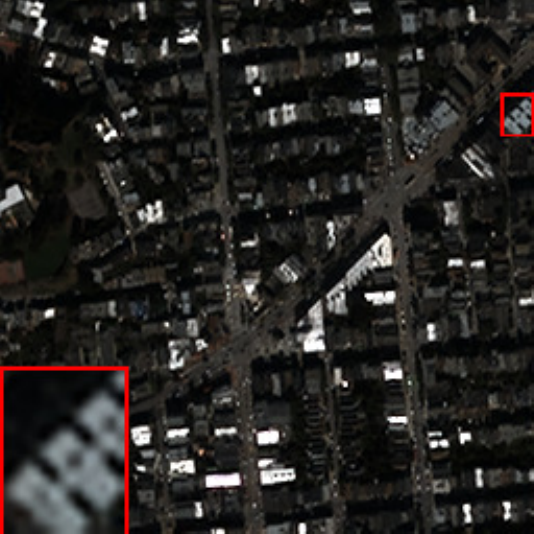} &
				\includegraphics[width=\myw]{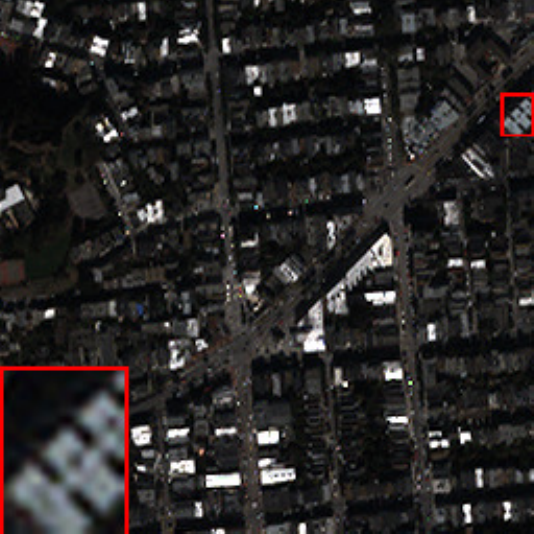} & \\[-1.5pt]
				
				\includegraphics[width=\myw]{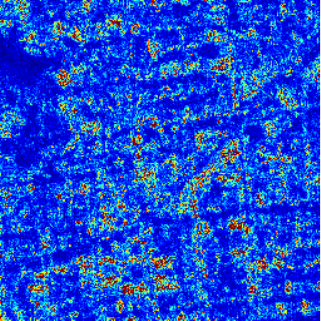} &
				\includegraphics[width=\myw]{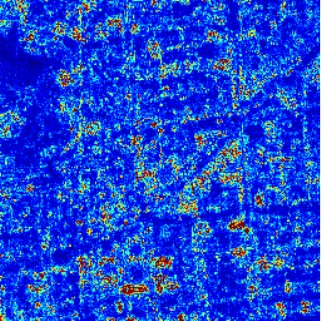} &
				\includegraphics[width=\myw]{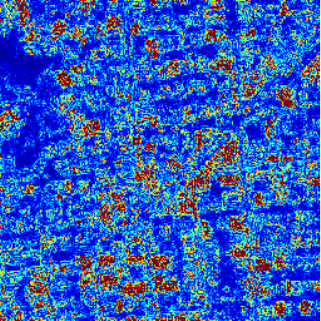} &
				\includegraphics[width=\myw]{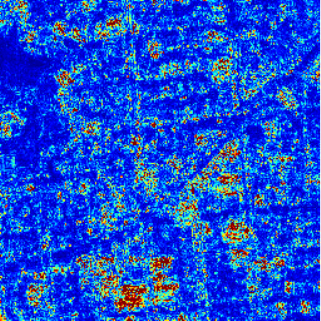} &
				\includegraphics[width=\myw]{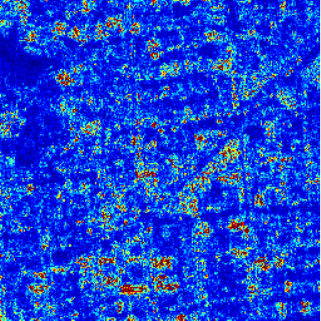} &
				\includegraphics[width=\myw]{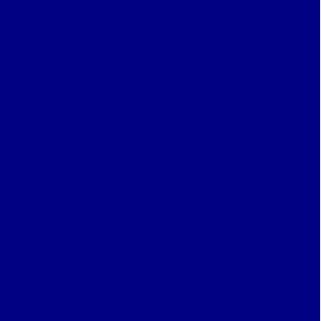} & \\[-2pt]
				
				\small PSGAN & \small MSAN & \small PanMamba & \small MARNet & \small GSPan & \small GT & \\[-6pt]
			\end{tabular}
			
			\caption{Qualitative comparison between the proposed GSPan and other methods on the QB dataset. Rows 1 and 3 show pansharpened fused images, and Rows 2 and 4 show corresponding absolute error maps against the ground truth (GT).}
			\label{fig:rr_qb_comparison}
		\end{figure*}

\begin{figure*}[htbp]
			\centering
			
			\captionsetup[subfloat]{labelformat=empty, skip=0pt}
			\setlength{\tabcolsep}{1pt}

			\newlength{\mywgf}
			\setlength{\mywgf}{0.14\textwidth} 
			
			\begin{tabular}{ccccccl} 

				\includegraphics[width=\mywgf]{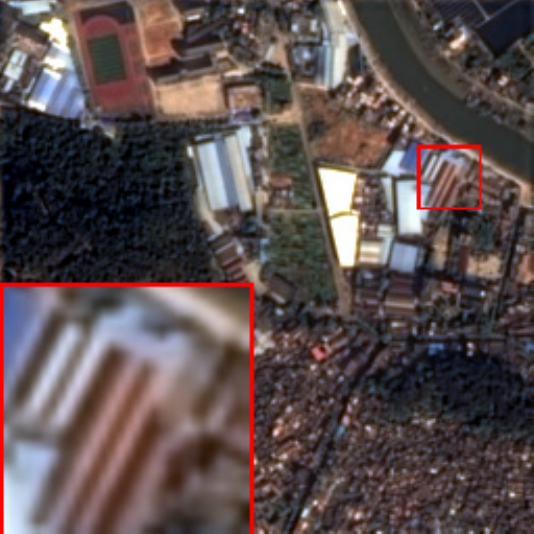} &
				\includegraphics[width=\mywgf]{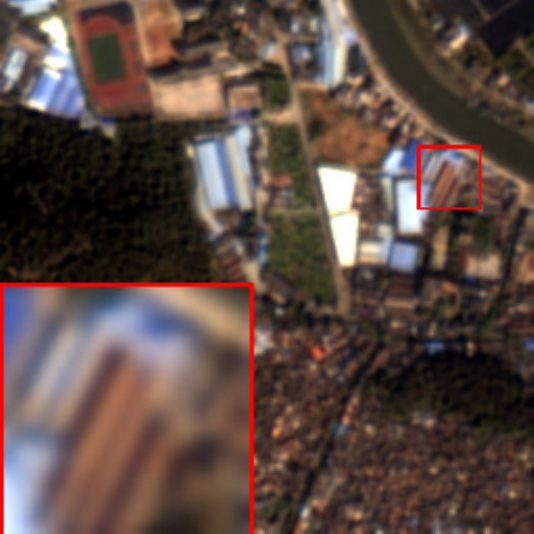} &
				\includegraphics[width=\mywgf]{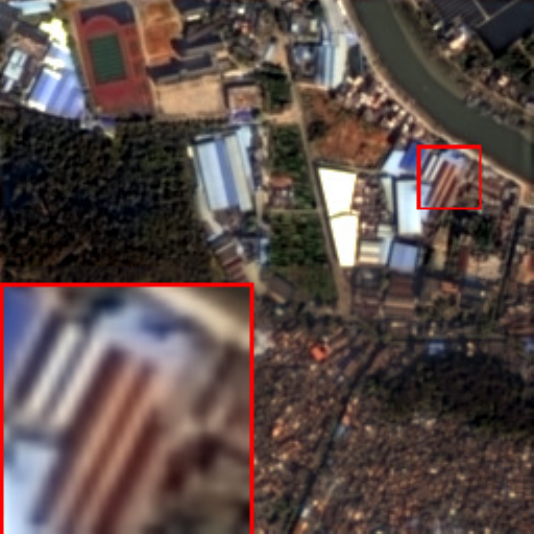} &
				\includegraphics[width=\mywgf]{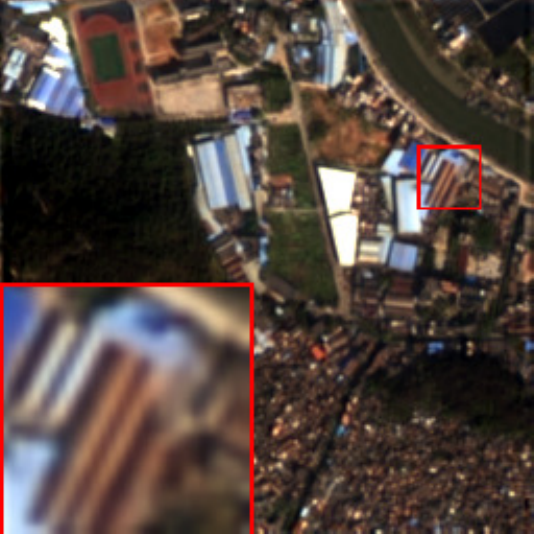} &
				\includegraphics[width=\mywgf]{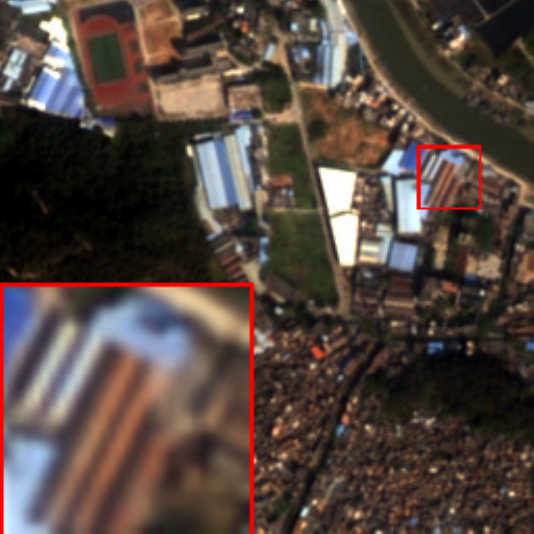} &
				\includegraphics[width=\mywgf]{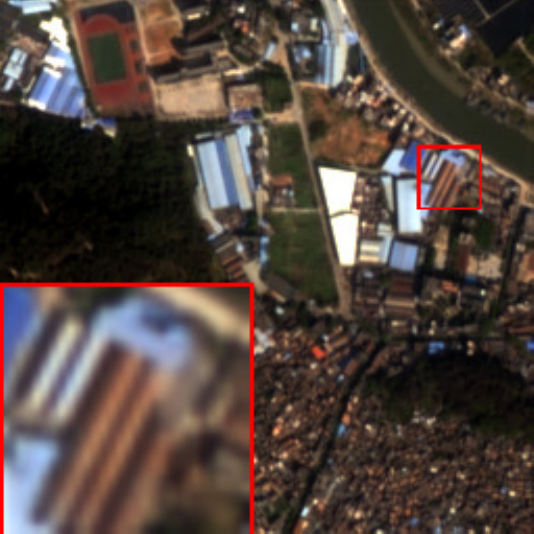} & 
		
				\multirow{6}{*}[\mywgf+0.5pt]{\includegraphics[height=\dimexpr 4\mywgf + 33pt\relax, trim=10 0 0 0, clip]{./images/qb/rr/error_map/global_colorbar.png}} \\[-1pt]
				
				\includegraphics[width=\mywgf]{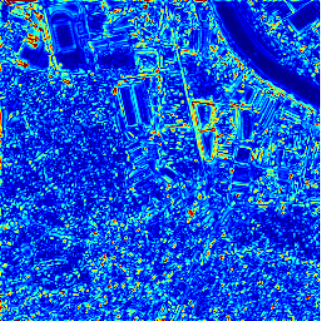} &
				\includegraphics[width=\mywgf]{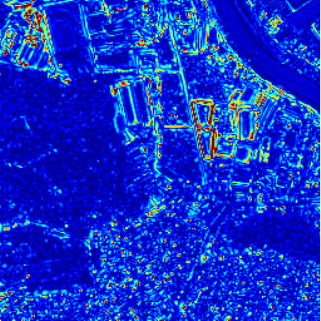} &
				\includegraphics[width=\mywgf]{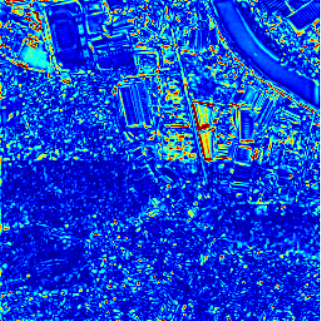} &
				\includegraphics[width=\mywgf]{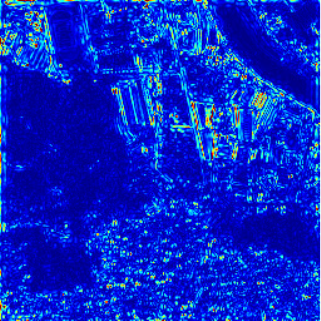} &
				\includegraphics[width=\mywgf]{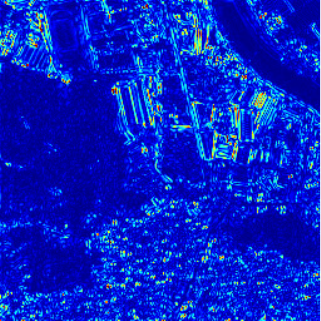} &
				\includegraphics[width=\mywgf]{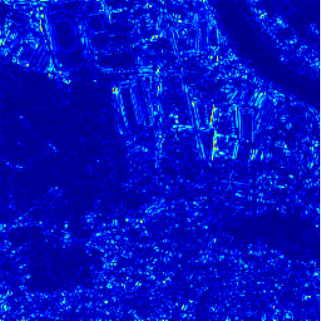} & \\[-2pt]
				
				\small AWLP & \small TV & \small BDSD & \small PNN & \small FusionNet & \small LAGConv & \\[-2pt]

				\includegraphics[width=\mywgf]{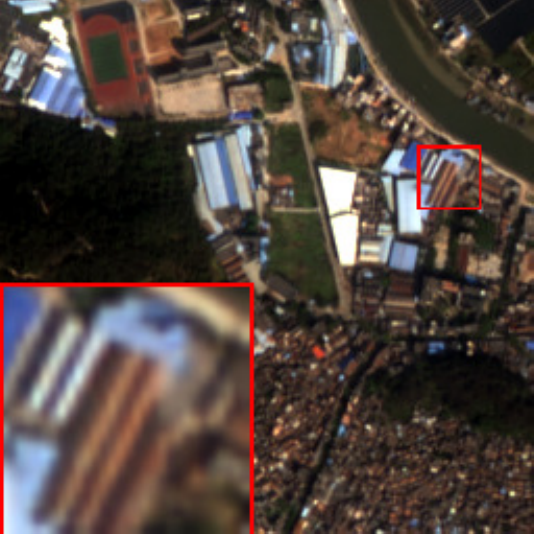} &
				\includegraphics[width=\mywgf]{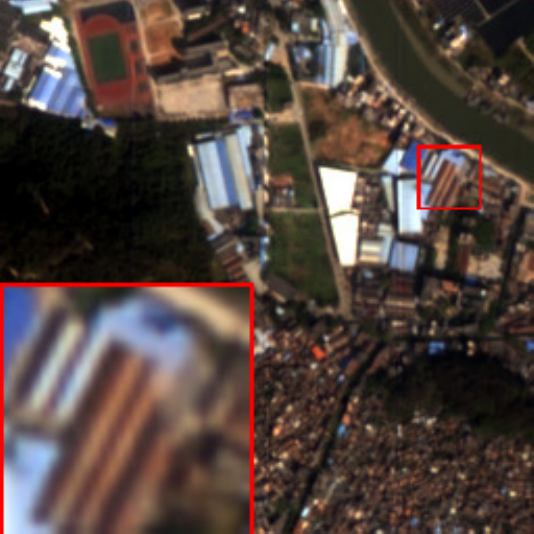} &
				\includegraphics[width=\mywgf]{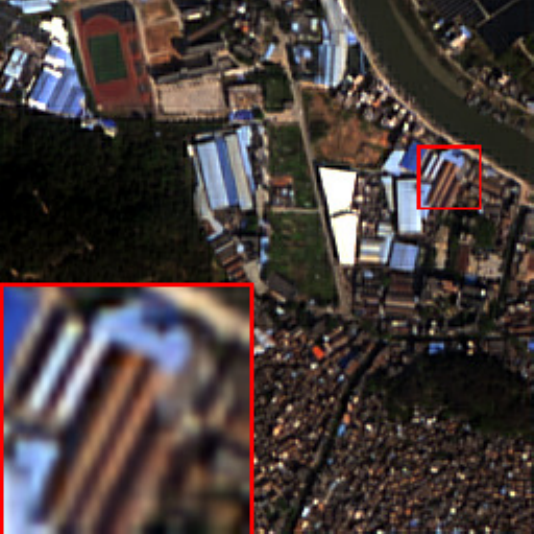} &
				\includegraphics[width=\mywgf]{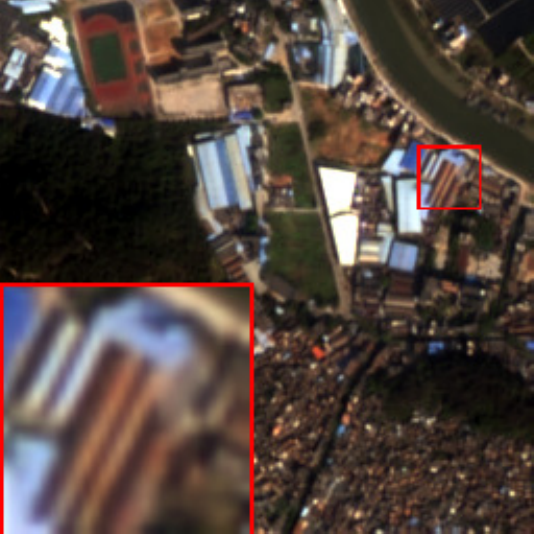} &
				\includegraphics[width=\mywgf]{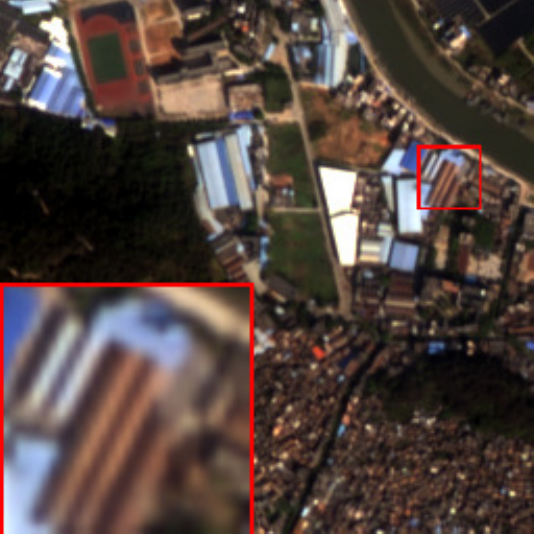} &
				\includegraphics[width=\mywgf]{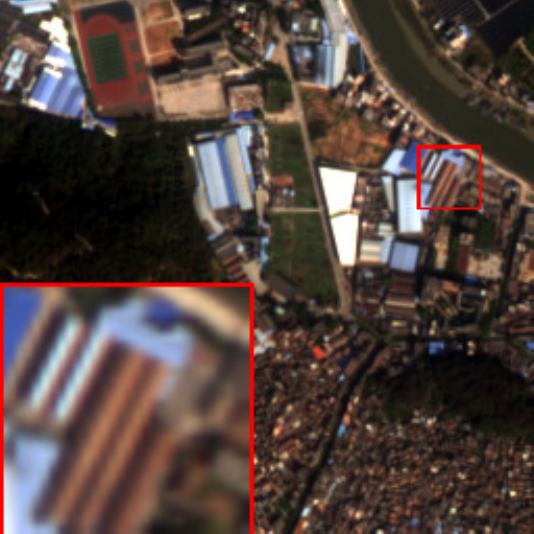} & \\[-1.5pt]
				
				\includegraphics[width=\mywgf]{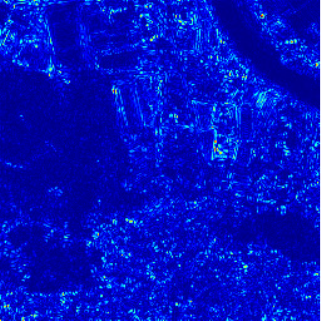} &
				\includegraphics[width=\mywgf]{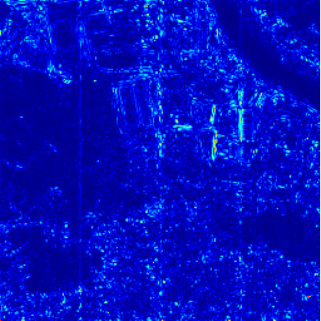} &
				\includegraphics[width=\mywgf]{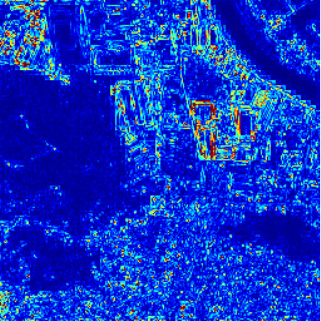} &
				\includegraphics[width=\mywgf]{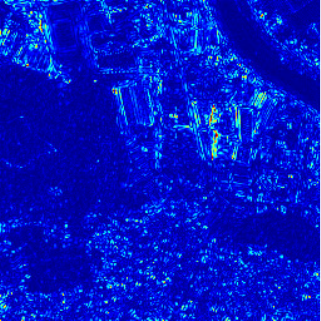} &
				\includegraphics[width=\mywgf]{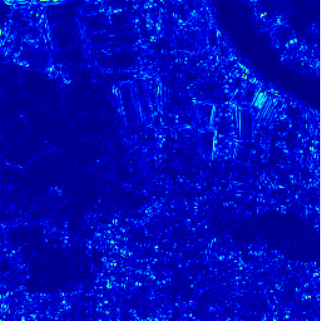} &
				\includegraphics[width=\mywgf]{./images/qb/rr/error_map/GT_GT_reference_rr1-eps-converted-to.pdf} & \\[-2pt]
				
				\small PSGAN & \small MSAN & \small PanMamba & \small MARNet & \small GSPan & \small GT & \\[-6pt]
			\end{tabular}
			
			\caption{Qualitative comparison between the proposed GSPan and other methods on the GF2 dataset. Rows 1 and 3 show pansharpened fused images, and Rows 2 and 4 show corresponding absolute error maps against the ground truth (GT).}
			\label{fig:rr_gf2_comparison}
		\end{figure*}

\begin{figure*}[htbp]
			\centering
			\captionsetup[subfloat]{labelformat=empty, skip=0pt}
			\setlength{\tabcolsep}{1pt} 
			
			\newlength{\mywwv}
			\setlength{\mywwv}{0.14\textwidth} 
			
			\begin{tabular}{ccccccl} 
				\includegraphics[width=\mywwv]{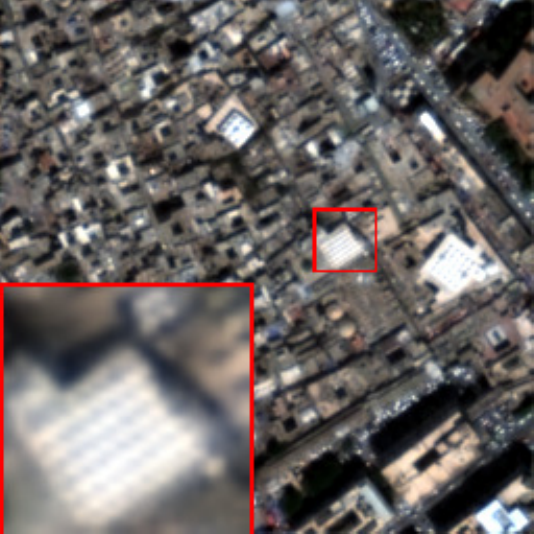} &
				\includegraphics[width=\mywwv]{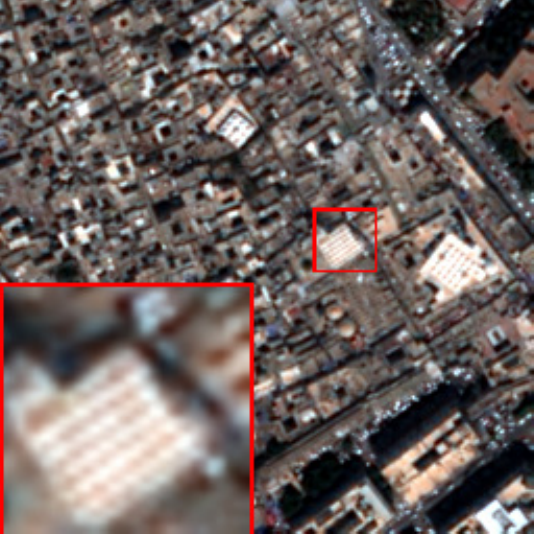} &
				\includegraphics[width=\mywwv]{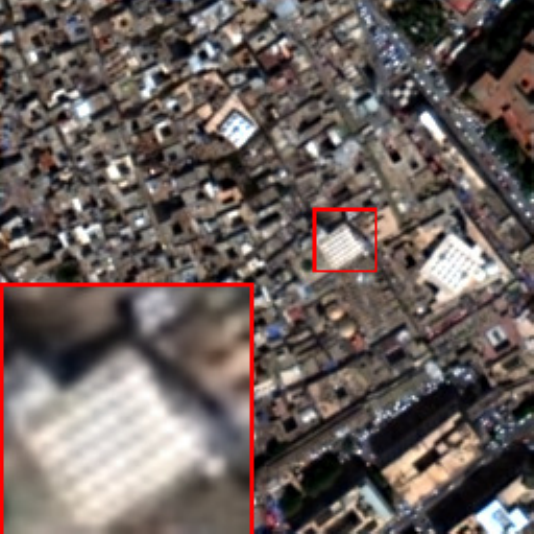} &
				\includegraphics[width=\mywwv]{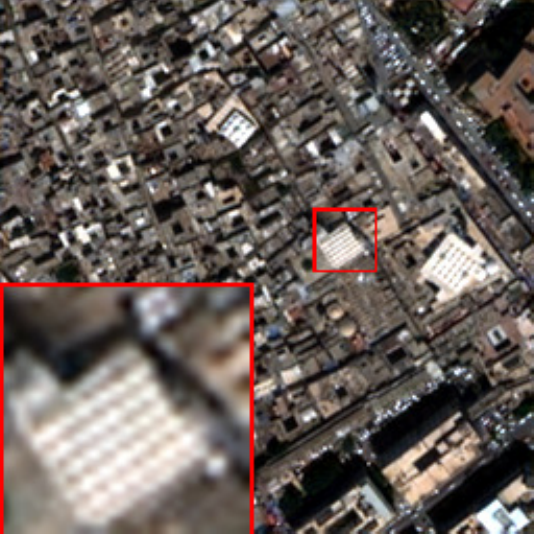} &
				\includegraphics[width=\mywwv]{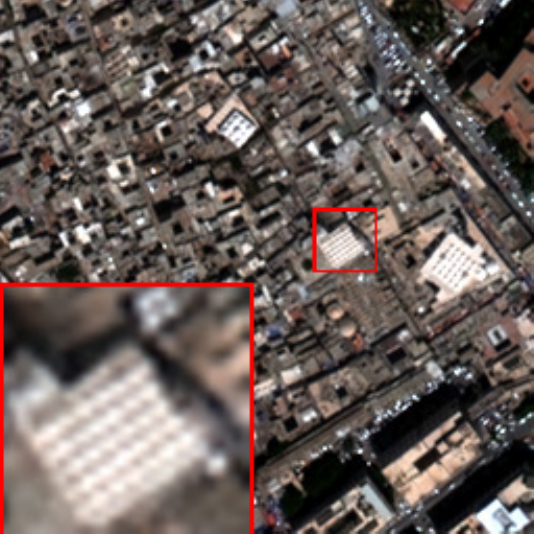} &
				\includegraphics[width=\mywwv]{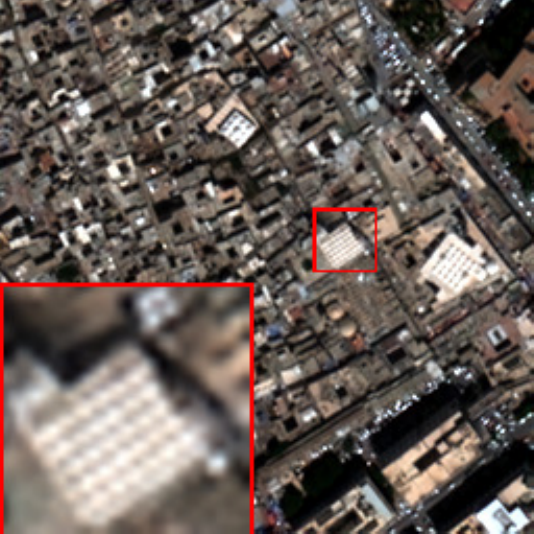} & 
		
				\multirow{6}{*}[\mywwv+0.5pt]{\includegraphics[height=\dimexpr 4\mywwv + 33pt\relax, trim=10 0 0 0, clip]{./images/qb/rr/error_map/global_colorbar.png}} \\[-1pt]
				
				\includegraphics[width=\mywwv]{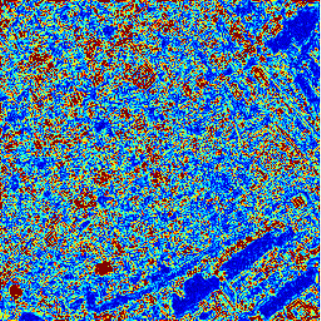} &
				\includegraphics[width=\mywwv]{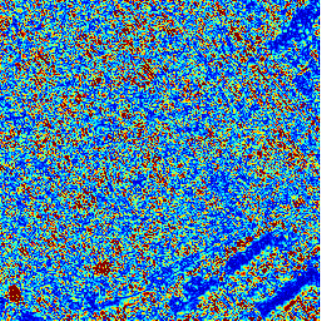} &
				\includegraphics[width=\mywwv]{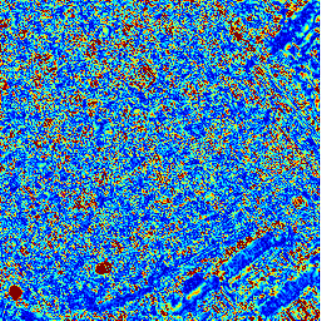} &
				\includegraphics[width=\mywwv]{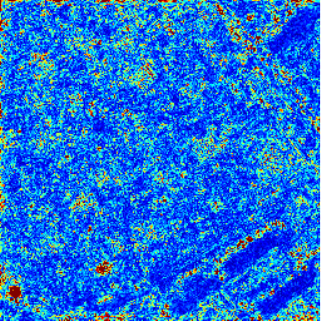} &
				\includegraphics[width=\mywwv]{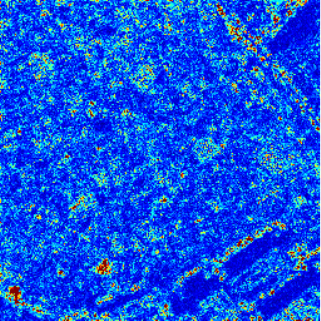} &
				\includegraphics[width=\mywwv]{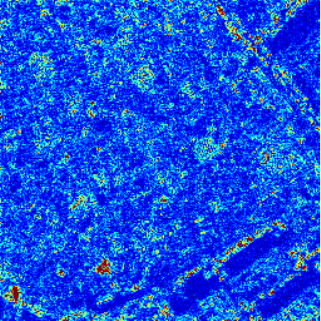} & \\[-2pt]
				
				\small AWLP & \small TV & \small BDSD & \small PNN & \small FusionNet & \small LAGConv & \\[-2pt] 
				
				\includegraphics[width=\mywwv]{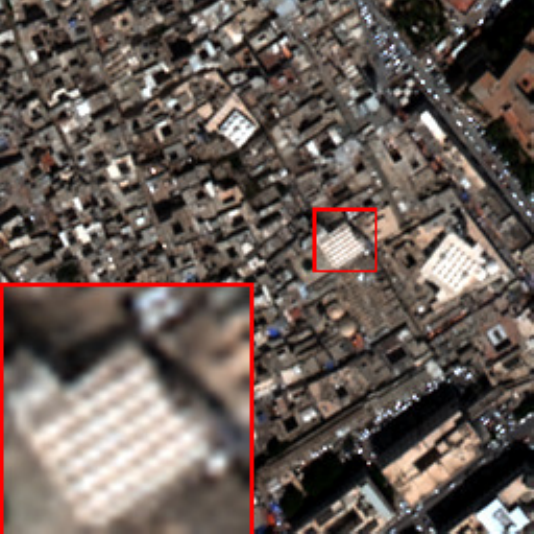} &
				\includegraphics[width=\mywwv]{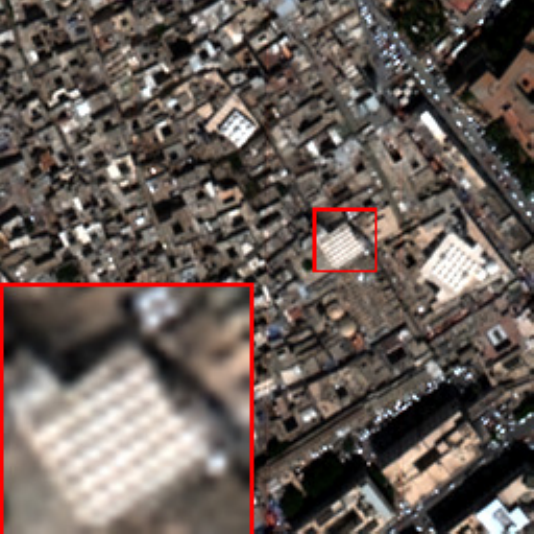} &
				\includegraphics[width=\mywwv]{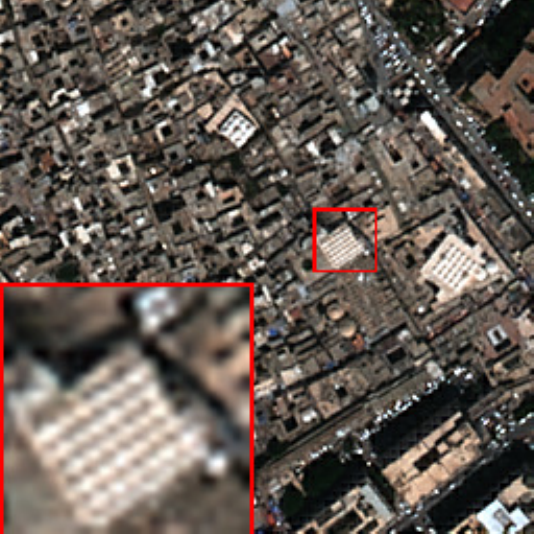} &
				\includegraphics[width=\mywwv]{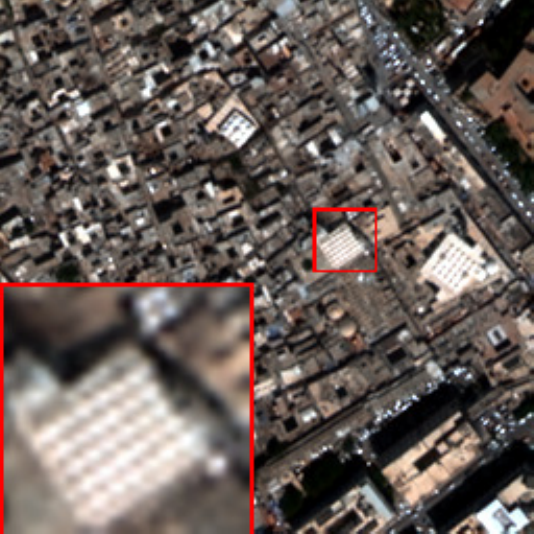} &
				\includegraphics[width=\mywwv]{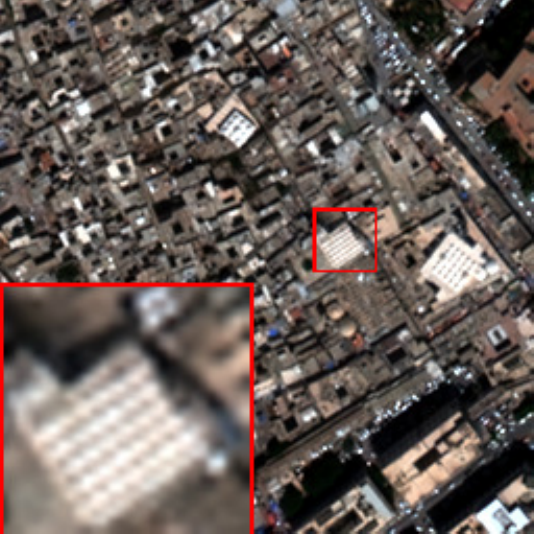} &
				\includegraphics[width=\mywwv]{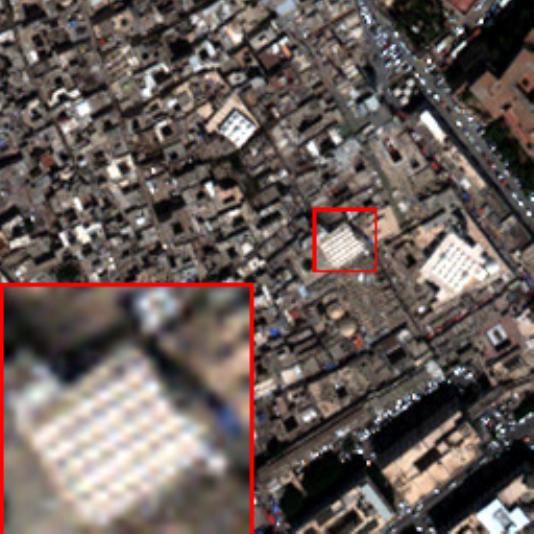} & \\[-1.5pt]
				
				\includegraphics[width=\mywwv]{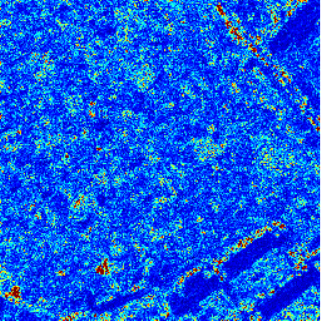} &
				\includegraphics[width=\mywwv]{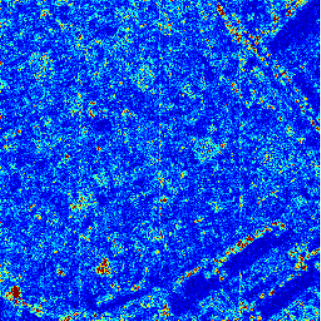} &
				\includegraphics[width=\mywwv]{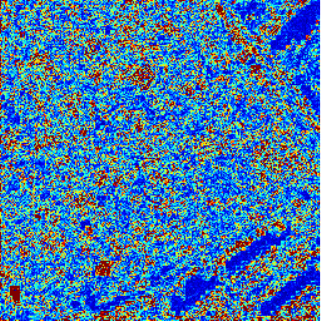} &
				\includegraphics[width=\mywwv]{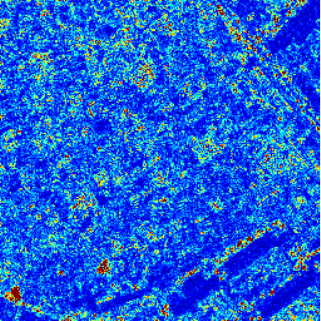} &
				\includegraphics[width=\mywwv]{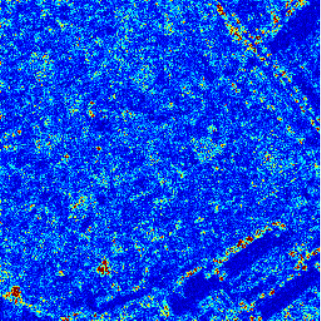} &
				\includegraphics[width=\mywwv]{./images/qb/rr/error_map/GT_GT_reference_rr1-eps-converted-to.pdf} & \\[-2pt]
				
				\small PSGAN & \small MSAN & \small PanMamba & \small MARNet & \small GSPan & \small GT & \\[-6pt]
			\end{tabular}
			
			\caption{Qualitative comparison between the proposed GSPan and other methods on the WV3 dataset. Rows 1 and 3 show pansharpened fused images, and Rows 2 and 4 show corresponding absolute error maps against the ground truth (GT).}
			\label{fig:rr_wv3_comparison}
		\end{figure*}

		Qualitatively, Fig.~\ref{fig:rr_qb_comparison}, Fig.~\ref{fig:rr_gf2_comparison}, and Fig.~\ref{fig:rr_wv3_comparison} illustrate that GSPan produces superior visual results with sharper edges and fewer artifacts compared to competing methods. In the zoomed-in patches of the QB and WV3 datasets, our method accurately restores complex spatial structures, such as building contours and fine roof textures. Furthermore, the absolute error maps in the bottom rows reveal that GSPan yields the most uniform and darkest blue regions, indicating the smallest residuals relative to the ground truth. These visual results suggest that our Gaussian synthesis mechanism effectively leverages spatial guidance from the PAN image to refine the MS bands while maintaining a high degree of spectral fidelity.

		\subsubsection{Evaluation on full-resolution scenes}

        For full-resolution evaluation, GSPan also demonstrates strong overall performance in terms of HQNR and its distortion components. On the QB and GF2 datasets, GSPan achieves the best results across all three full-resolution metrics, including $D_\lambda$, $D_s$, and HQNR. On the WV3 dataset, GSPan obtains the lowest spectral distortion $D_\lambda$ and the highest HQNR, while its spatial distortion $D_s$ is slightly higher than the best-performing competing method. This result suggests that GSPan tends to preserve spectral consistency well, while maintaining competitive spatial fidelity in real-world full-resolution scenes.

\begin{figure*}[htbp]
			\centering
			
			\captionsetup[subfloat]{labelformat=empty, skip=0pt}
			\setlength{\tabcolsep}{1pt} 
			
			\newlength{\mywfrqb}
			\setlength{\mywfrqb}{0.14\textwidth} 
			
			\begin{tabular}{ccccccl} 
				\includegraphics[width=\mywfrqb]{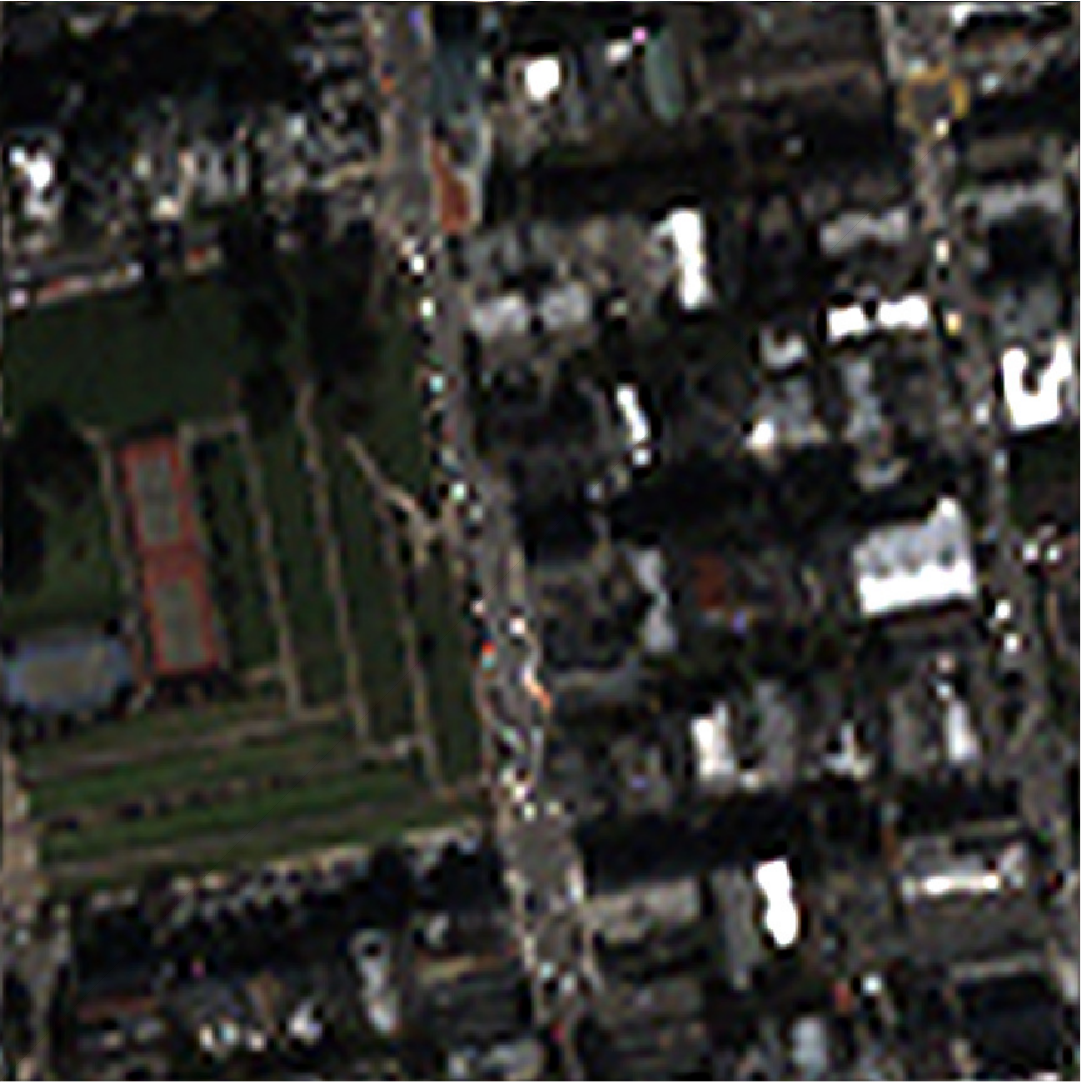} &
				\includegraphics[width=\mywfrqb]{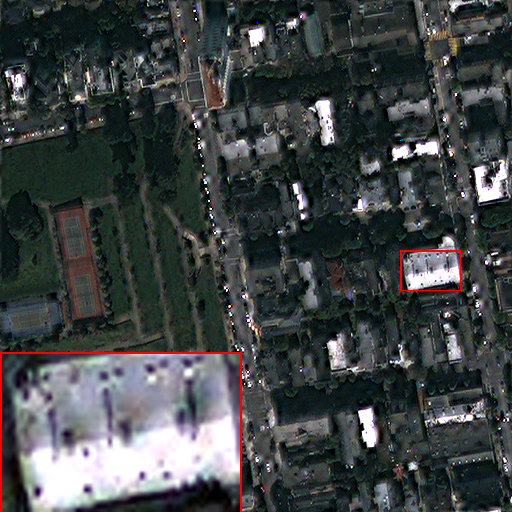} &
				\includegraphics[width=\mywfrqb]{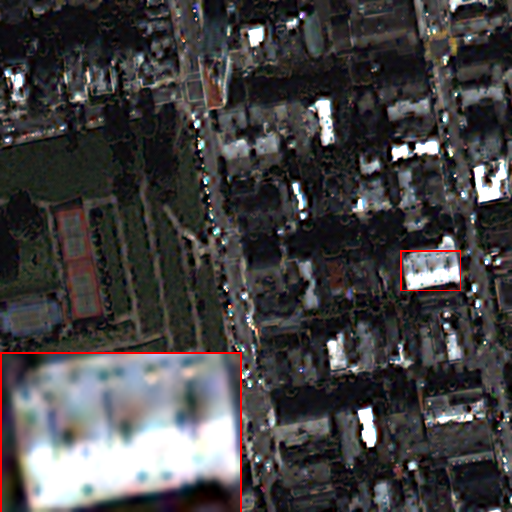} &
				\includegraphics[width=\mywfrqb]{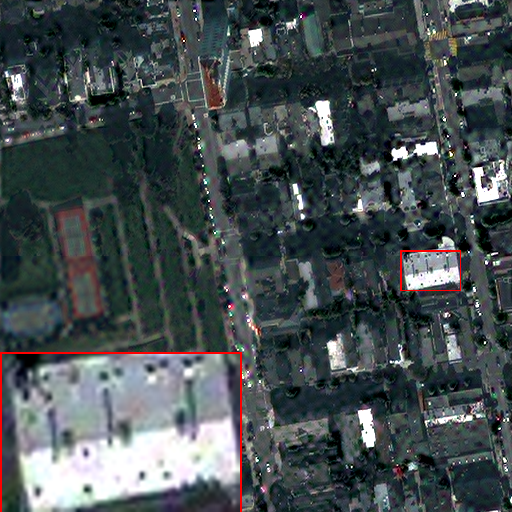} &
				\includegraphics[width=\mywfrqb]{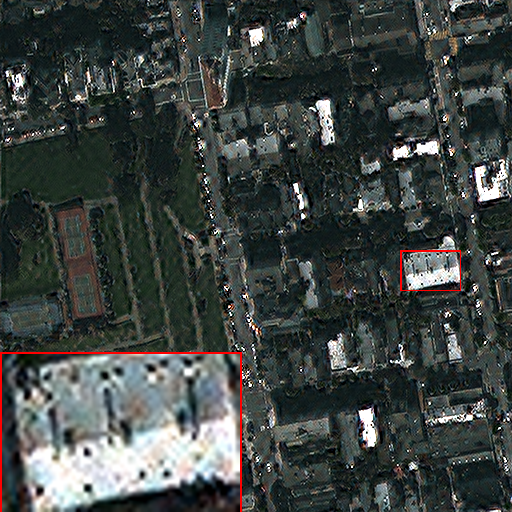} &
				\includegraphics[width=\mywfrqb]{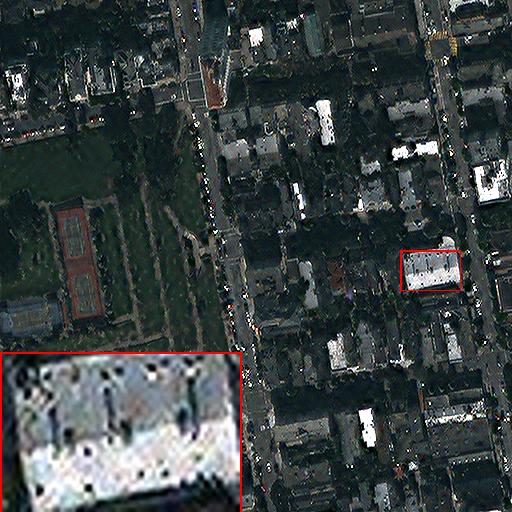} & 
				
				\multirow{6}{*}[\mywfrqb+1pt]{\includegraphics[height=\dimexpr 4\mywfrqb + 36pt\relax, trim=8 0 0 0, clip]{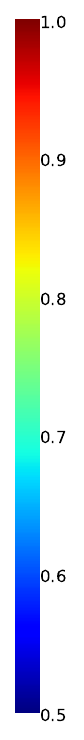}} \\[-1pt]
				
				\includegraphics[width=\mywfrqb]{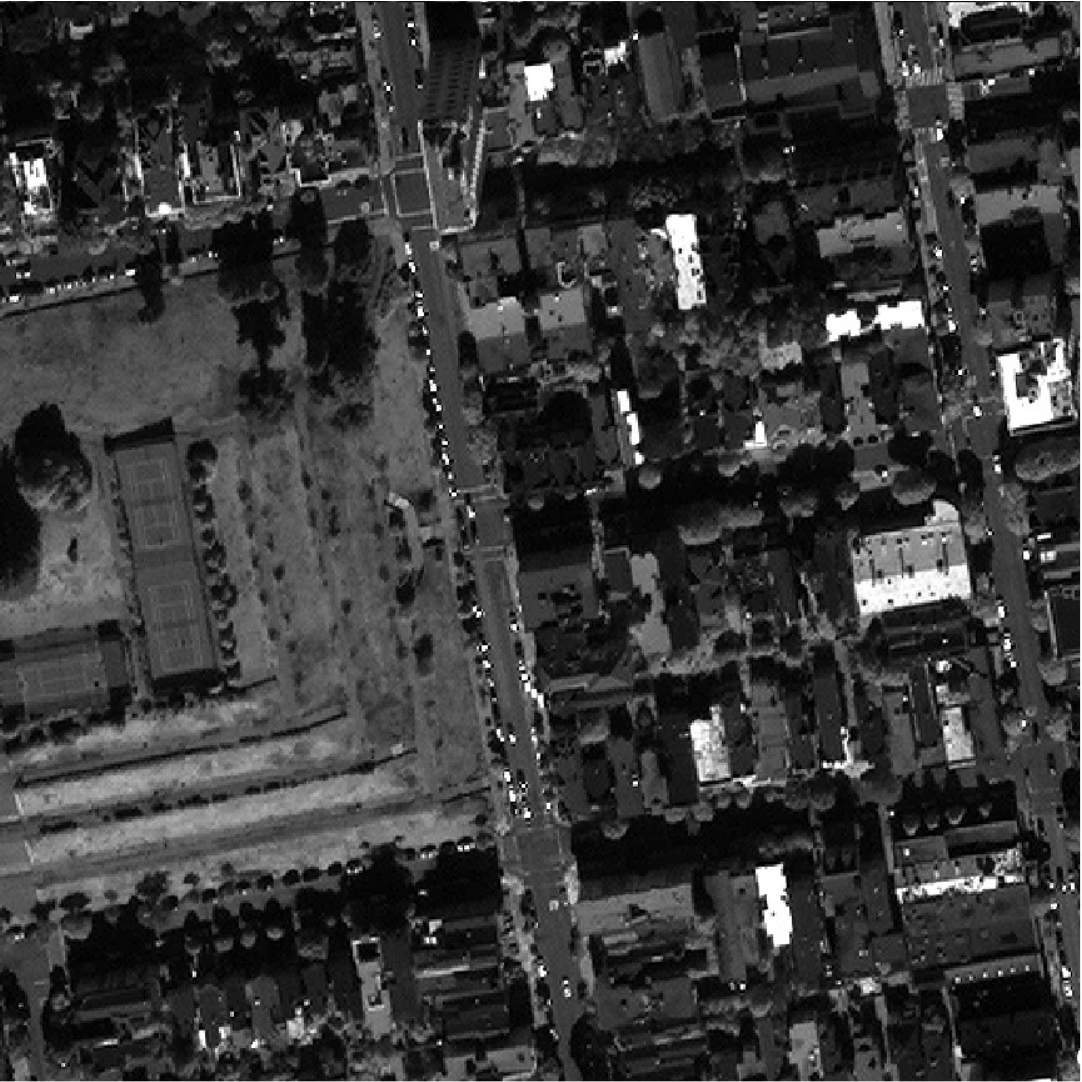} &
				\includegraphics[width=\mywfrqb]{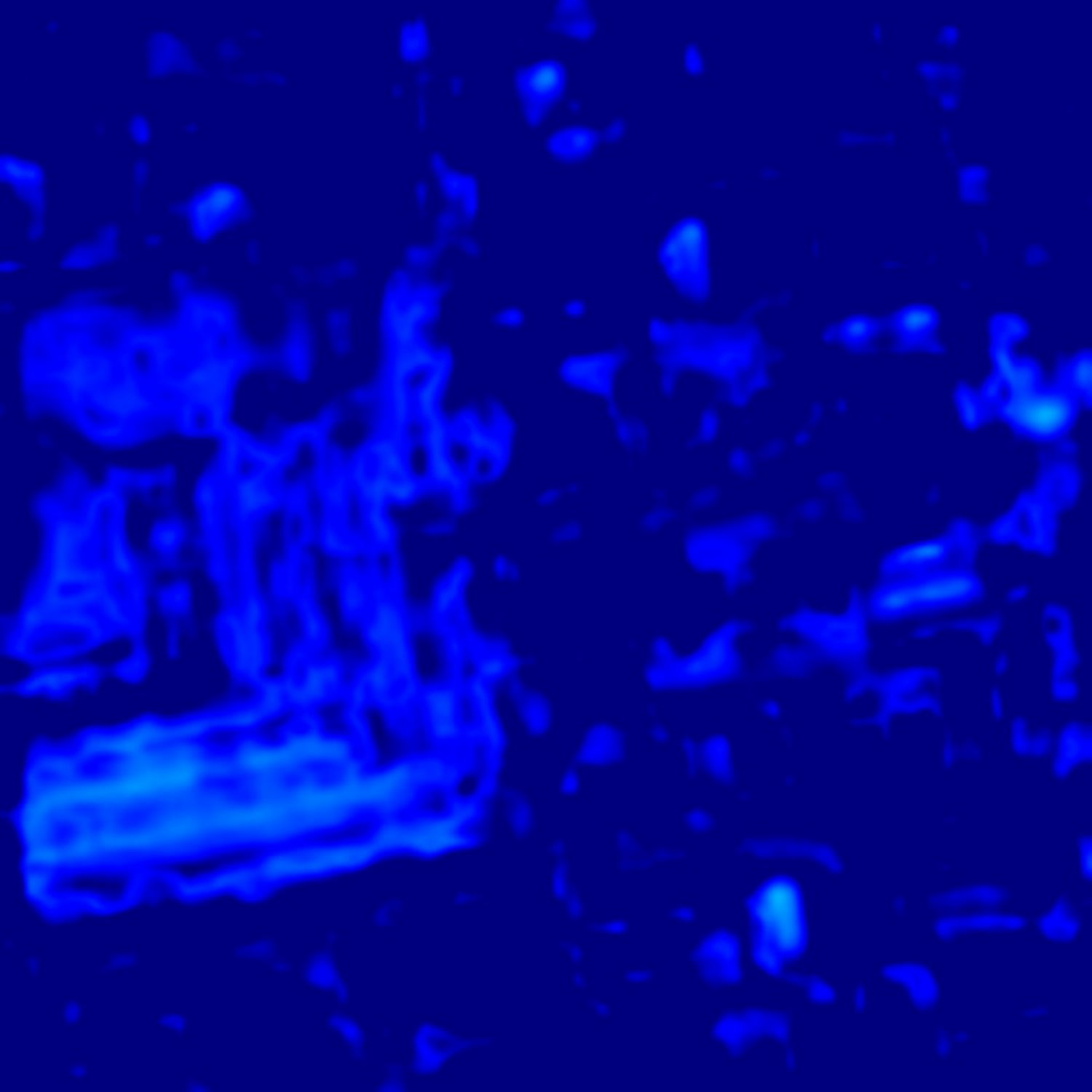} &
				\includegraphics[width=\mywfrqb]{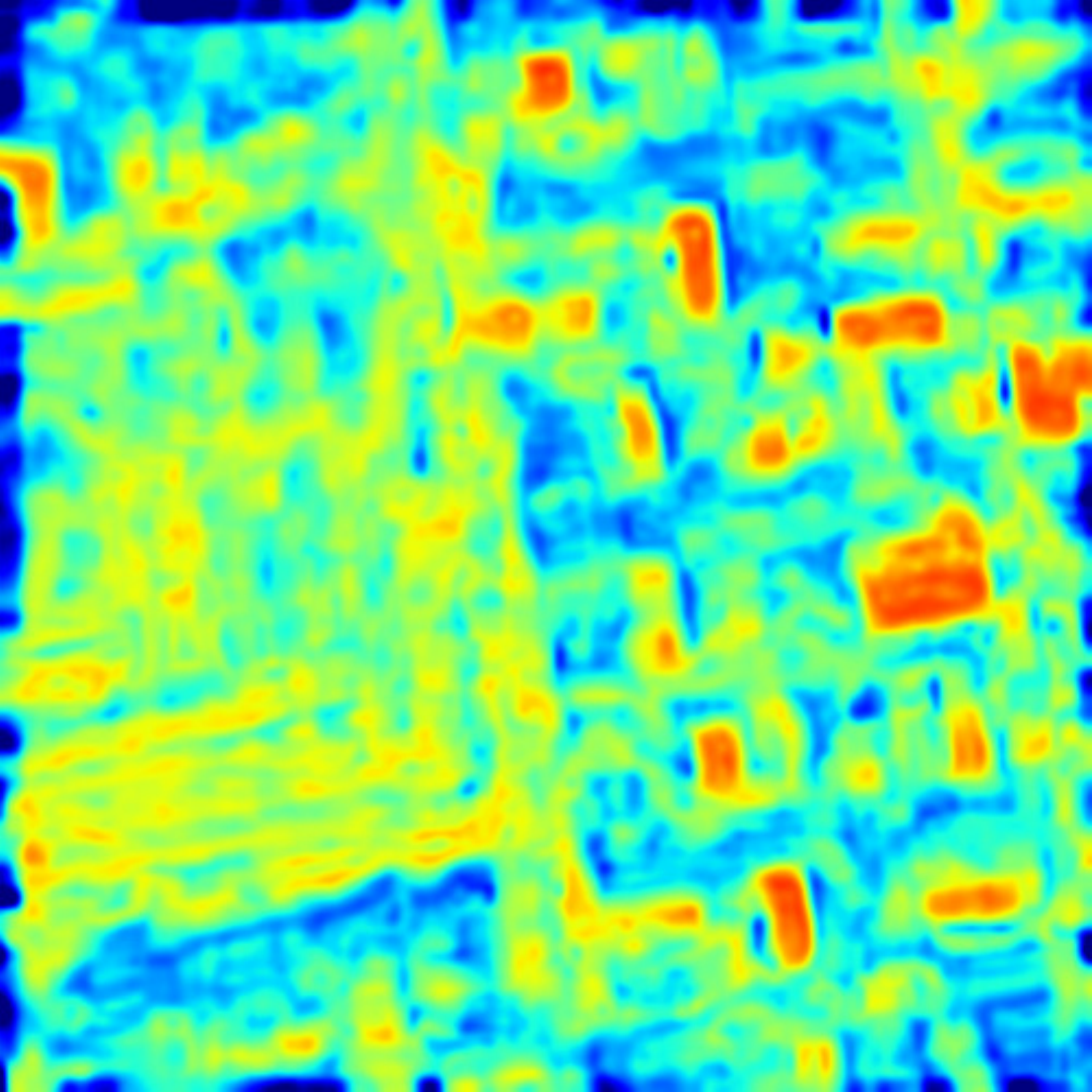} &
				\includegraphics[width=\mywfrqb]{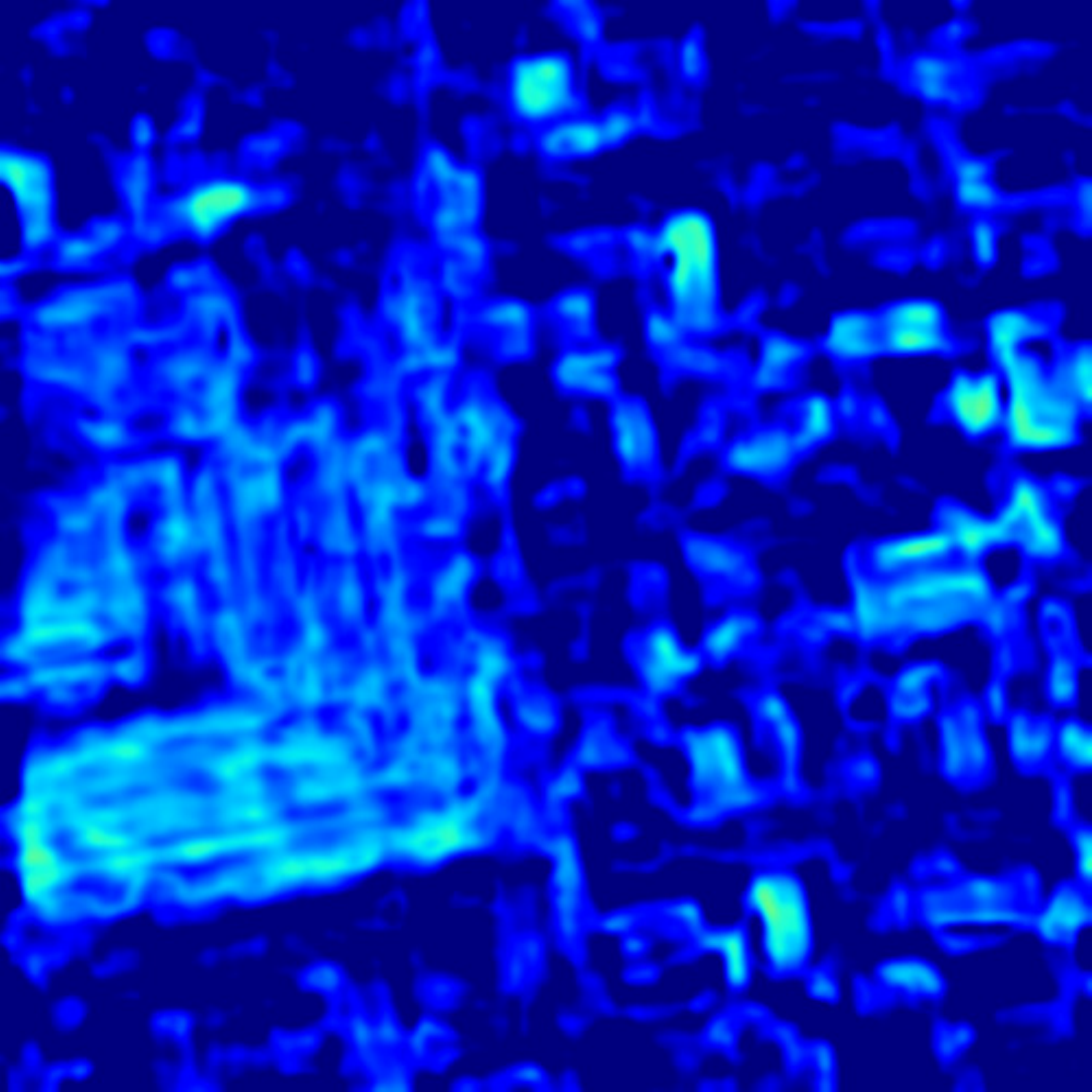} &
				\includegraphics[width=\mywfrqb]{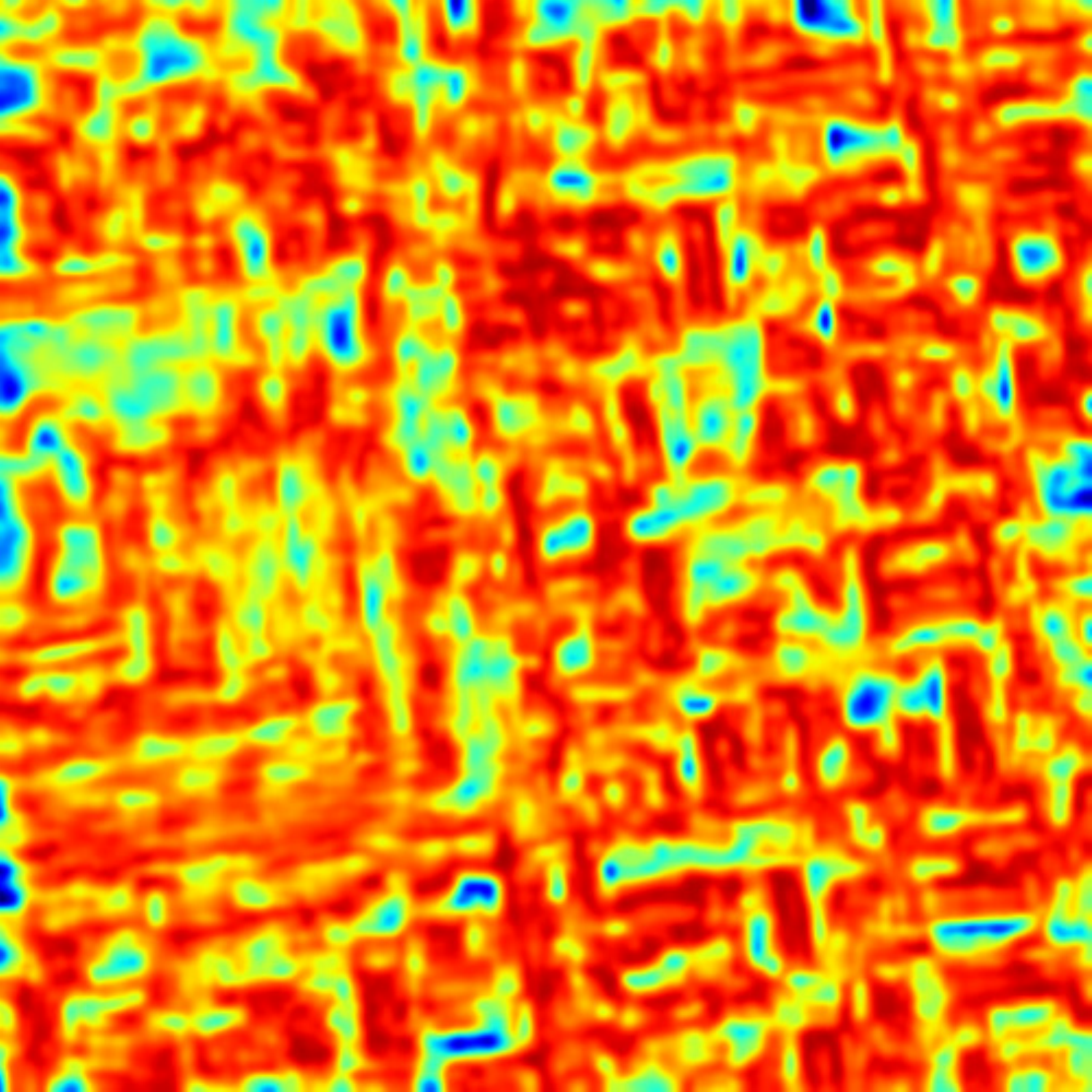} &
				\includegraphics[width=\mywfrqb]{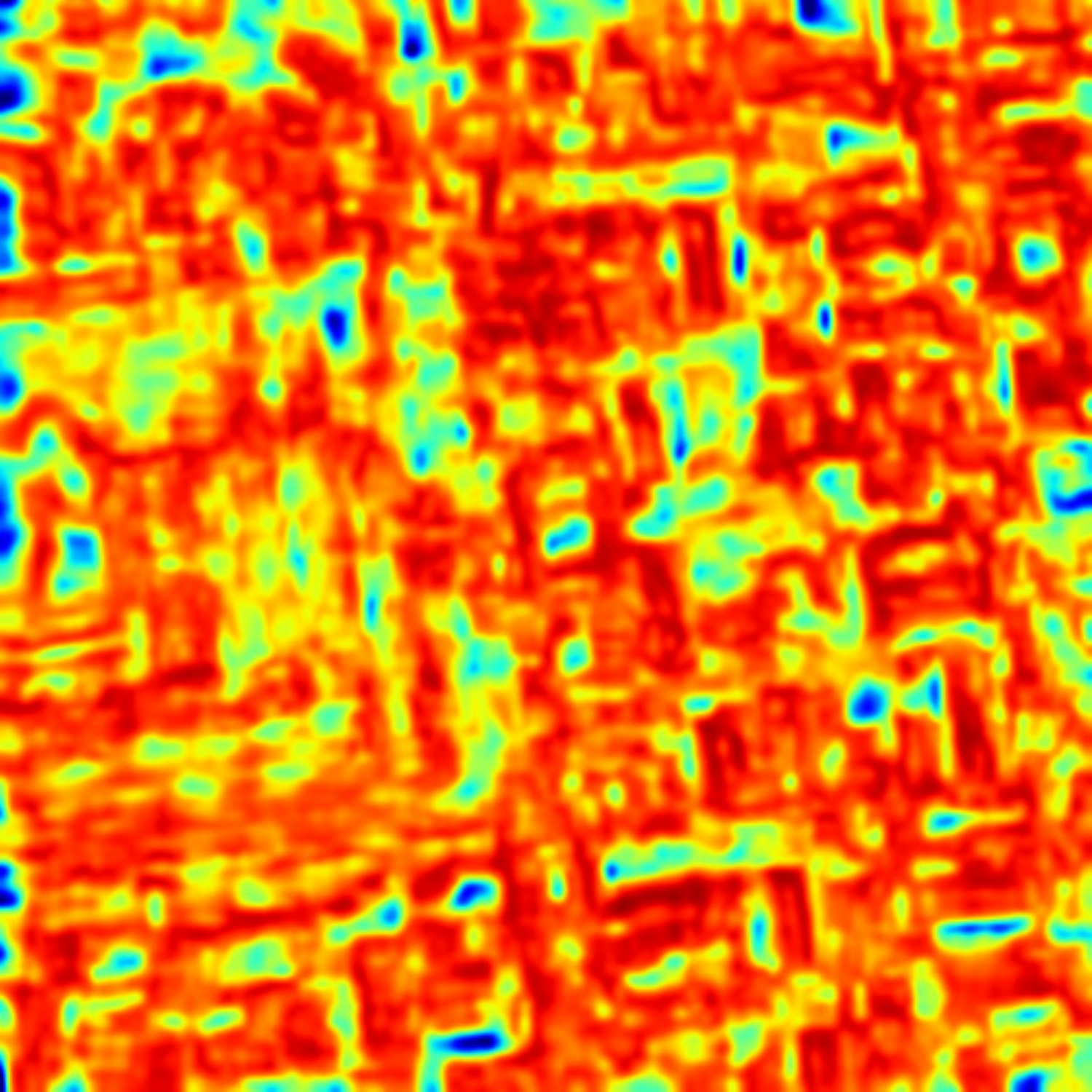} & \\[-2pt]
				
				\small $\widetilde{MS}$/PAN & \small AWLP & \small TV & \small BDSD & \small PNN & \small FusionNet & \\[-2pt] 
				
				\includegraphics[width=\mywfrqb]{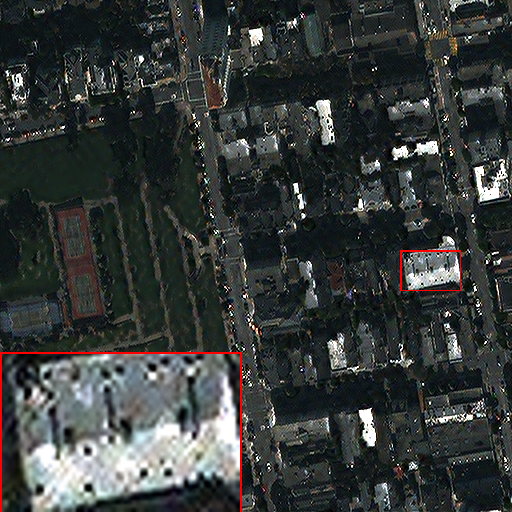} &
				\includegraphics[width=\mywfrqb]{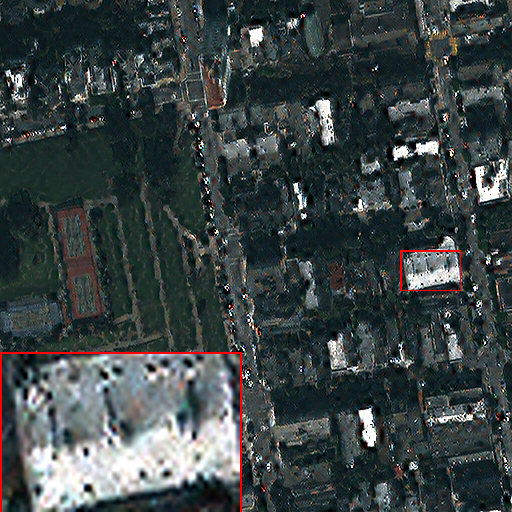} &
				\includegraphics[width=\mywfrqb]{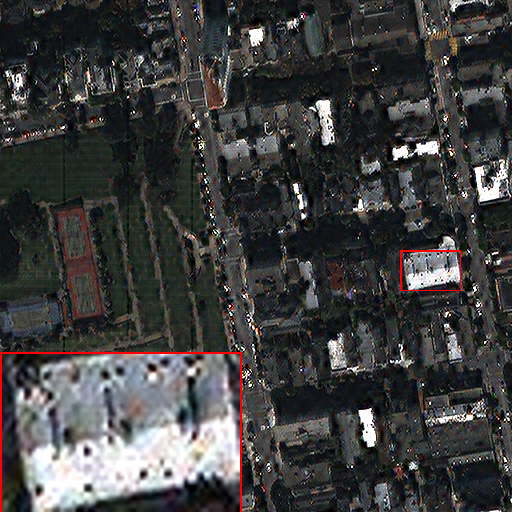} &
				\includegraphics[width=\mywfrqb]{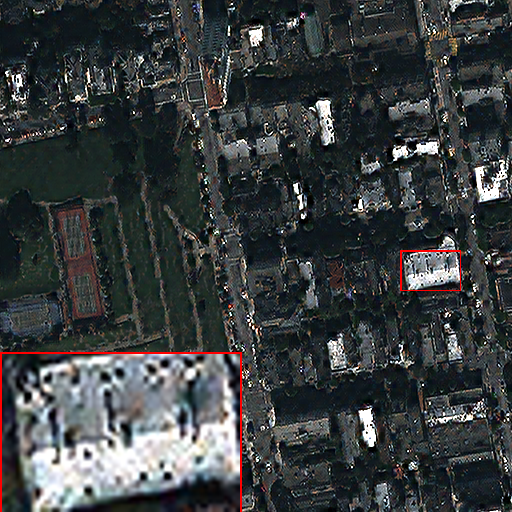} &
				\includegraphics[width=\mywfrqb]{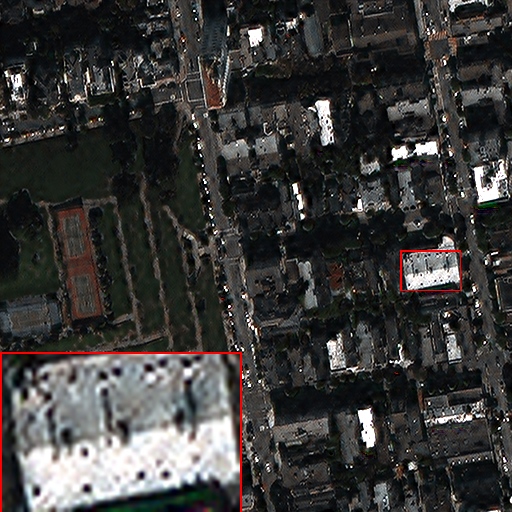} &
				\includegraphics[width=\mywfrqb]{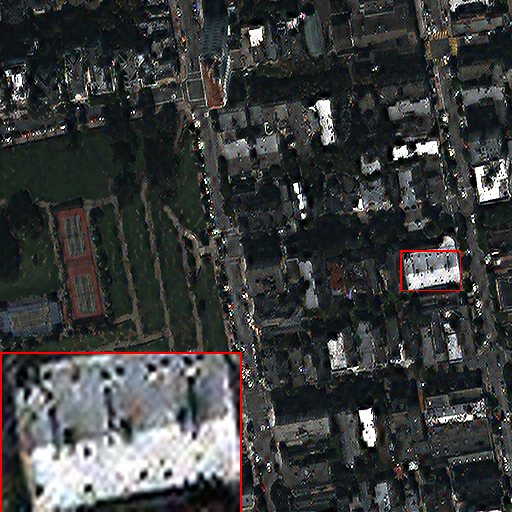} & \\[-1pt]
				
				\includegraphics[width=\mywfrqb]{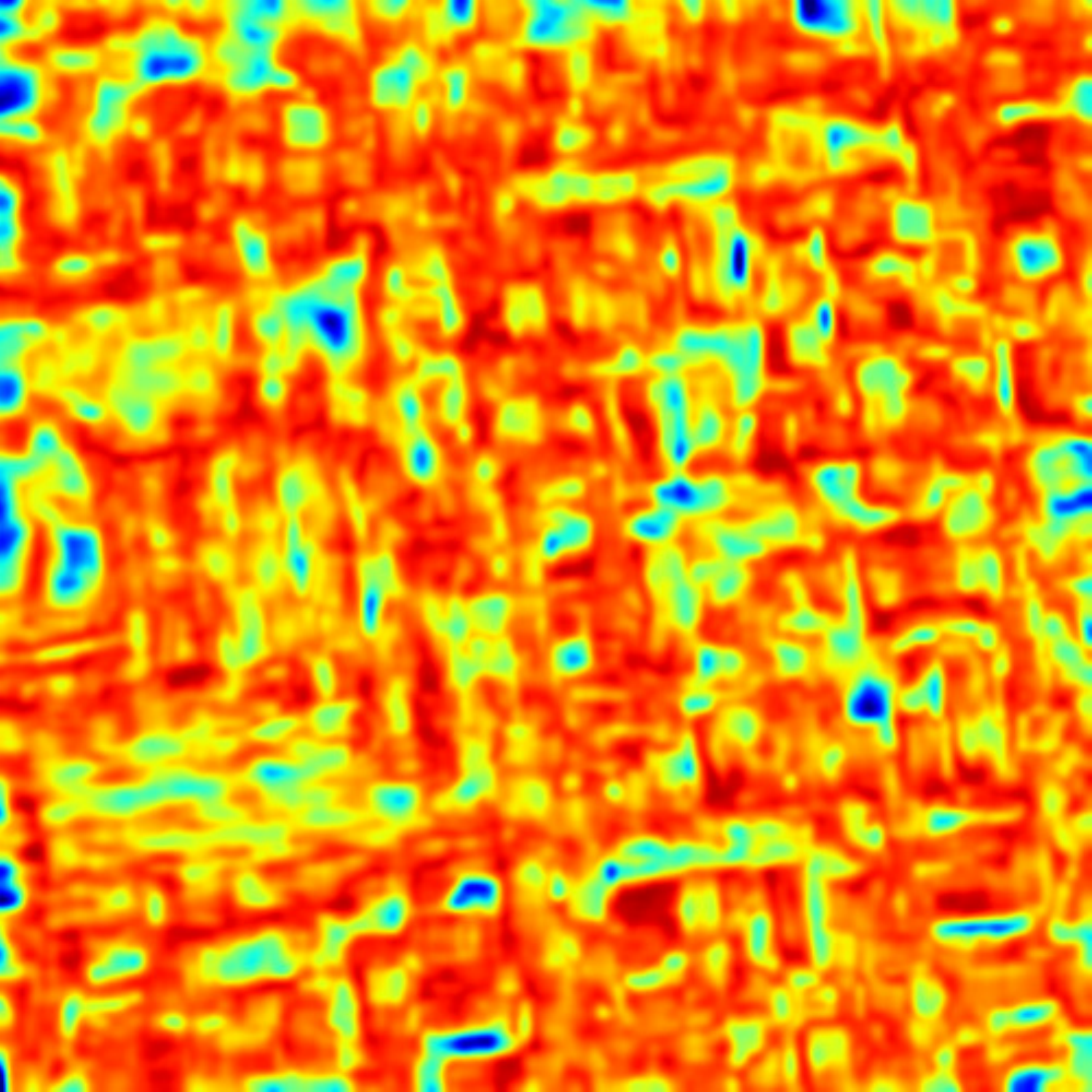} &
				\includegraphics[width=\mywfrqb]{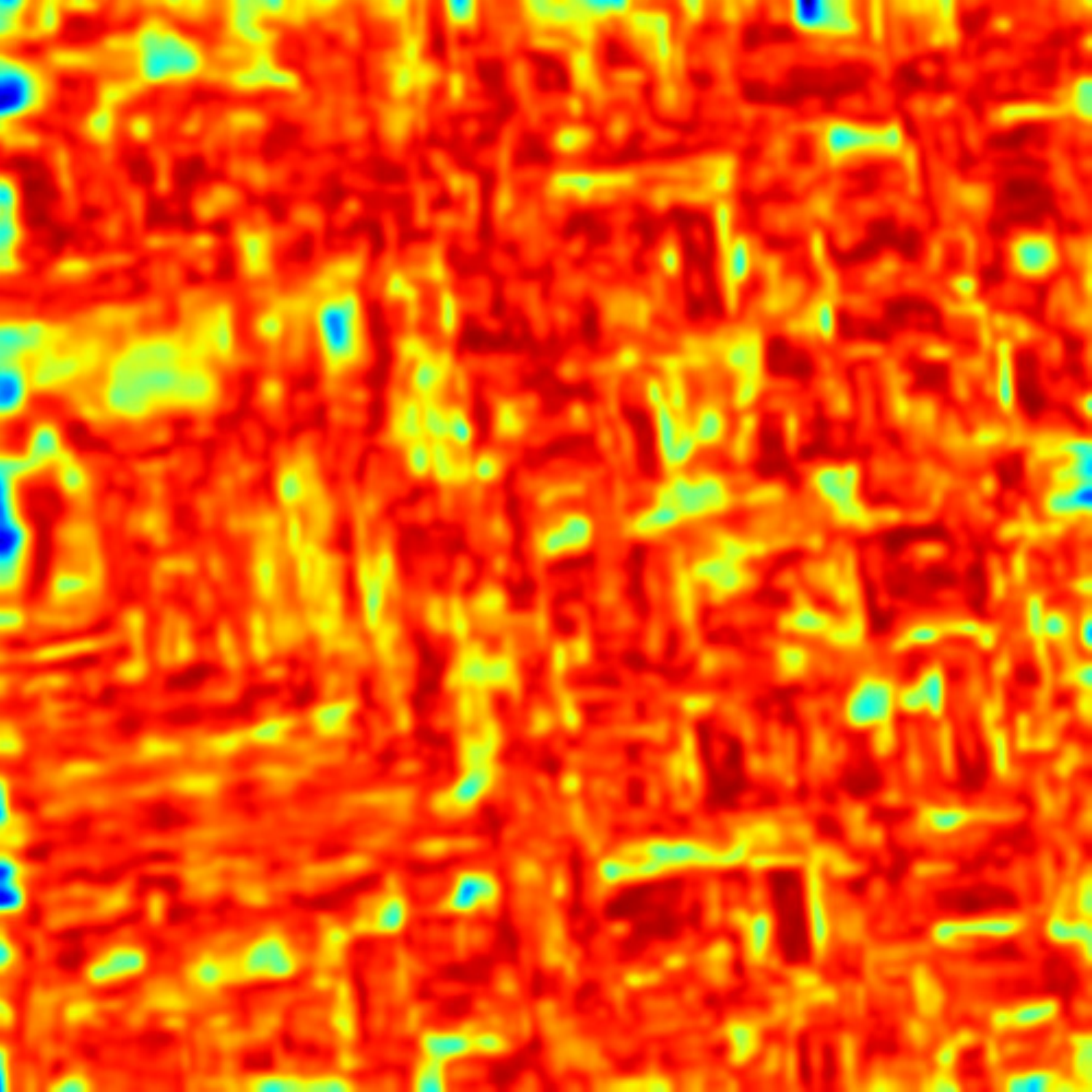} &
				\includegraphics[width=\mywfrqb]{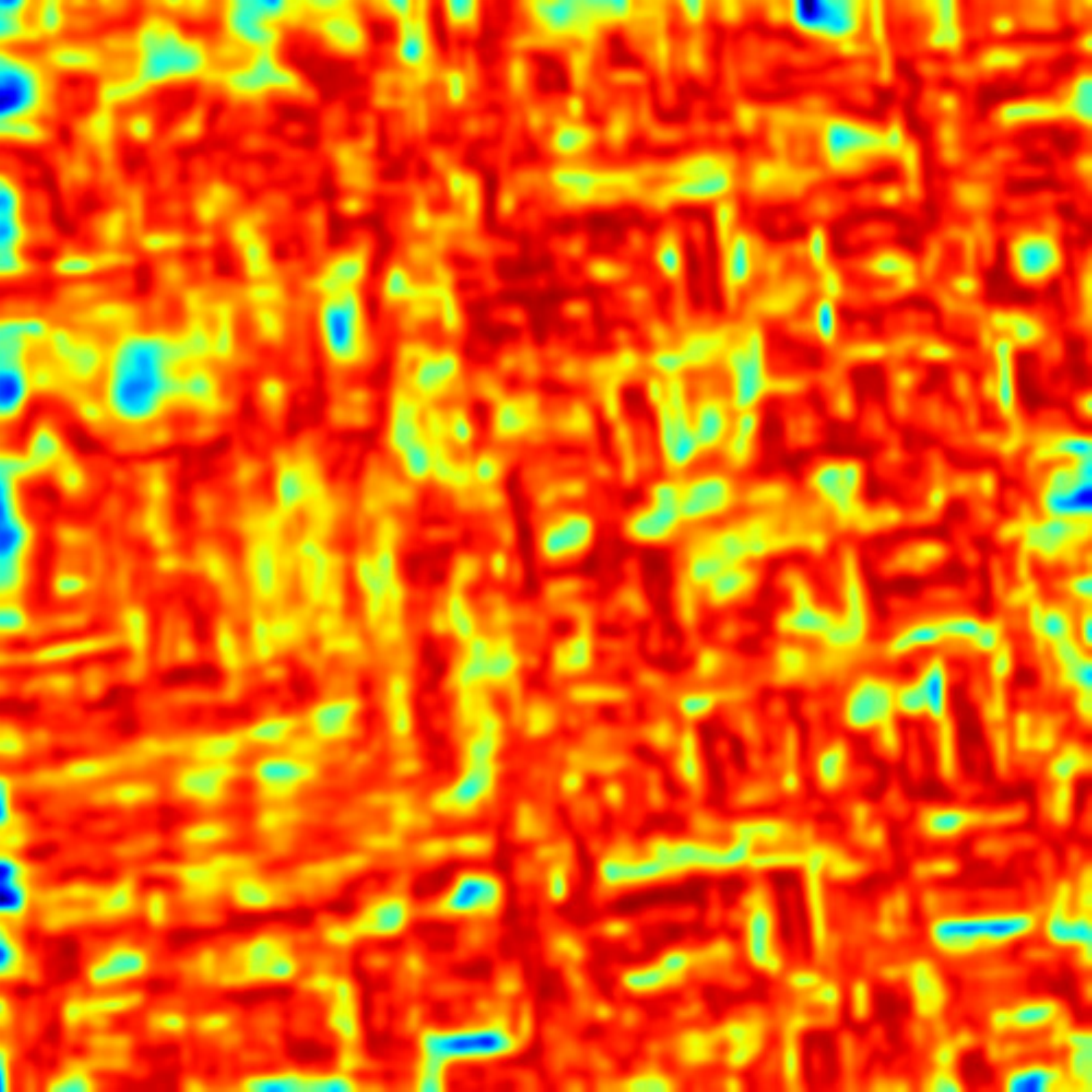} &
				\includegraphics[width=\mywfrqb]{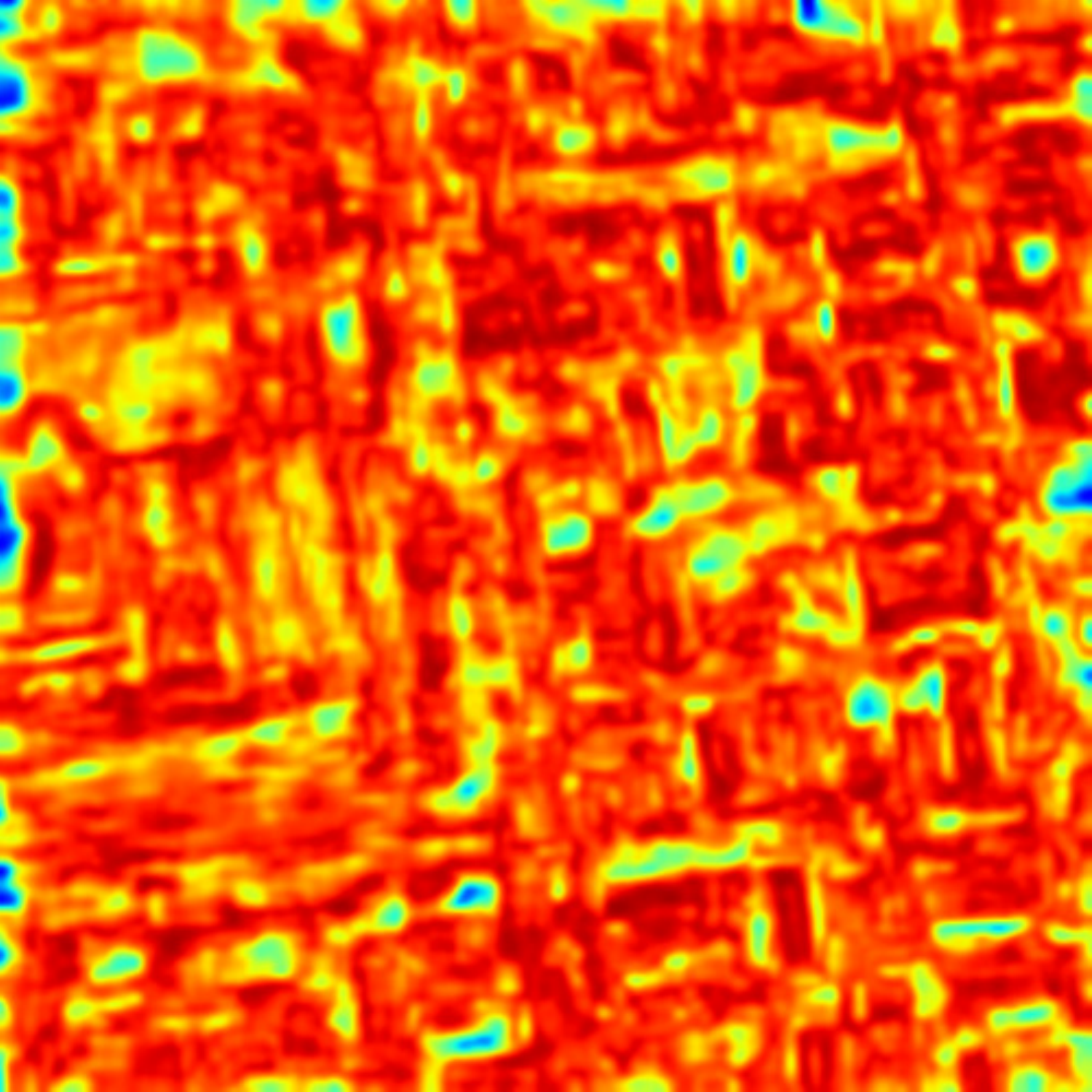} &
				\includegraphics[width=\mywfrqb]{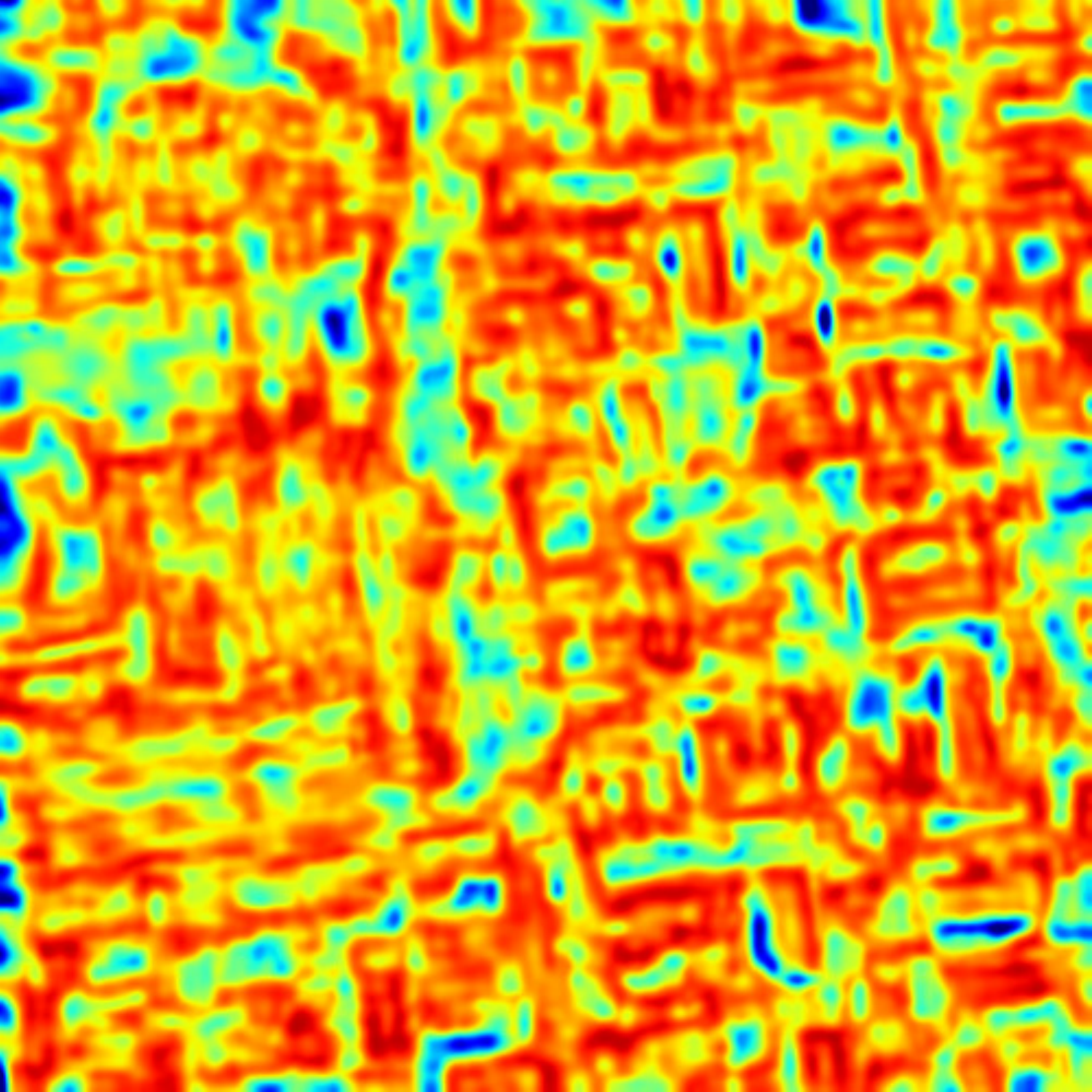} &
				\includegraphics[width=\mywfrqb]{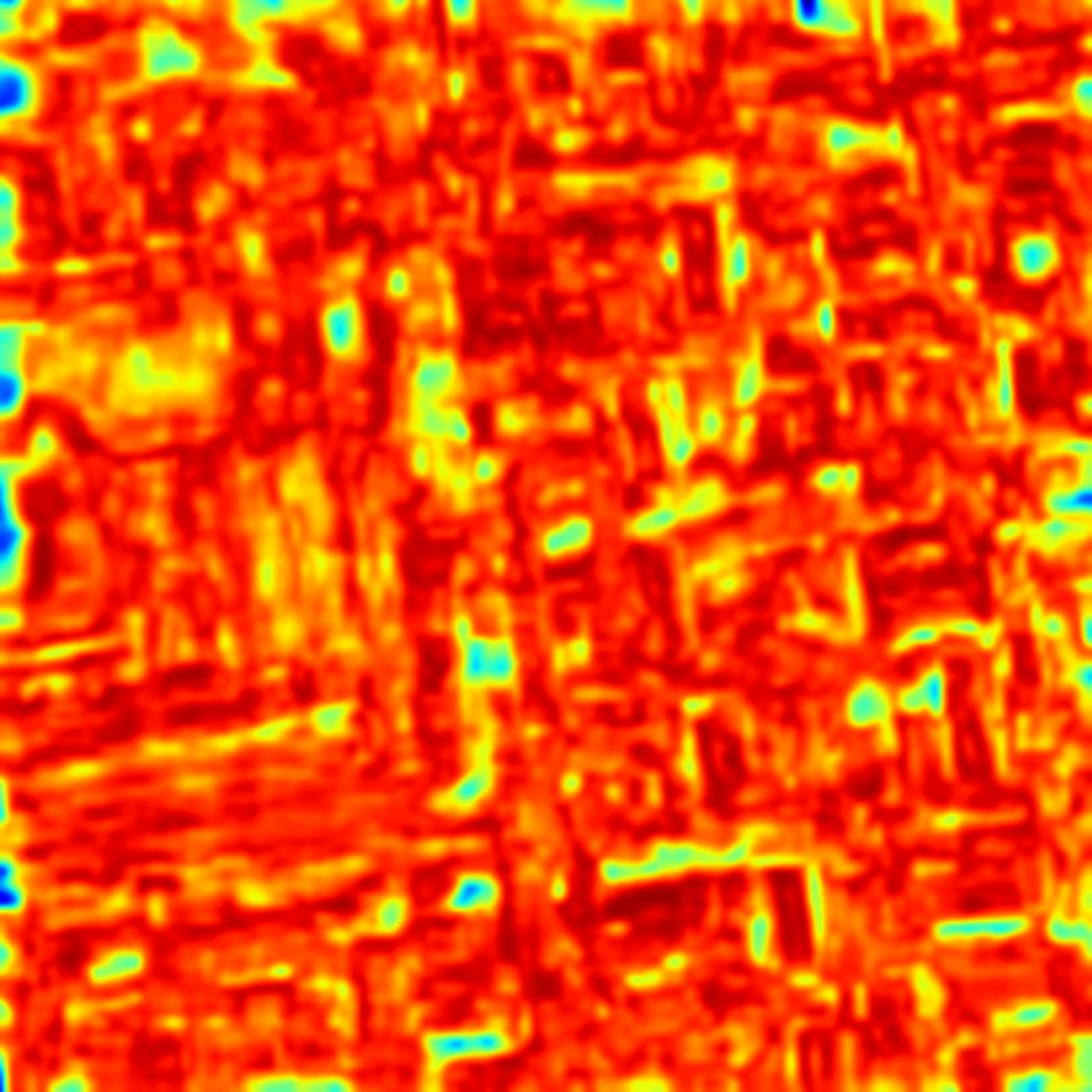} & \\[-2pt]
				
				\small LAGConv & \small PSGAN & \small MSAN & \small PanMamba & \small MARNet & \small GSPan & \\[-6pt]
			\end{tabular}
			
			\caption{Qualitative comparison on the QB dataset under full-resolution testing. The top and third rows show the fused images, while the second and fourth rows display the HQNR maps.}
			\label{fig:qb_fr_comparison}
		\end{figure*}

\begin{figure*}[htbp]
		\centering
		
		\captionsetup[subfloat]{labelformat=empty, skip=0pt}
		\setlength{\tabcolsep}{1pt}

		\newlength{\mywfr}
		\setlength{\mywfr}{0.145\textwidth} 
		
		\begin{tabular}{ccccccl} 
			\includegraphics[width=\mywfr]{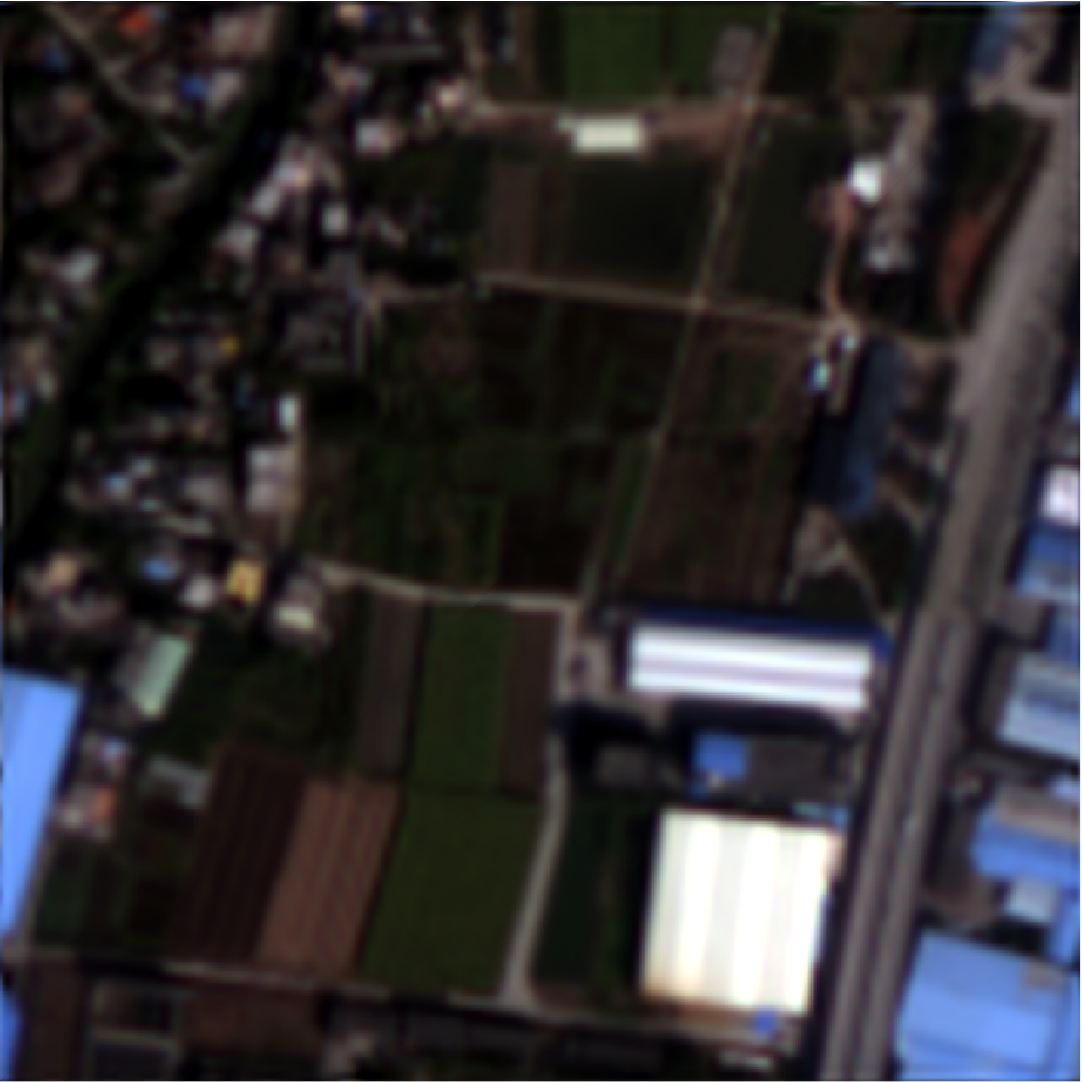} &
			\includegraphics[width=\mywfr]{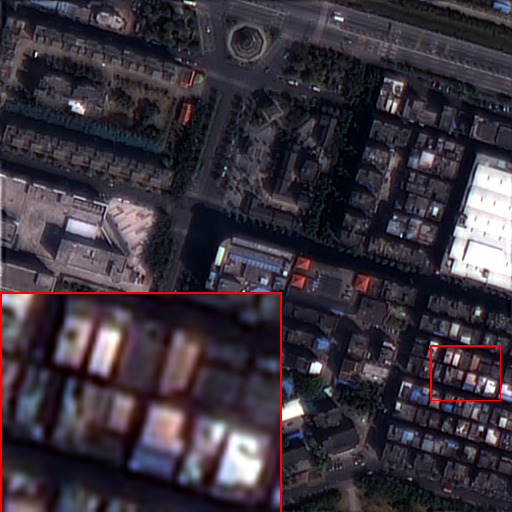} &
			\includegraphics[width=\mywfr]{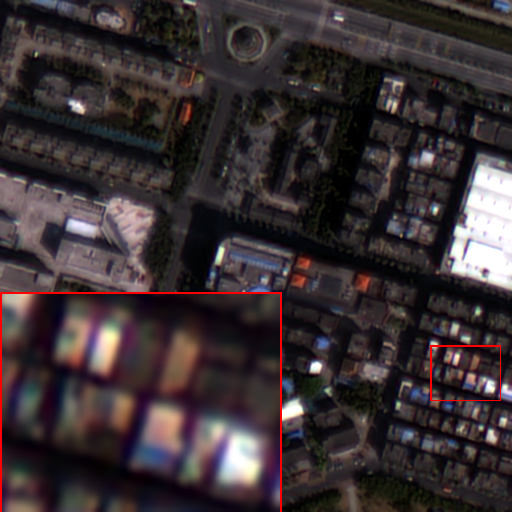} &
			\includegraphics[width=\mywfr]{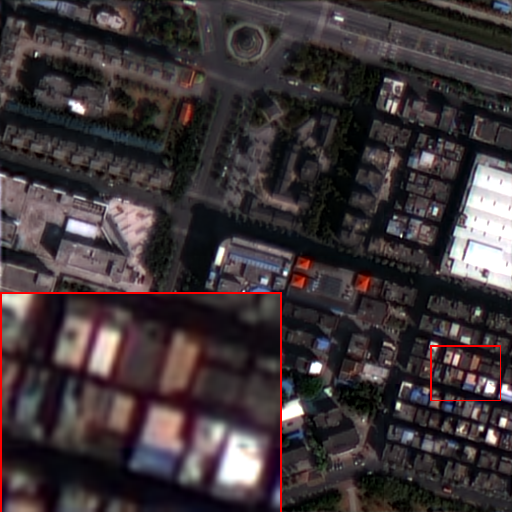} &
			\includegraphics[width=\mywfr]{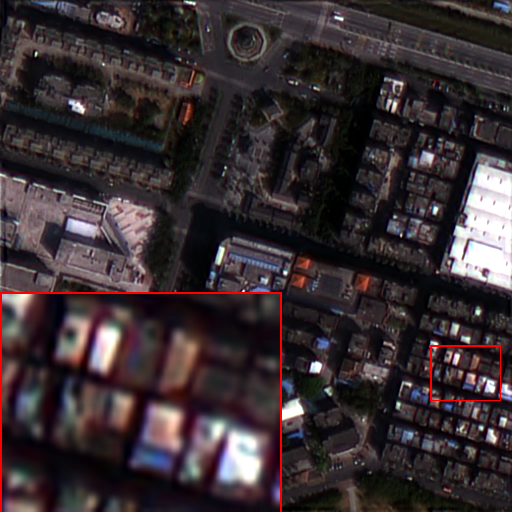} &
			\includegraphics[width=\mywfr]{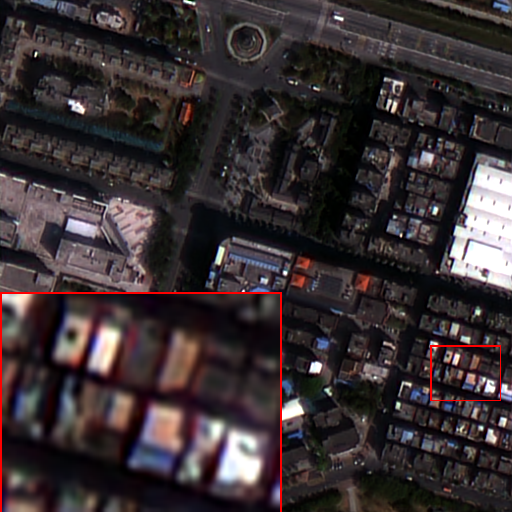} & 
			
			\multirow{6}{*}[\mywfr+1pt]{\includegraphics[height=\dimexpr 4\mywfr + 36pt\relax, trim=8 0 0 0, clip]{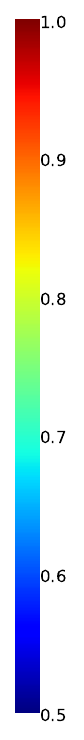}} \\[-1pt]
			
			\includegraphics[width=\mywfr]{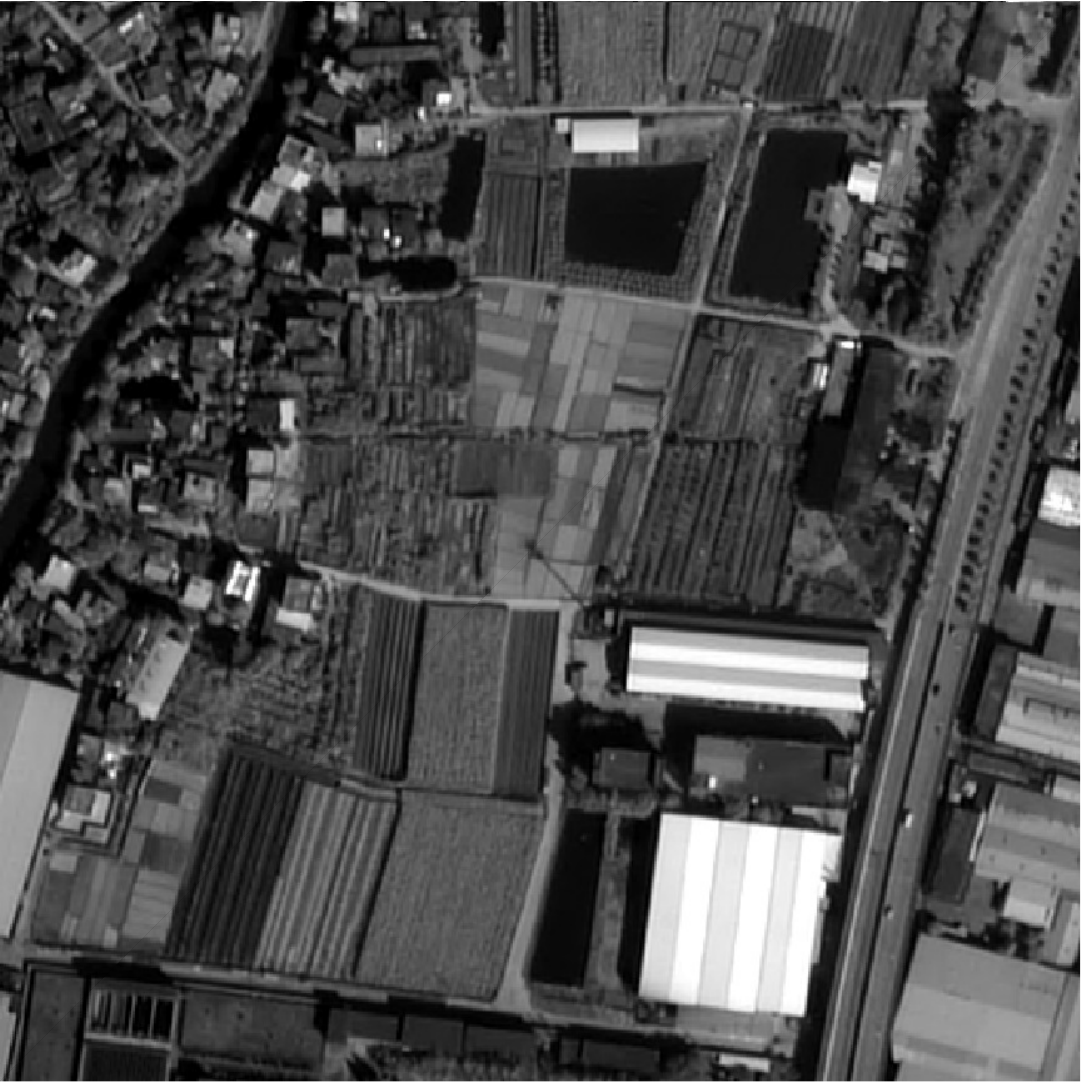} &
			\includegraphics[width=\mywfr]{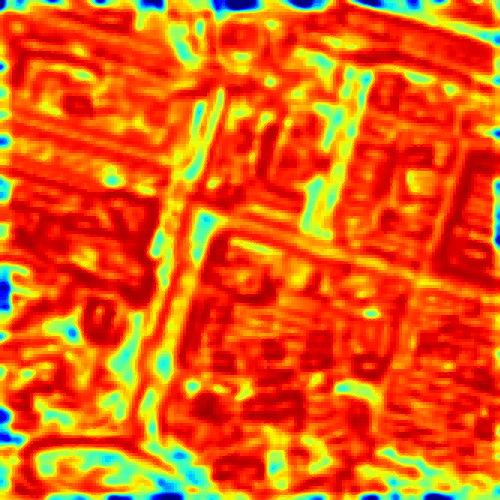} &
			\includegraphics[width=\mywfr]{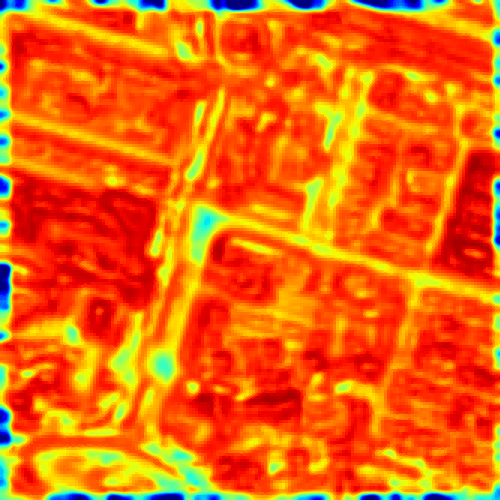} &
			\includegraphics[width=\mywfr]{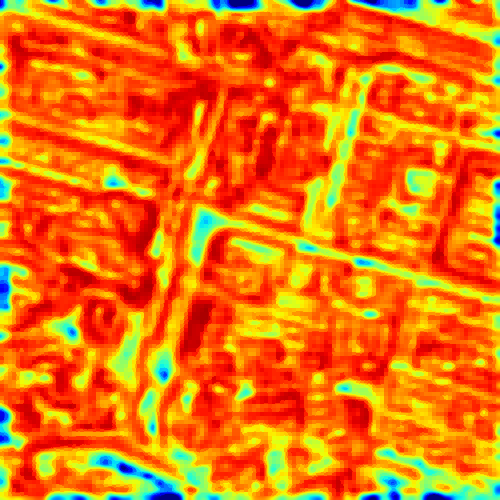} &
			\includegraphics[width=\mywfr]{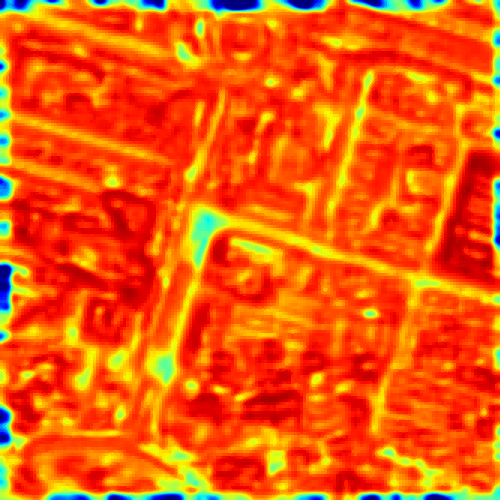} &
			\includegraphics[width=\mywfr]{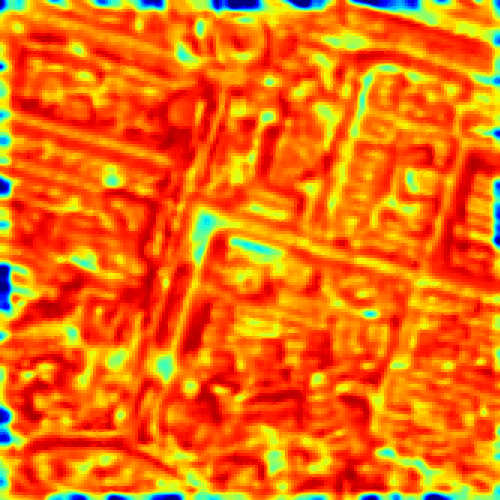} & \\[-2pt]
			
			\small $\widetilde{MS}$/PAN & \small AWLP & \small TV & \small BDSD & \small PNN & \small FusionNet & \\[-2pt] 
			
			\includegraphics[width=\mywfr]{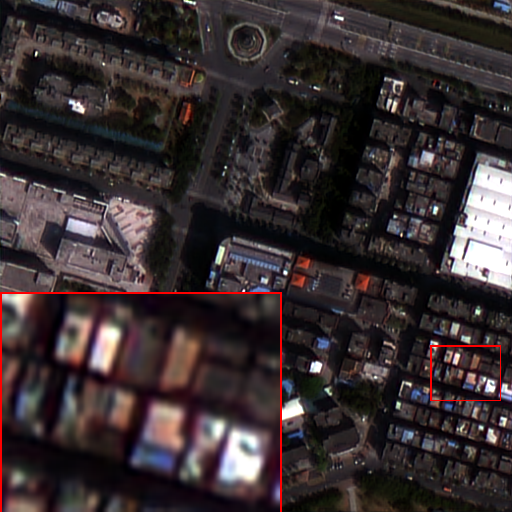} &
			\includegraphics[width=\mywfr]{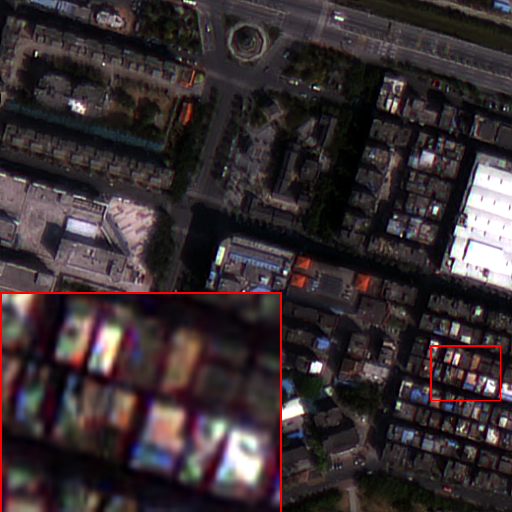} &
			\includegraphics[width=\mywfr]{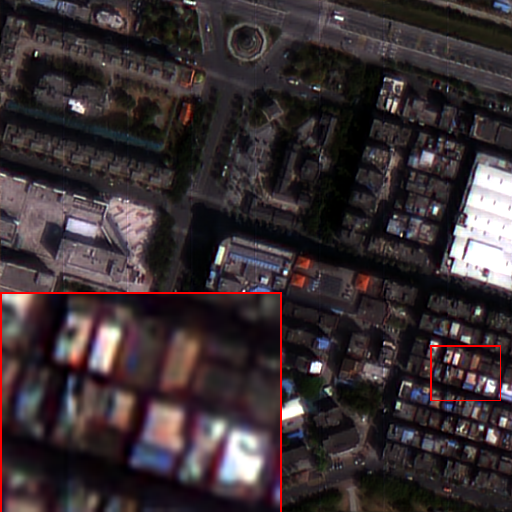} &
			\includegraphics[width=\mywfr]{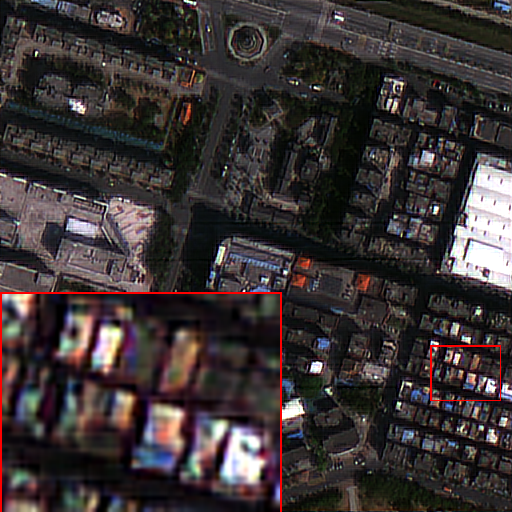} &
			\includegraphics[width=\mywfr]{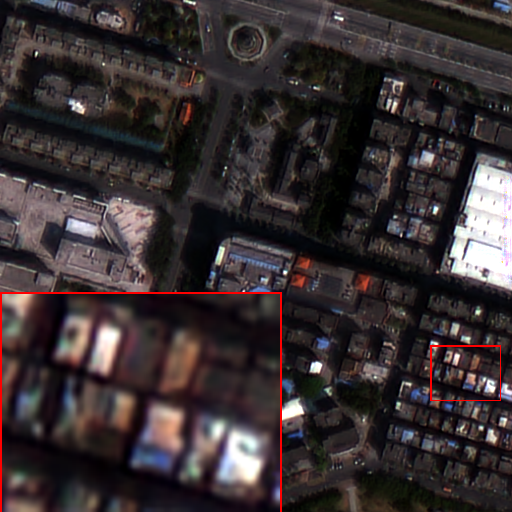} &
			\includegraphics[width=\mywfr]{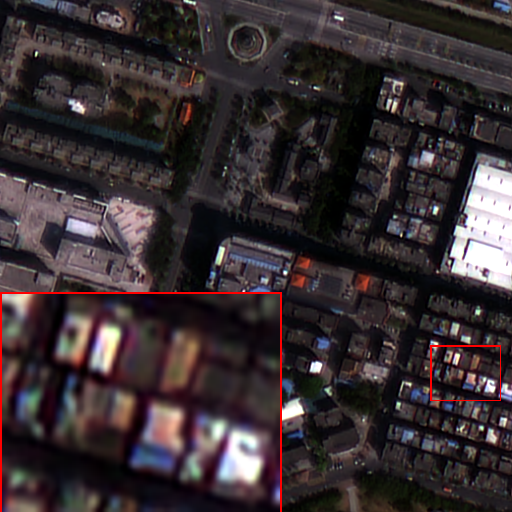} & \\[-1pt]
			
			\includegraphics[width=\mywfr]{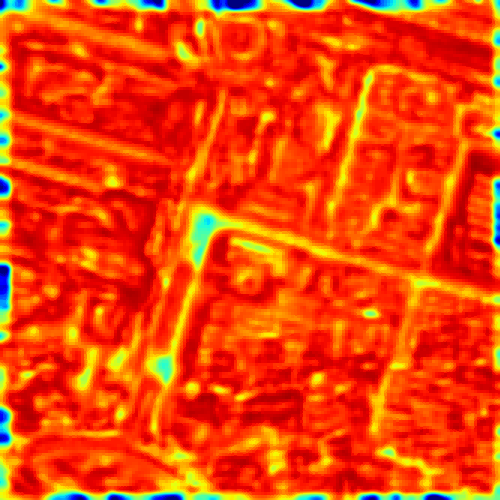} &
			\includegraphics[width=\mywfr]{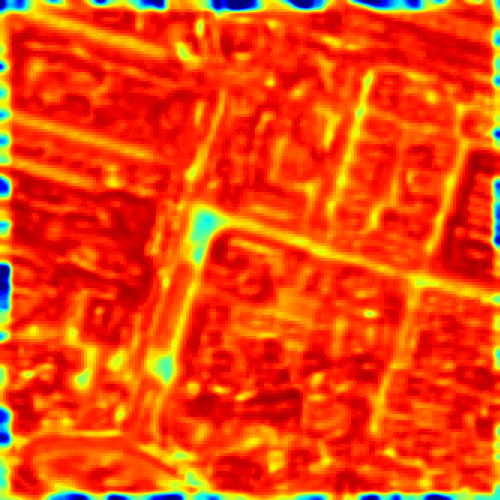} &
			\includegraphics[width=\mywfr]{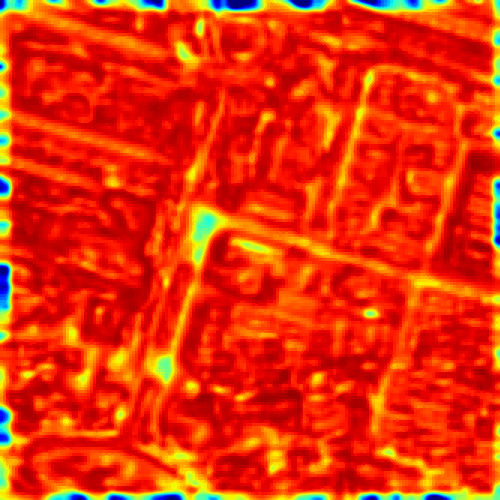} &
			\includegraphics[width=\mywfr]{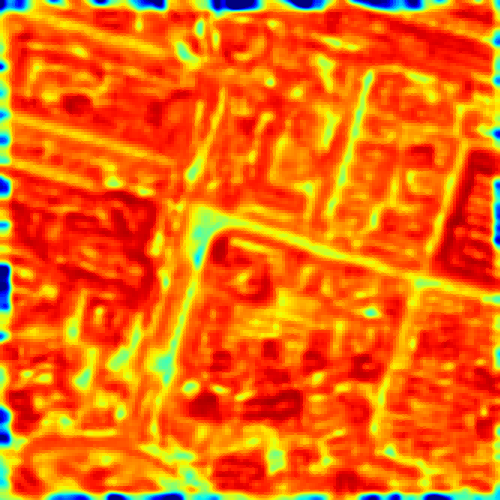} &
			\includegraphics[width=\mywfr]{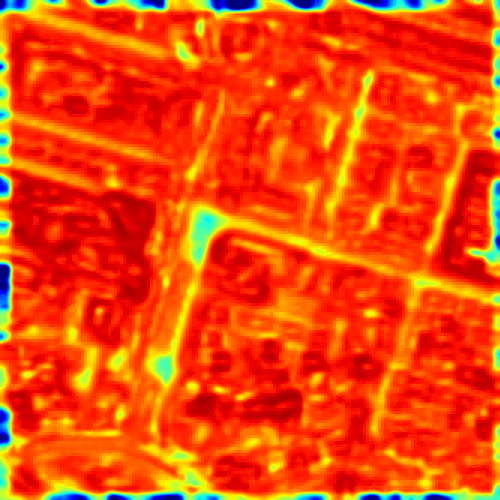} &
			\includegraphics[width=\mywfr]{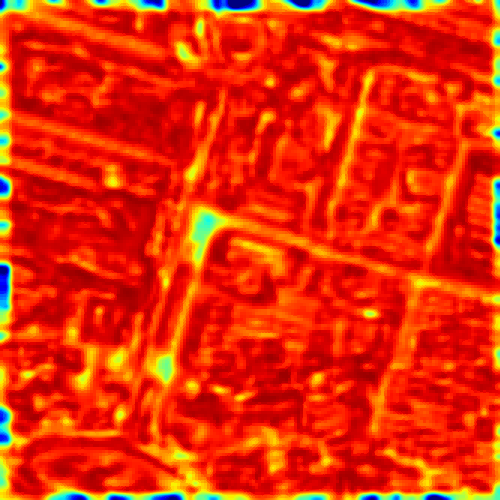} & \\[-2pt]
			
			\small LAGConv & \small PSGAN & \small MSAN & \small PanMamba & \small MARNet & \small GSPan & \\[-6pt]
		\end{tabular}
		
		\caption{Qualitative comparison on the GF2 dataset under full-resolution testing. The top and third rows show the fused images, while the second and fourth rows display the HQNR maps. }
		\label{fig:gf2_fr_comparison}
	\end{figure*}

\begin{figure*}[htbp]
			\centering
			
			\captionsetup[subfloat]{labelformat=empty, skip=0pt}
			\setlength{\tabcolsep}{1pt}

			\newlength{\mywfrwv}
			\setlength{\mywfrwv}{0.145\textwidth} 
			
			\begin{tabular}{ccccccl} 
				
				\includegraphics[width=\mywfrwv]{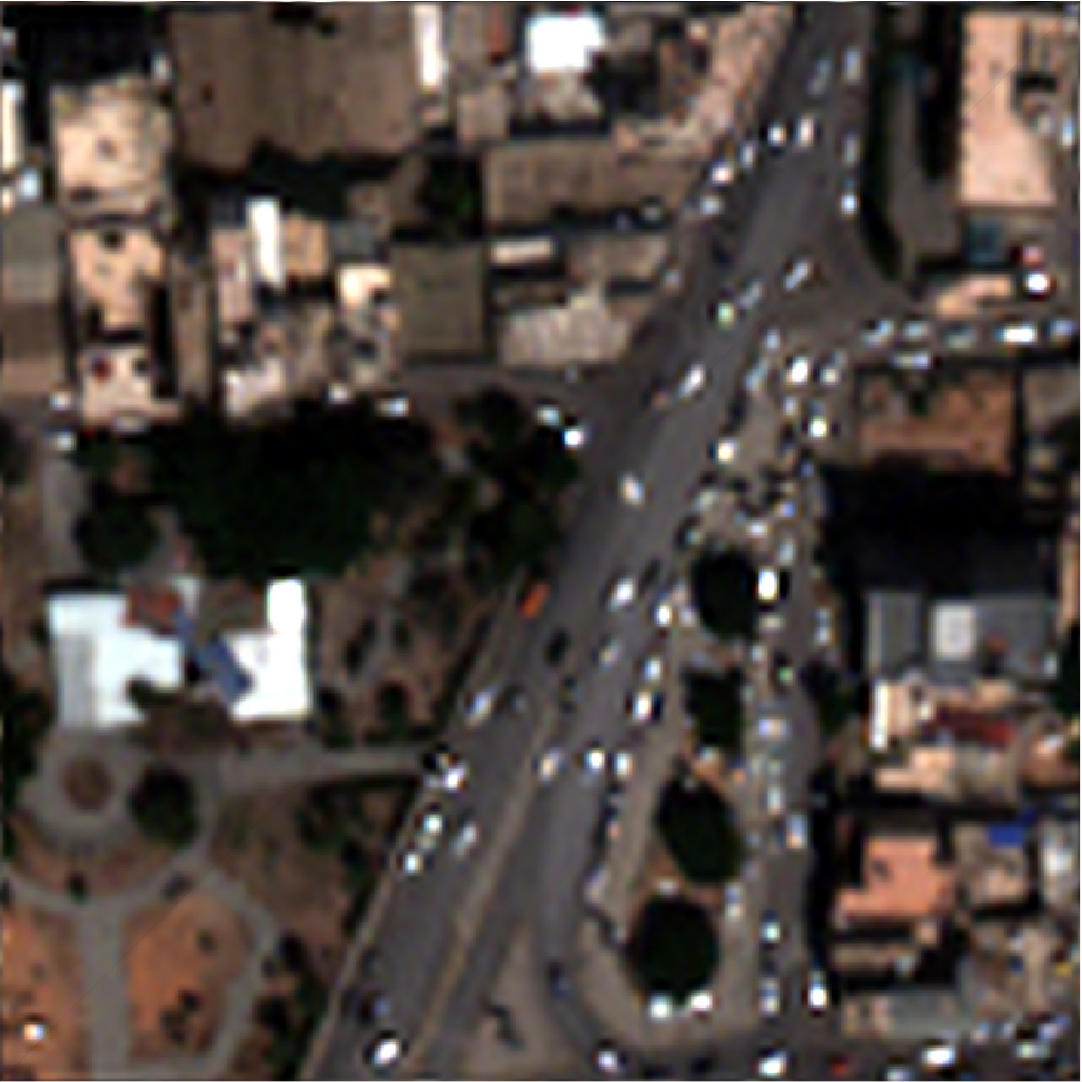} &
				\includegraphics[width=\mywfrwv]{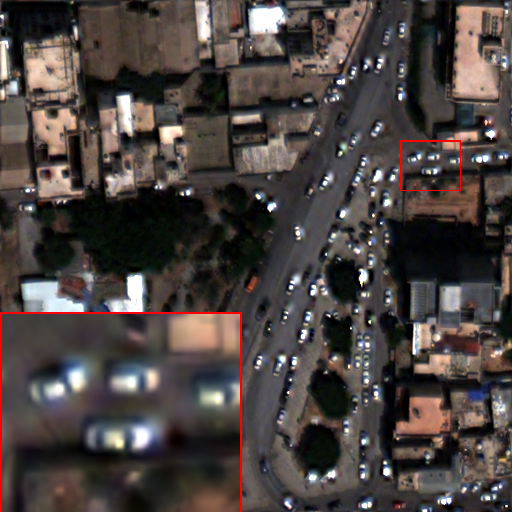} &
				\includegraphics[width=\mywfrwv]{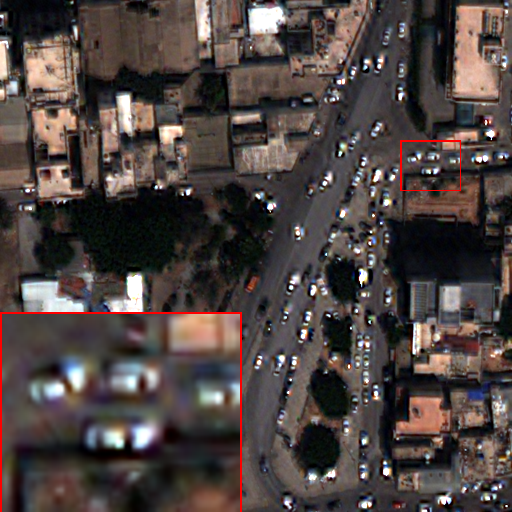} &
				\includegraphics[width=\mywfrwv]{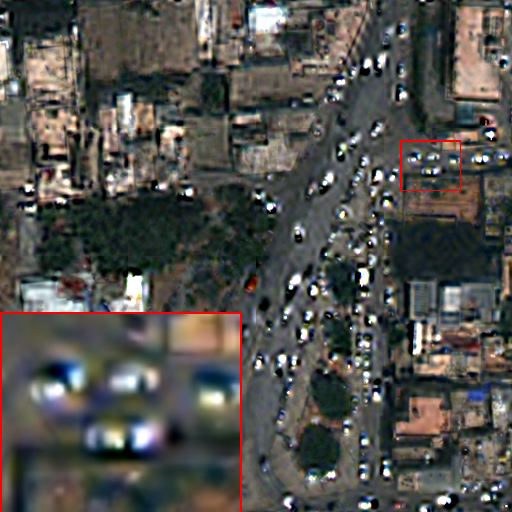} &
				\includegraphics[width=\mywfrwv]{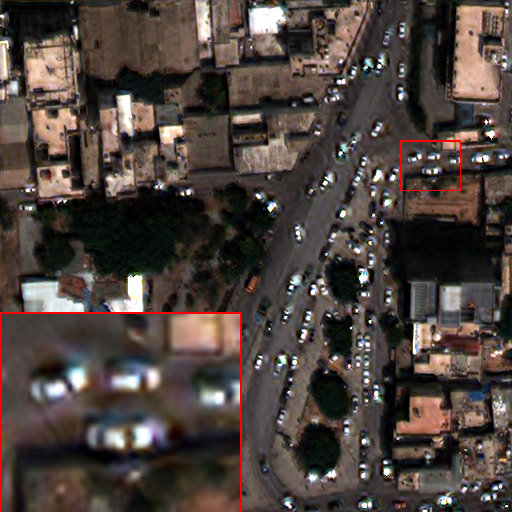} &
				\includegraphics[width=\mywfrwv]{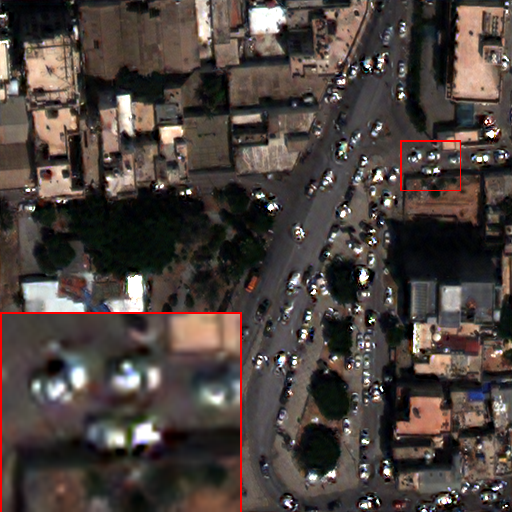} & 
				
				\multirow{6}{*}[\mywfrwv+1pt]{\includegraphics[height=\dimexpr 4\mywfrwv + 36pt\relax, trim=8 0 0 0, clip]{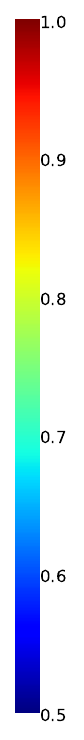}} \\[-1pt]

				\includegraphics[width=\mywfrwv]{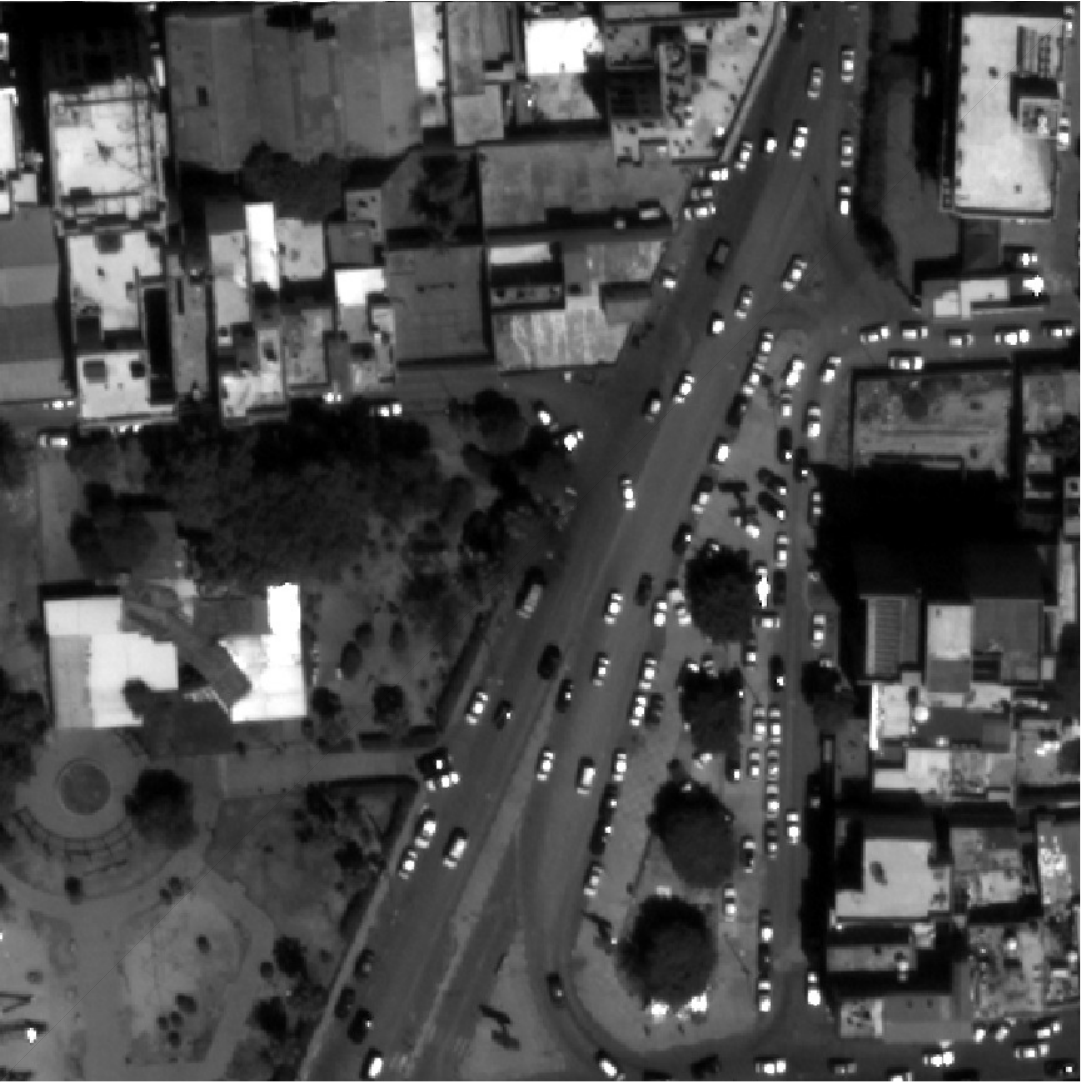} &
				\includegraphics[width=\mywfrwv]{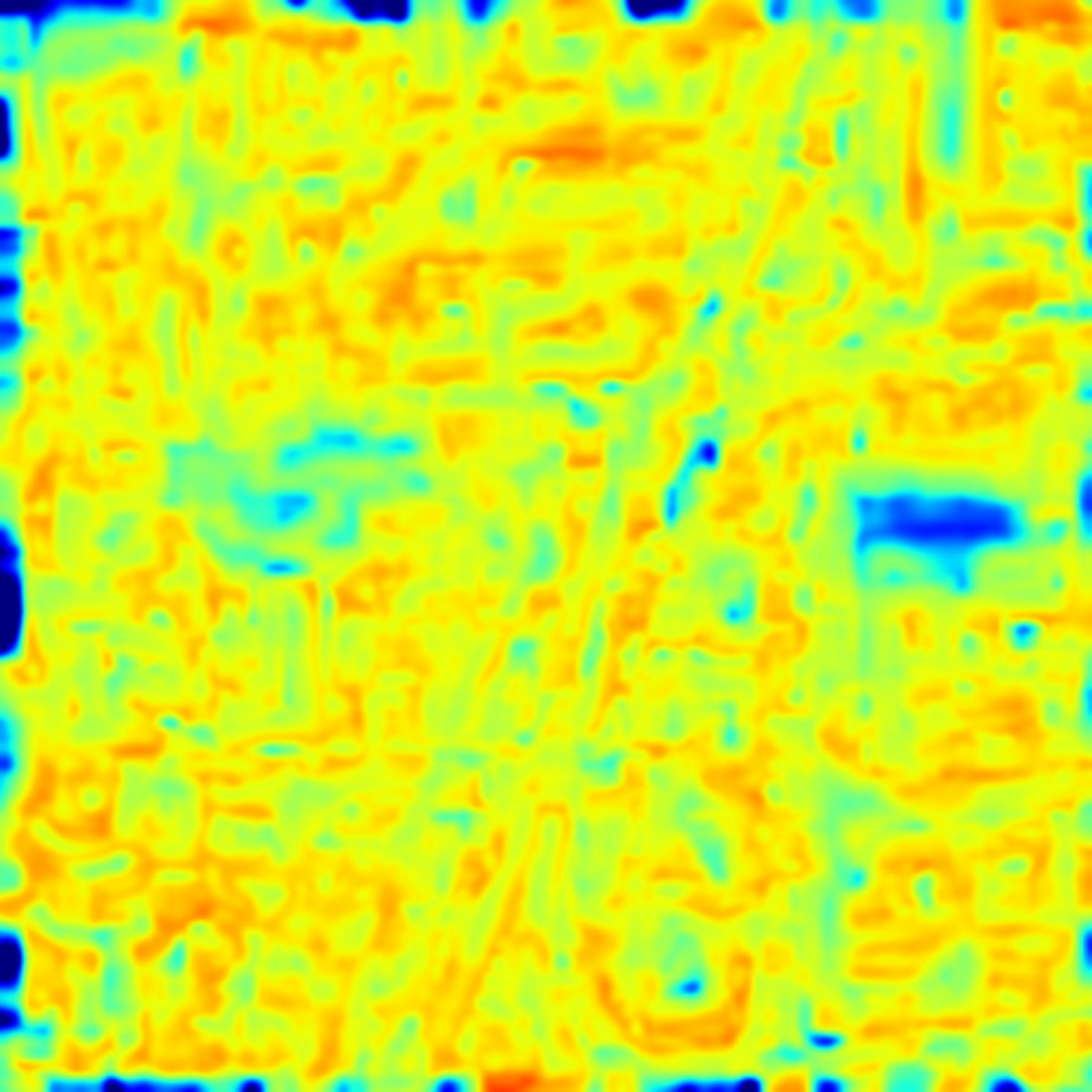} &
				\includegraphics[width=\mywfrwv]{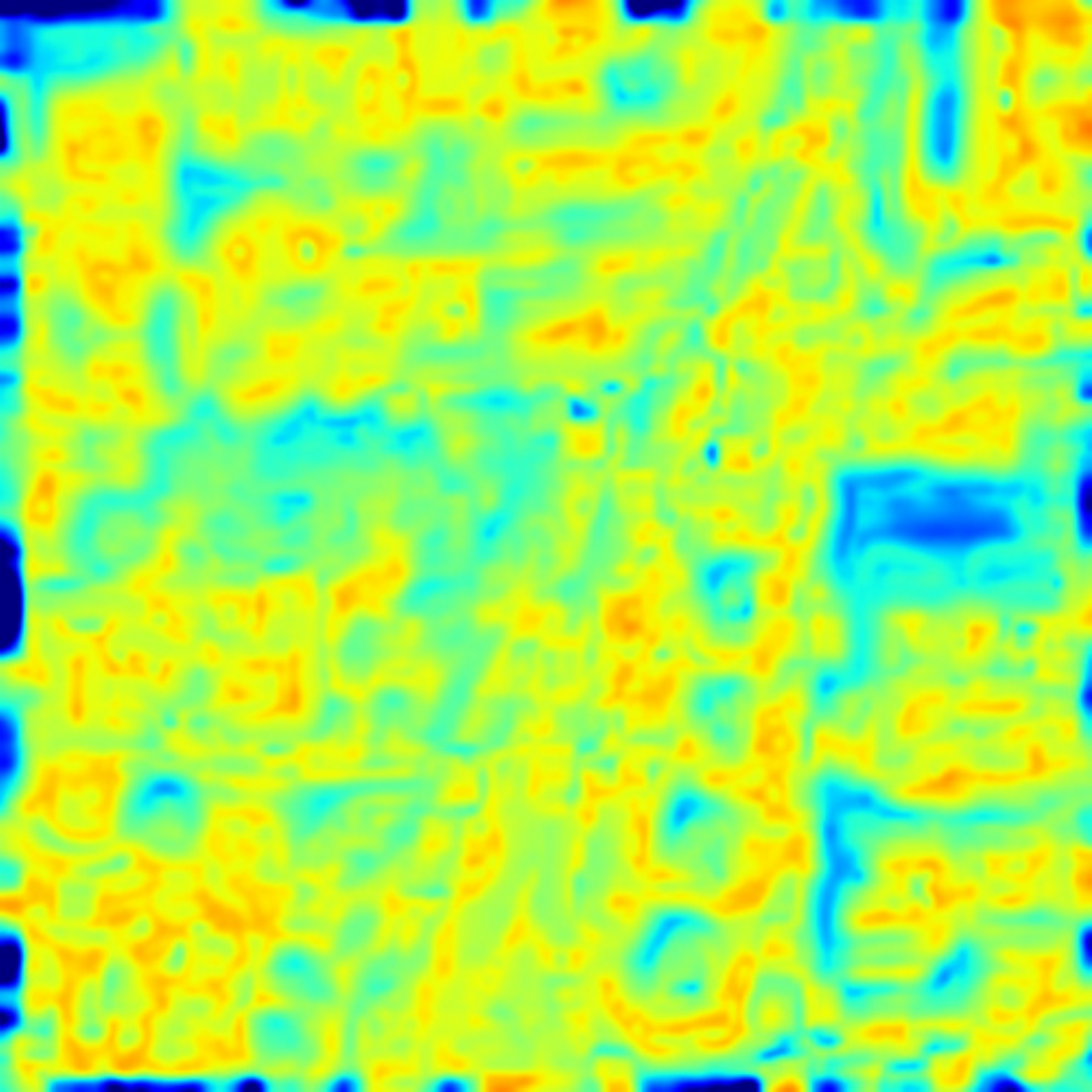} &
				\includegraphics[width=\mywfrwv]{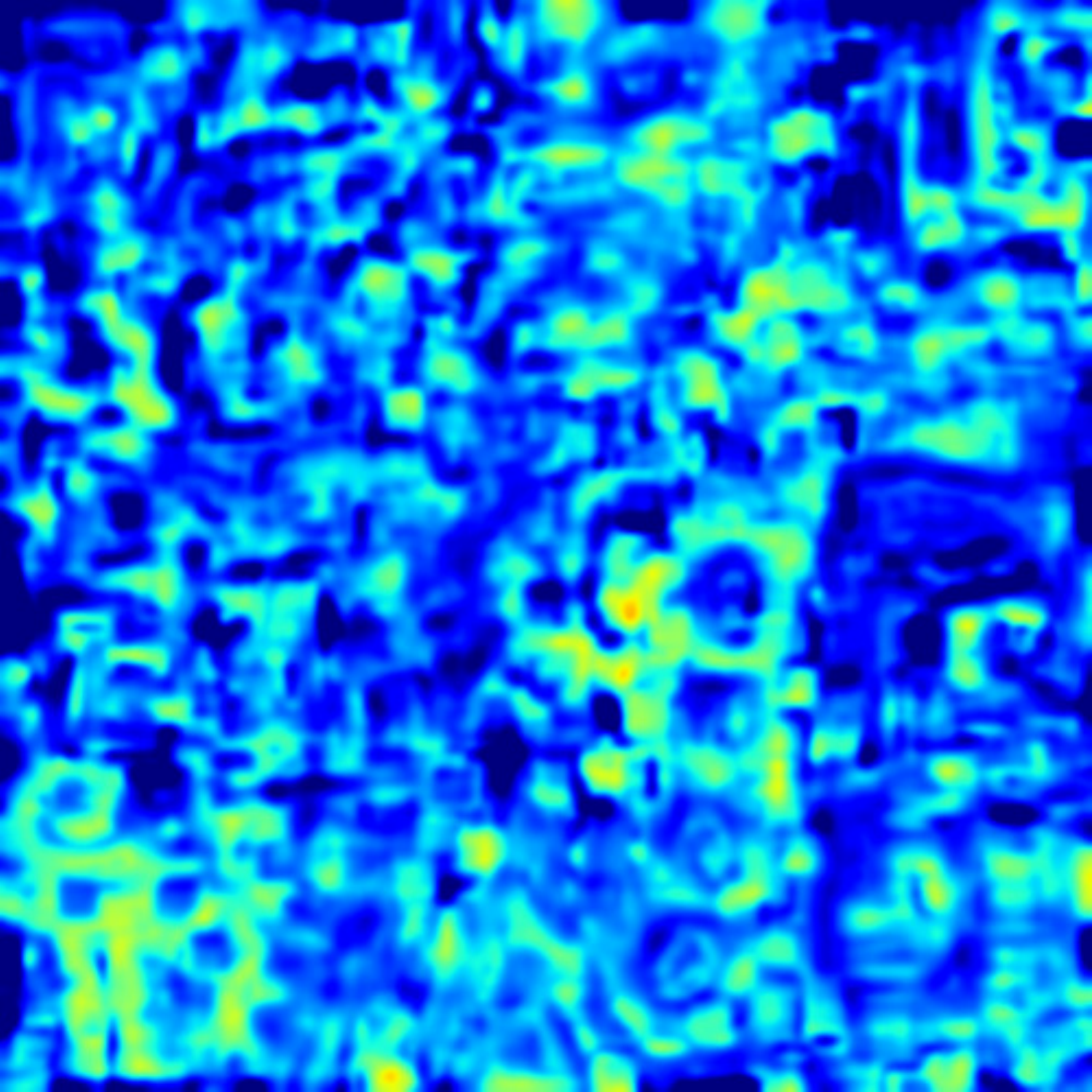} &
				\includegraphics[width=\mywfrwv]{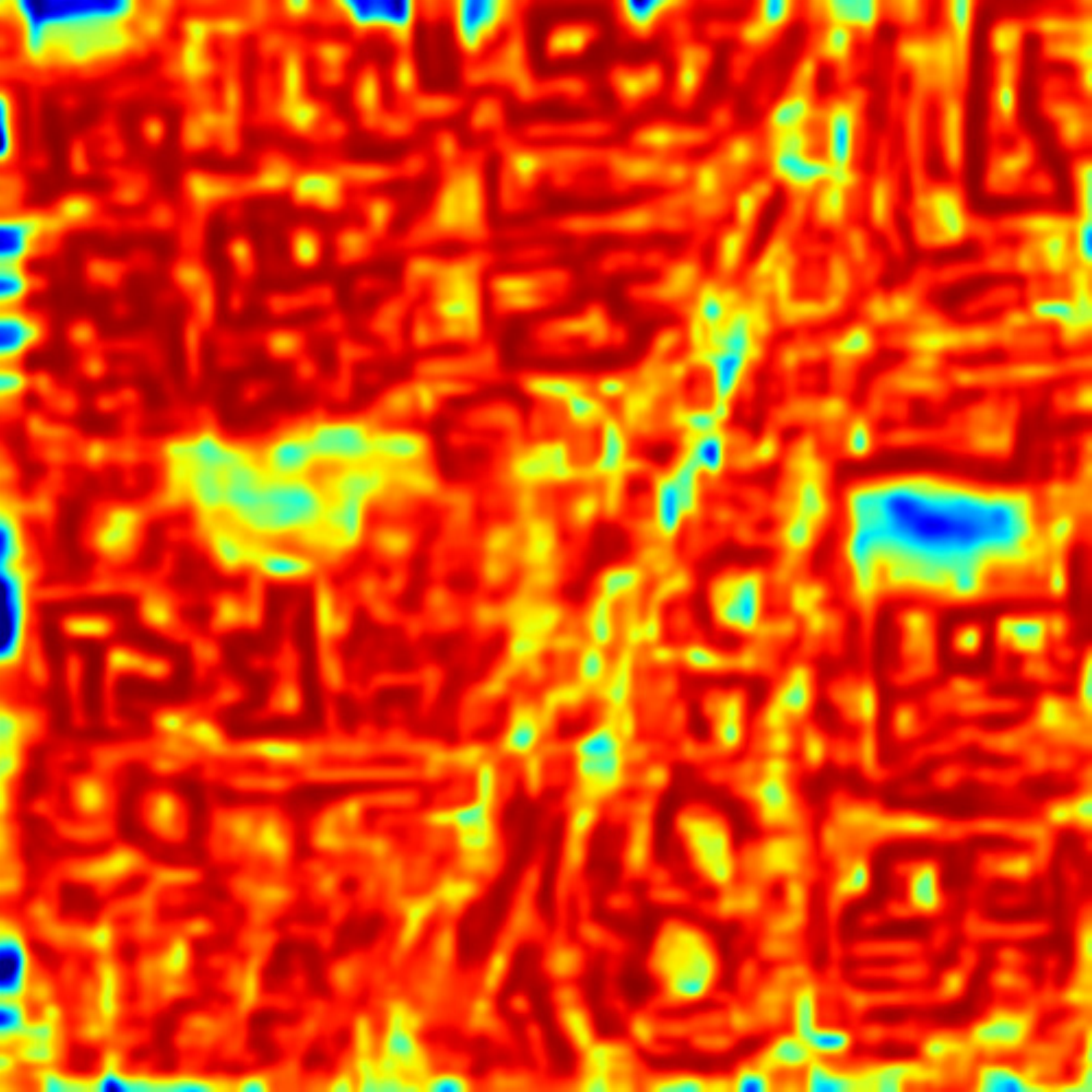} &
				\includegraphics[width=\mywfrwv]{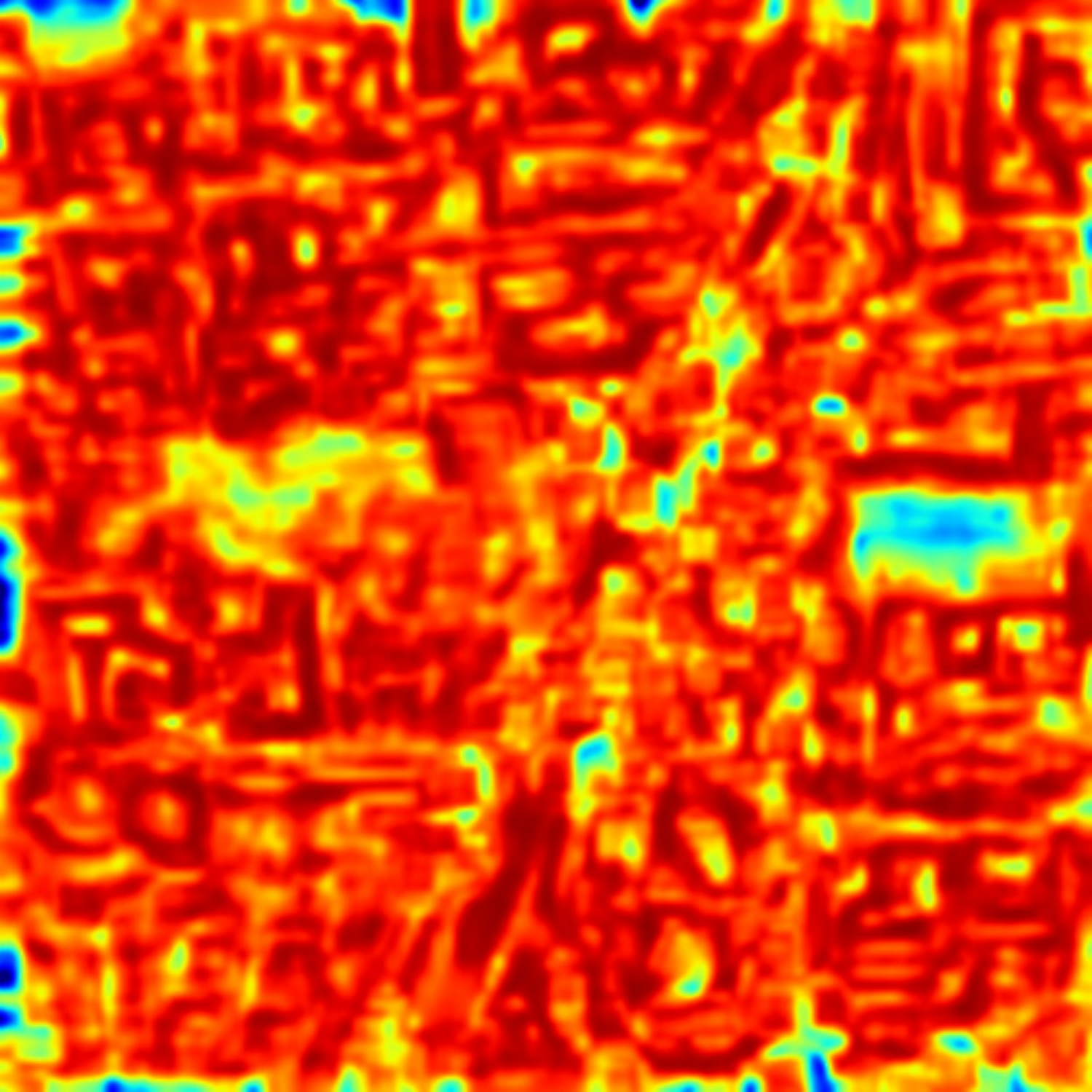} & \\[-2pt]
				
				\small $\widetilde{MS}$/PAN & \small AWLP & \small TV & \small BDSD & \small PNN & \small FusionNet & \\[-2pt]

				\includegraphics[width=\mywfrwv]{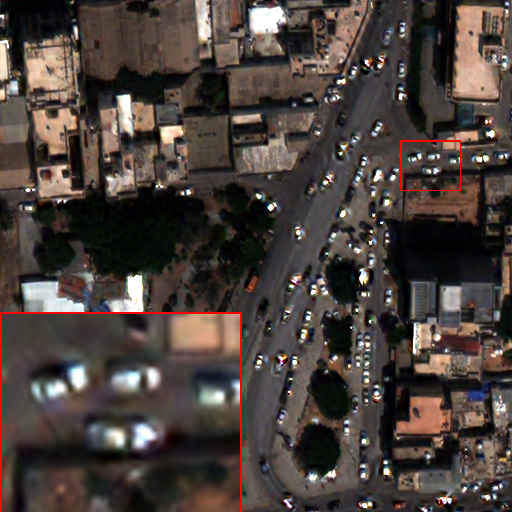} &
				\includegraphics[width=\mywfrwv]{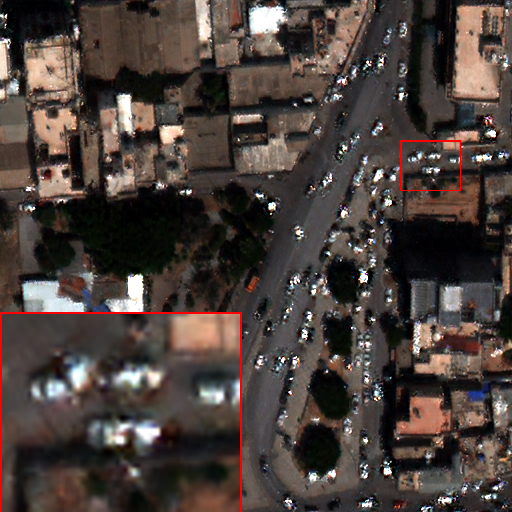} &
				\includegraphics[width=\mywfrwv]{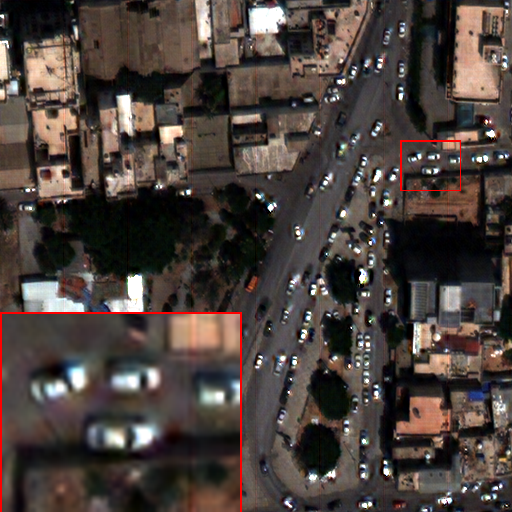} &
				\includegraphics[width=\mywfrwv]{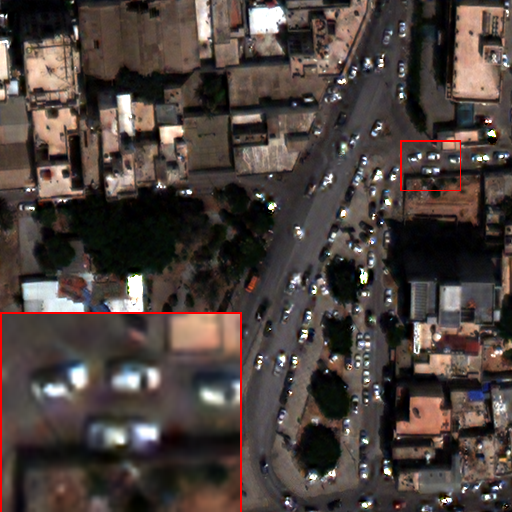} &
				\includegraphics[width=\mywfrwv]{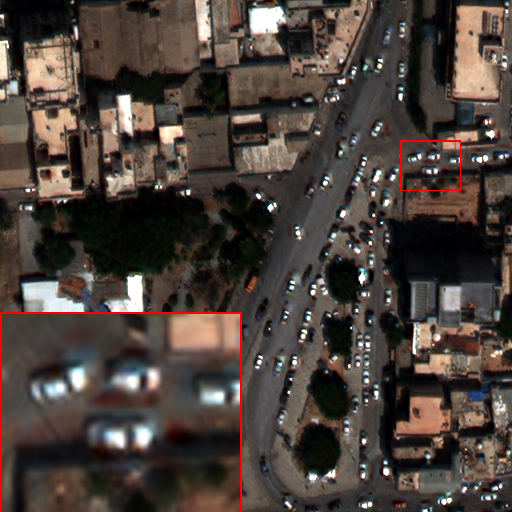} &
				\includegraphics[width=\mywfrwv]{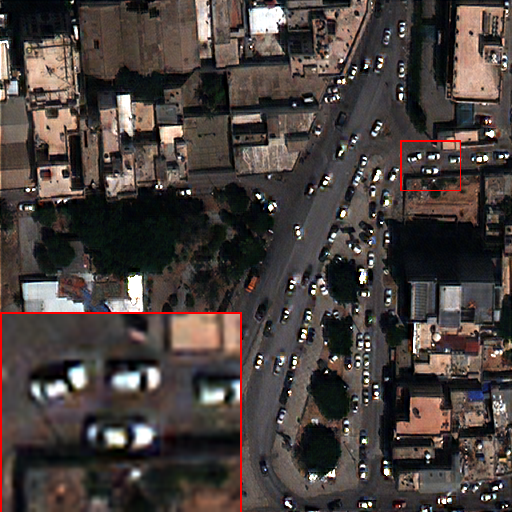} & \\[-1pt]
				
				\includegraphics[width=\mywfrwv]{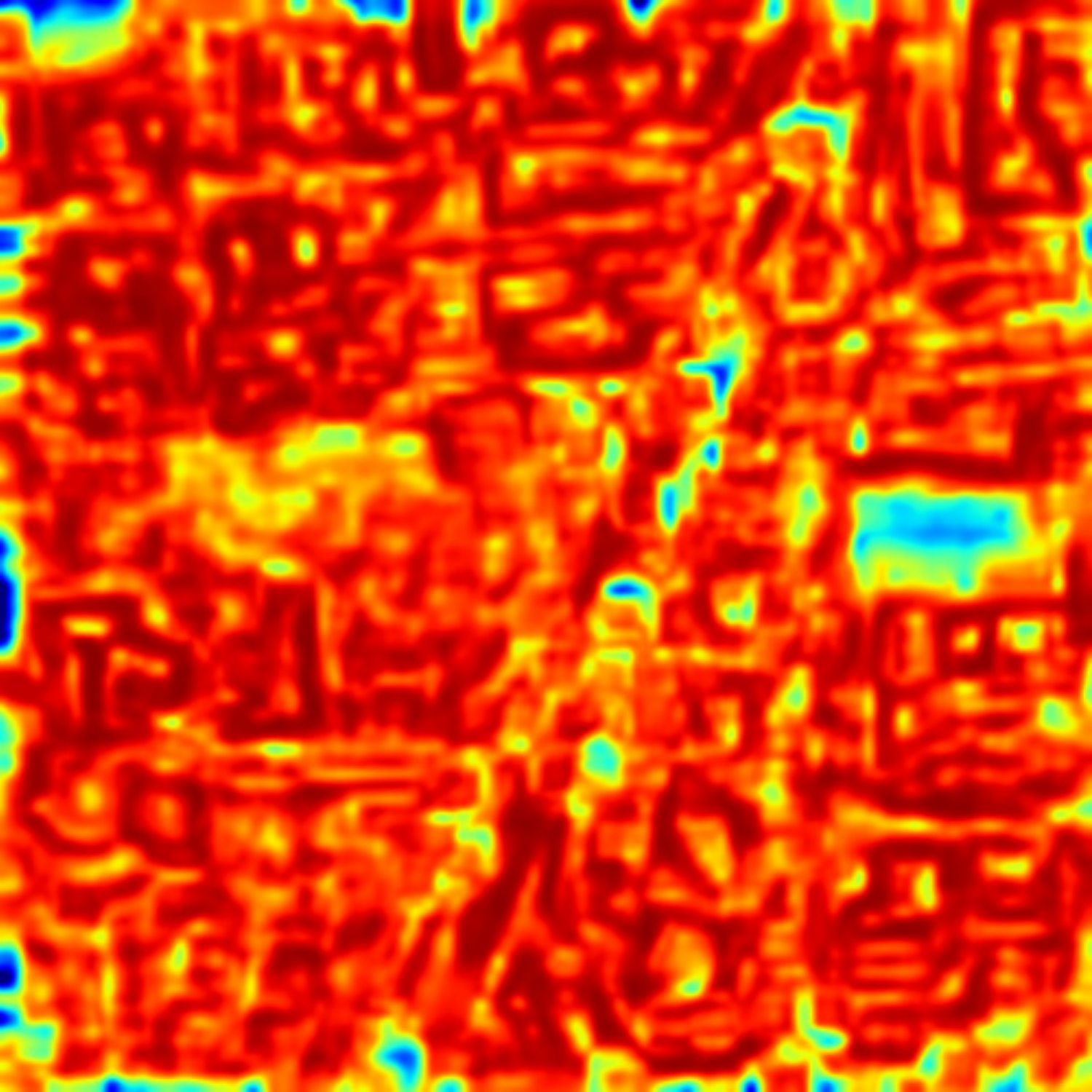} &
				\includegraphics[width=\mywfrwv]{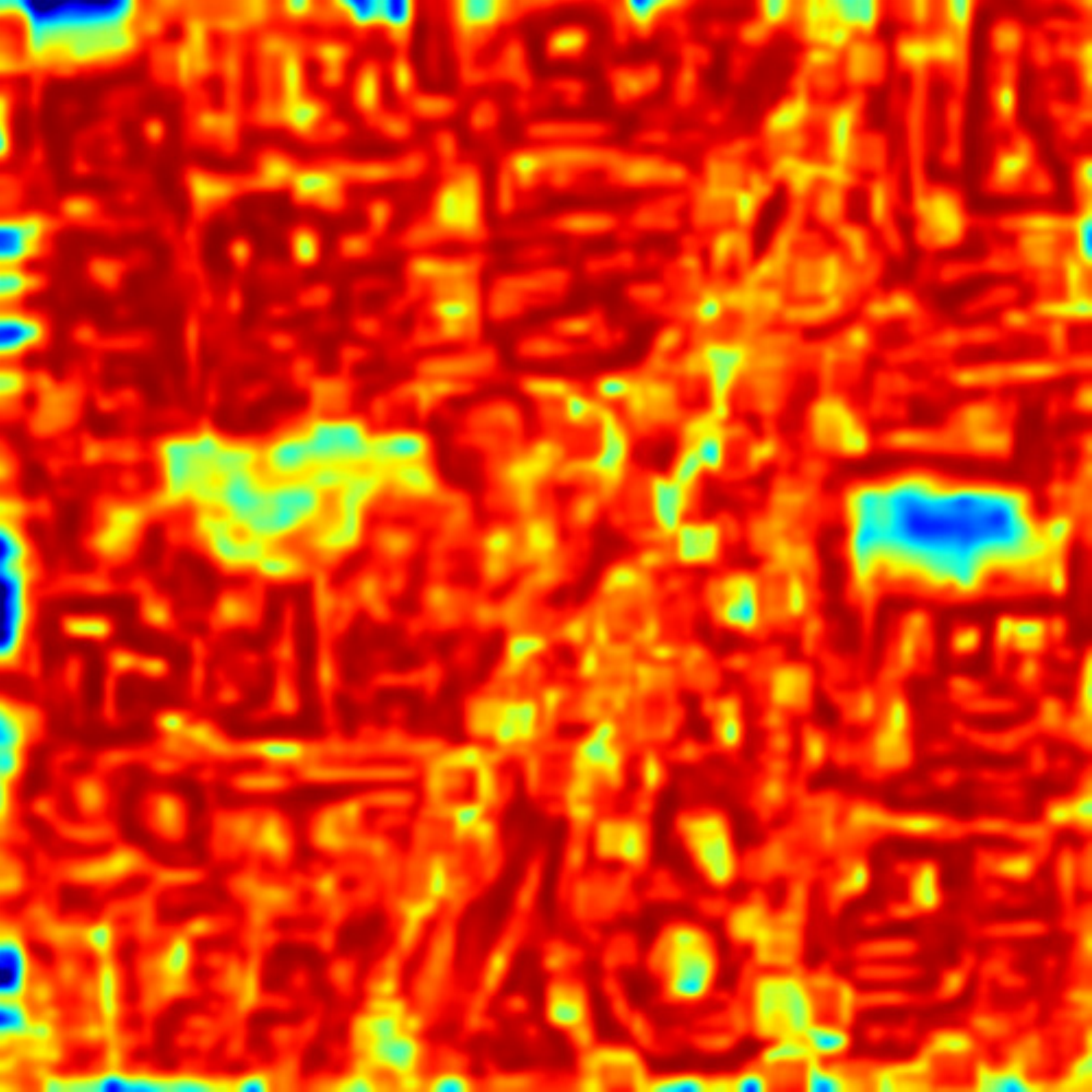} &
				\includegraphics[width=\mywfrwv]{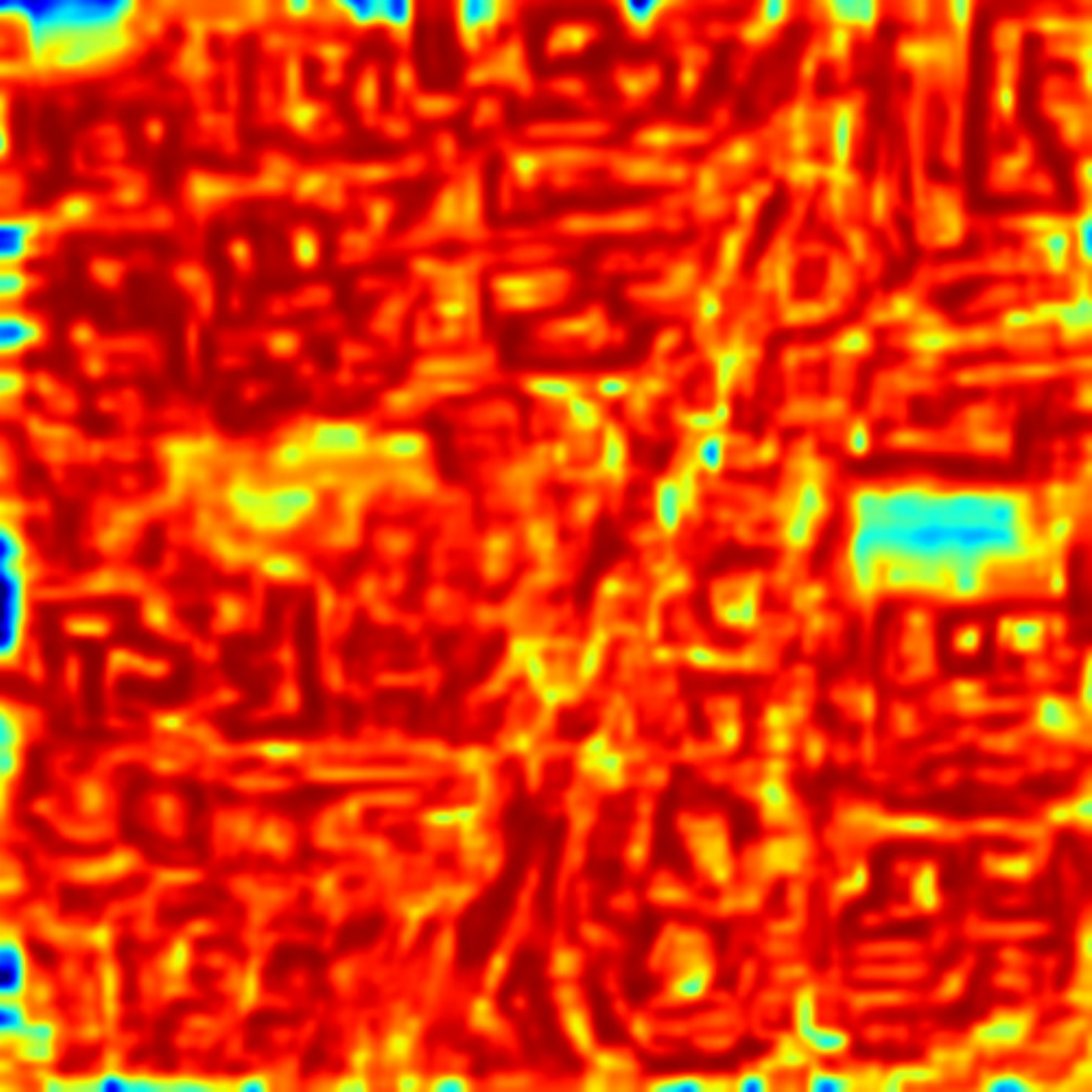} &
				\includegraphics[width=\mywfrwv]{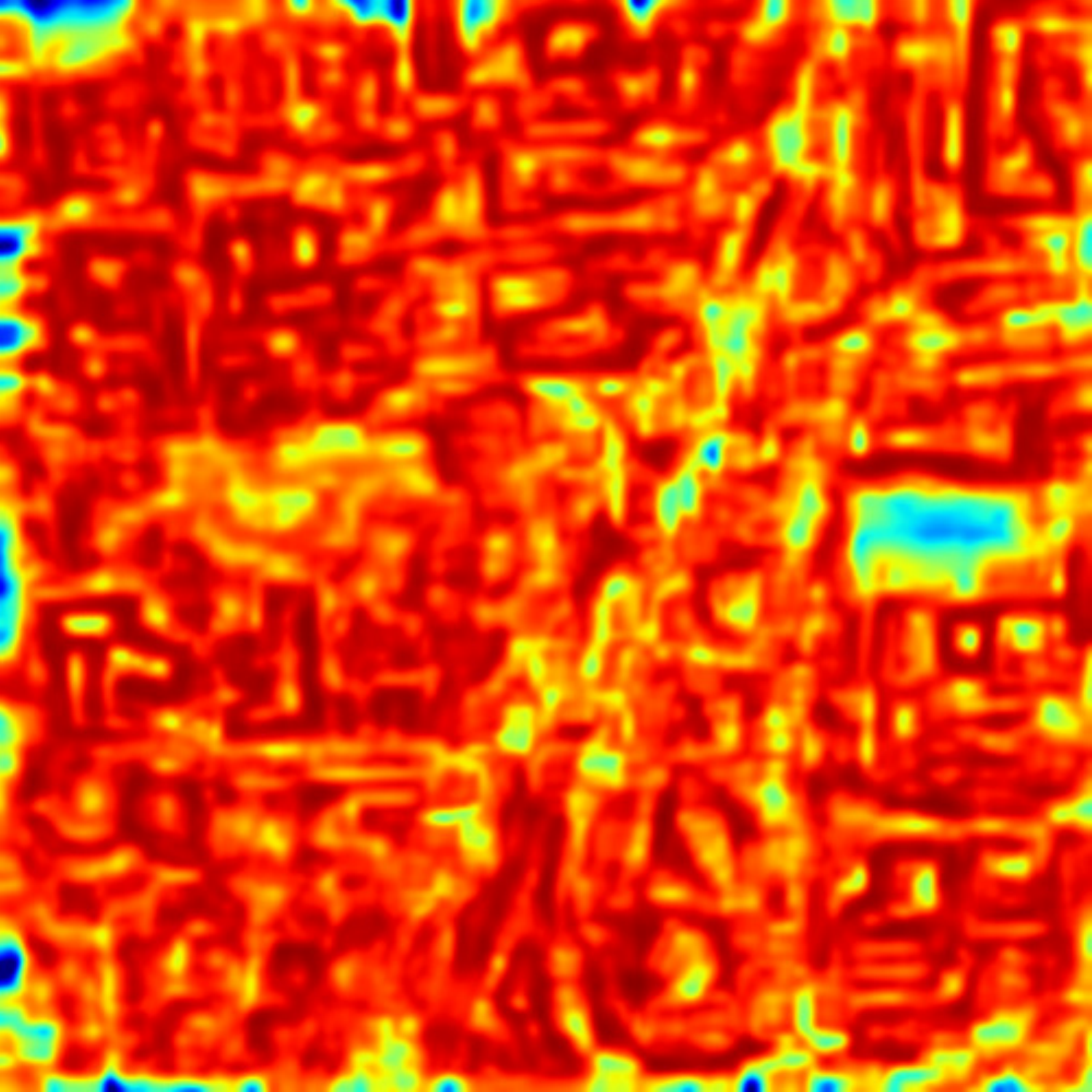} &
				\includegraphics[width=\mywfrwv]{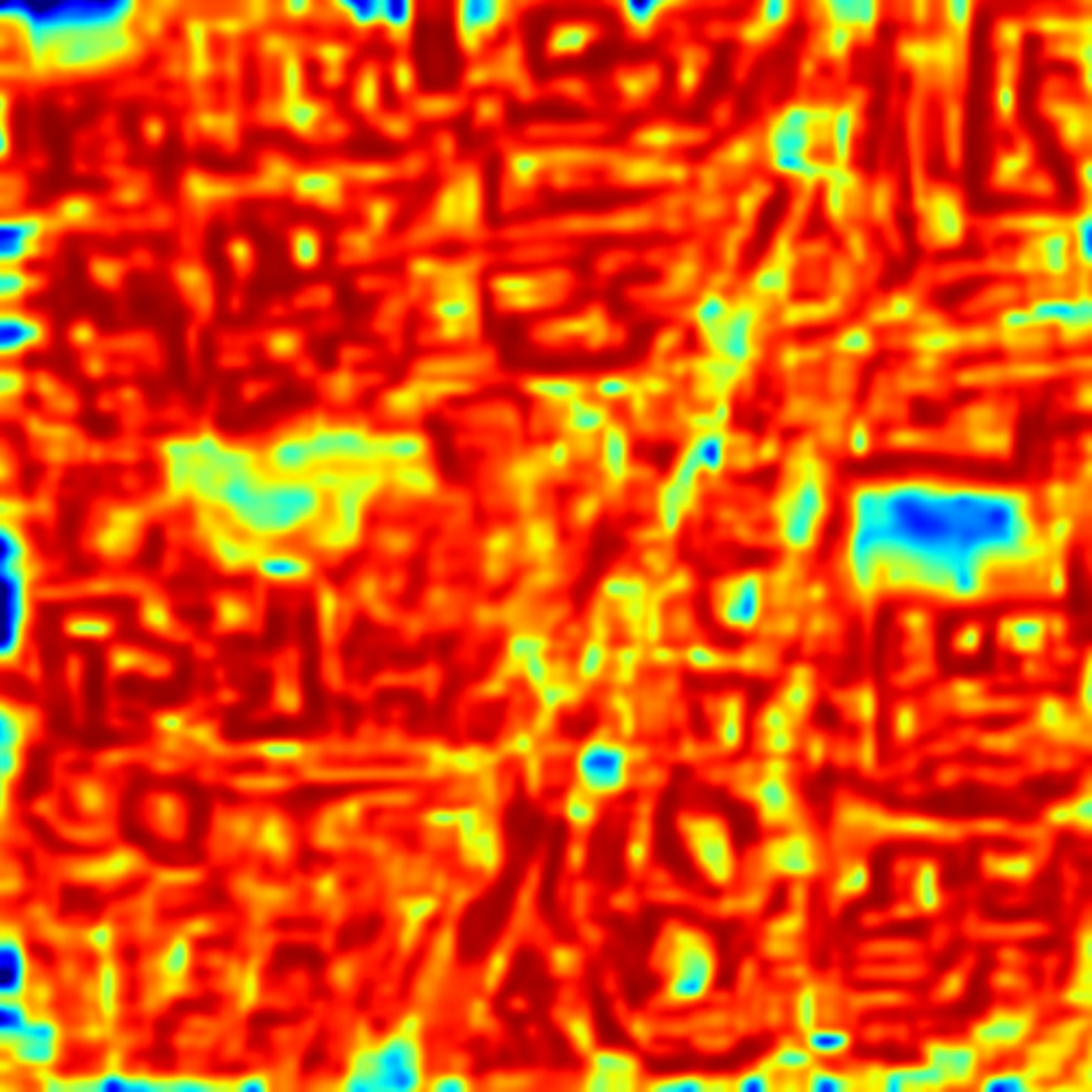} &
				\includegraphics[width=\mywfrwv]{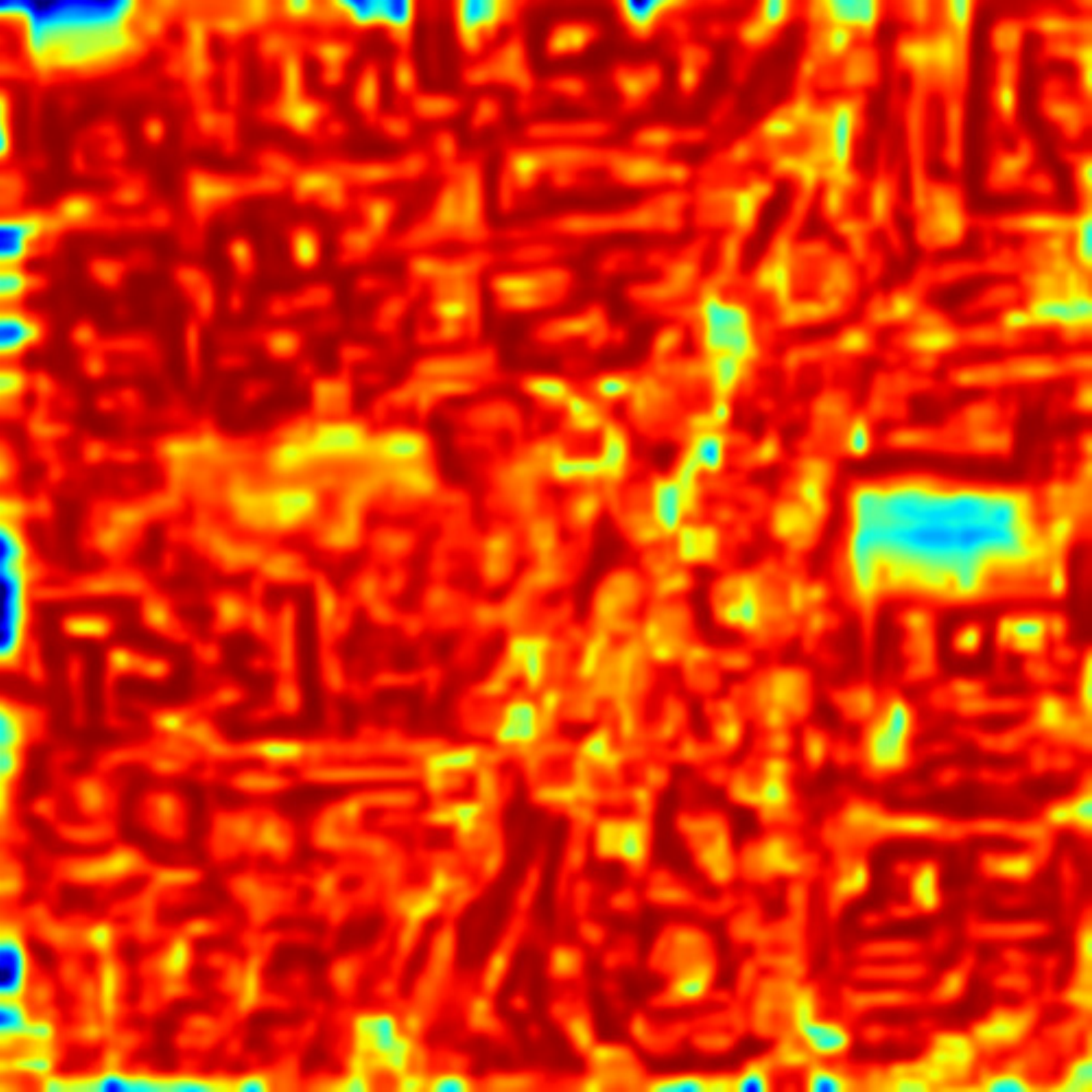} & \\[-2pt]
				
				\small LAGConv & \small PSGAN & \small MSAN & \small PanMamba & \small MARNet & \small GSPan& \\[-6pt]
			\end{tabular}
			
			\caption{Qualitative comparison on the WV3 dataset under full-resolution testing. The top and third rows show the fused images, while the second and fourth rows display the HQNR maps.}
			\label{fig:wv3_fr_comparison}
		\end{figure*}

		Fig.~\ref{fig:qb_fr_comparison}, Fig.~\ref{fig:gf2_fr_comparison}, and Fig.~\ref{fig:wv3_fr_comparison} present the visual fusion results and the corresponding HQNR index maps for the QB, GF2, and WV3 datasets. The HQNR maps provide a localized quality assessment, where colors closer to dark red (value 1.0) represent higher fidelity, while blue and cyan tones signify significant distortion. Across all datasets, the HQNR maps of GSPan are dominated by deep red tones, indicating that our method preserves local spectral and spatial information more effectively than its competitors. Visually, the fused images produced by GSPan exhibit the sharpest textures and the most natural color reproduction, as seen in the building boundaries of the QB dataset and the dense agricultural structures in the GF2 dataset (Fig.~\ref{fig:gf2_fr_comparison}). These visual results, together with the HQNR maps, suggest that GSPan maintains robust full-resolution fusion performance across different sensors.

		\begin{table*}[htbp]
  \centering
  \footnotesize
	\setlength{\tabcolsep}{3pt}
  \caption{Evaluation on the WV3-4K dataset.  $\mathrm{GSPan}^*$ denotes the proposed SDAI mode. Params denotes the number of learnable parameters. Best results are highlighted in red, second-best ones in blue.}
    \begin{tabular}{ccccccccccccc}
    \toprule
    Methods & AWLP  & TV    & BDSD  & PNN   & FusionNet & LAGConv & PSGAN & MSAN  & PanMamba & MARNet & GSPan & $\mathrm{GSPan}^*$ \\
    \midrule
    $D_\lambda$   & 0.0412  & 0.0710  & 0.2880  & 0.117 & 0.1321 & 0.3531 & 0.348 & 0.0732 & 0.2297 & 0.2191 & \textcolor[rgb]{ 0,  0,  1}{0.0527} & \textcolor[rgb]{ 1,  0,  0}{0.0136} \\
     $D_s$ & 0.0918  & 0.1200  & 0.1844  & \textcolor[rgb]{ 0,  0,  1}{0.0379} & 0.1425 & 0.0697 & 0.0691 & 0.0794 & 0.0508 & 0.1963 & \textcolor[rgb]{ 1,  0,  0}{0.0247} & 0.1106 \\
    HQNR  & 0.8708  & 0.8175  & 0.5807  & 0.8496 & 0.7445 & 0.6018 & 0.607 & 0.8532 & 0.7317 & 0.6291 & \textcolor[rgb]{ 1,  0,  0}{0.9239} & \textcolor[rgb]{ 0,  0,  1}{0.8772} \\
    Inference Time(s) & -     & -     & -     & \textcolor[rgb]{ 0,  0,  1}{2.89} & \textcolor[rgb]{ 1,  0,  0}{2.03} & 16.72 & 3.05  & 9.76  & 20.64 & 81.7  & 76.48 & 6.89 \\
    Params(M) & -     & -     & -     & \textcolor[rgb]{ 0,  0,  1}{0.1044} & \textcolor[rgb]{ 1,  0,  0}{0.0786} & 0.1514 & 2.2515 & 2.6236 & 0.4744 & 1.7873 & 2.6241 & 2.6241 \\
     \bottomrule
    \end{tabular}%
  \label{tab:wv3_4k_fr}%
\end{table*}%
		\subsubsection{Evaluation on Large-Scene Full-Resolution Inference}

		Real-world satellite images are usually delivered as large continuous scenes rather than small cropped patches. Therefore, we further evaluate the proposed method on the WV3-4K dataset to examine its large-scene inference capability under $4096 \times 4096$ settings. Although the WV3-4K dataset is captured by the identical WorldView-3 sensor as the WV3 dataset, it is more challenging because large scenes cover a larger spatial extent and contain more complex heterogeneous land-cover distributions.

		As reported in Table~\ref{tab:wv3_4k_fr}, under full-resolution inference, GSPan achieves the highest HQNR and the lowest spatial distortion $D_s$ among the compared methods. This indicates that the main advantage of GSPan is not an isolated improvement in one metric, but a better overall spatial-spectral fidelity in large-scene inference. 
        
        The qualitative results in Fig.~\ref{fig:wv3_4k_comparison} further support the quantitative observations. Compared with competing methods, GSPan produces clearer structural boundaries and more coherent textures over the large continuous scene. Some conventional and DL-based methods tend to generate over-smoothed regions, local artifacts, or visually inconsistent details when applied to the $4096 \times 4096$ image. In contrast, GSPan preserves sharper spatial structures while maintaining a relatively natural spectral appearance. The corresponding HQNR map also contains broader high-quality regions and fewer low-quality local areas, suggesting that GSPan maintains more stable fusion quality across heterogeneous land-cover regions.
        
        The favorable large-scene performance of GSPan is attributed to its continuous Gaussian primitive representation. Most CNN-, Transformer-, GAN-, and SSM-based methods directly learn discrete pixel-level mappings. When applied to 4K scenes, their learned fusion rules are tightly coupled with the output grid, which may increase the risk of PAN over-injection, local spectral drift, or window-wise inconsistency in large-scene inference. In contrast, GSPan predicts Gaussian primitives and renders the fused image in a continuous coordinate space. Moreover, the residual synthesis formulation uses the upsampled MS image as the spectral base and lets Gaussian primitives model only the residual details field, which helps preserve spectral consistency while restoring sharp spatial structures.

		\begin{figure*}[htbp]
			\centering
			\includegraphics[width=\textwidth]{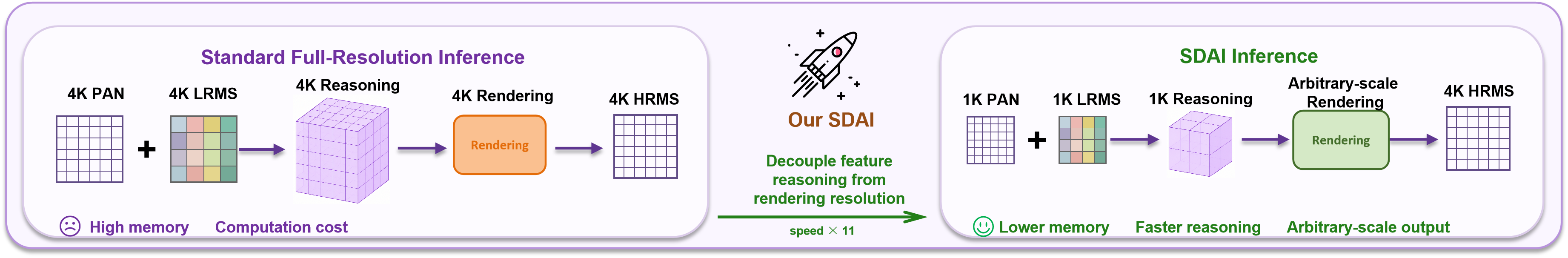}  
			\caption{Illustration of Scale-Decoupled Asymmetric Inference (SDAI). Unlike standard full-resolution inference, which estimates Gaussian primitive attributes and renders the output on the same full-resolution grid, SDAI estimates primitive attributes at a reduced resolution and renders the fused image on the target high-resolution grid. This design reduces large-scene inference cost by decoupling primitive attribute estimation from final-resolution rendering.
			}
			\label{fig:sdia}
		\end{figure*}
		
		\begin{figure*}[htbp]
			\centering
			\captionsetup[subfloat]{labelformat=empty, skip=0pt}
			\setlength{\tabcolsep}{1pt} 
			
			\newlength{\mywkv}
			\setlength{\mywkv}{0.145\textwidth} 
			
			\begin{tabular}{cccccc l} 
				\includegraphics[width=\mywkv]{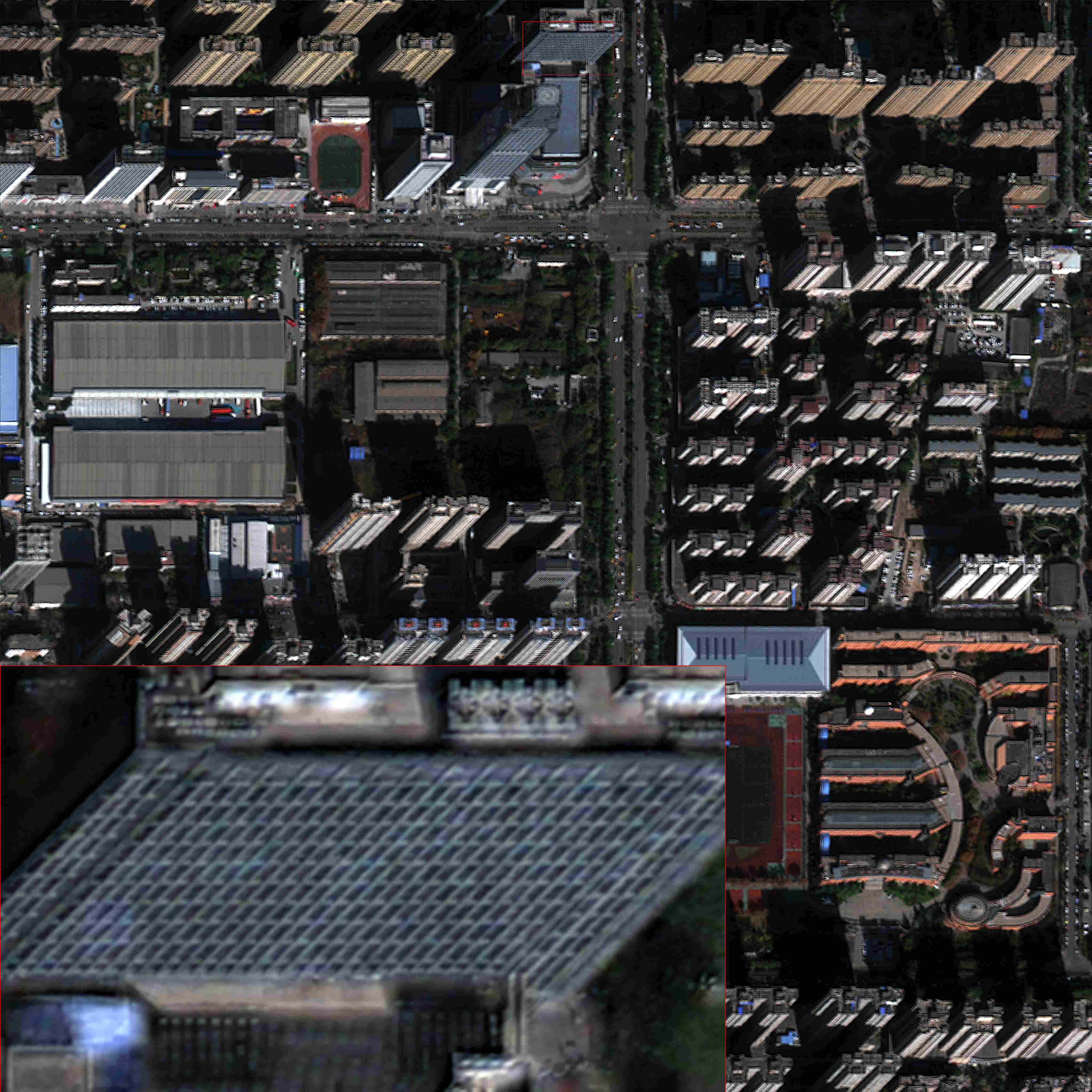} &
				\includegraphics[width=\mywkv]{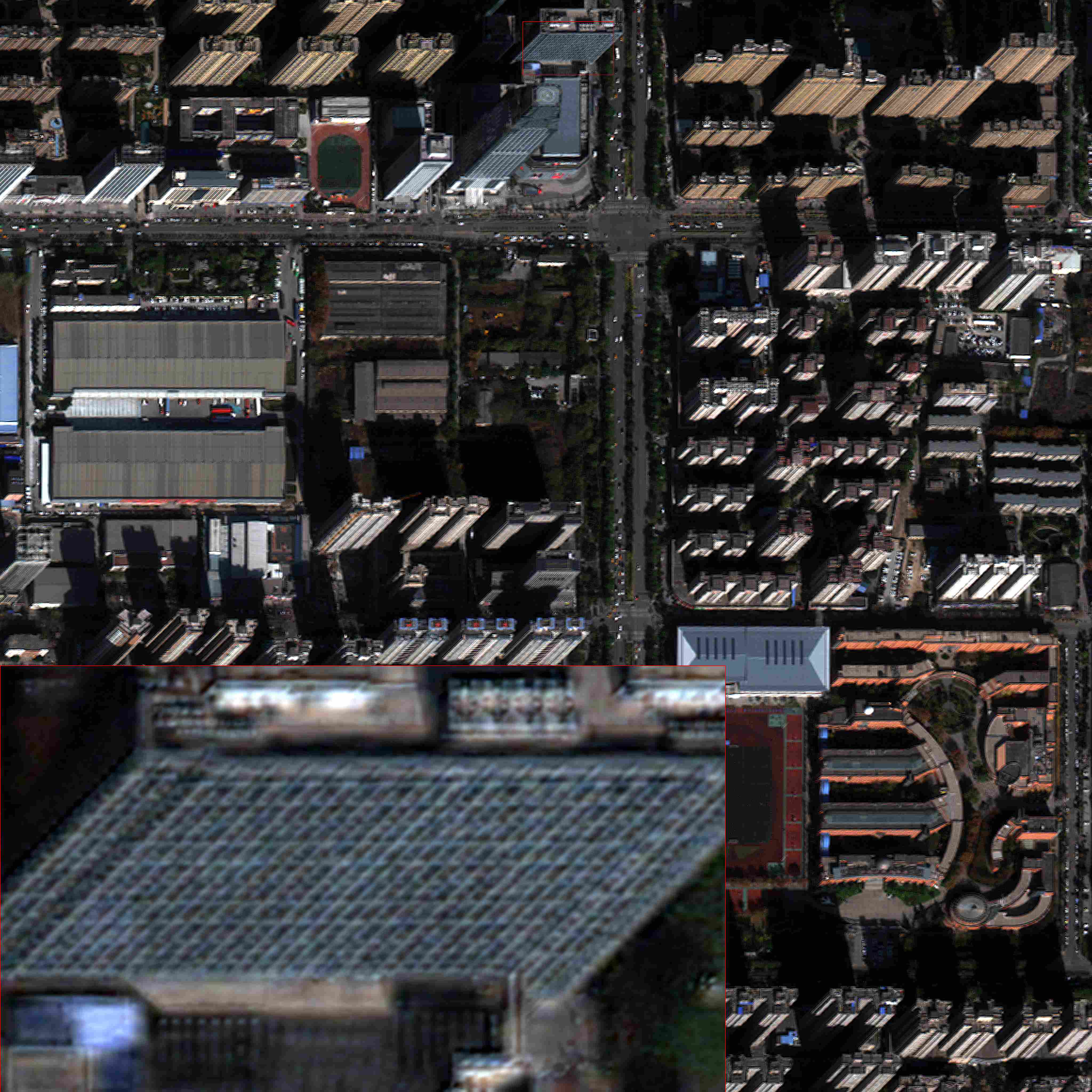} &
				\includegraphics[width=\mywkv]{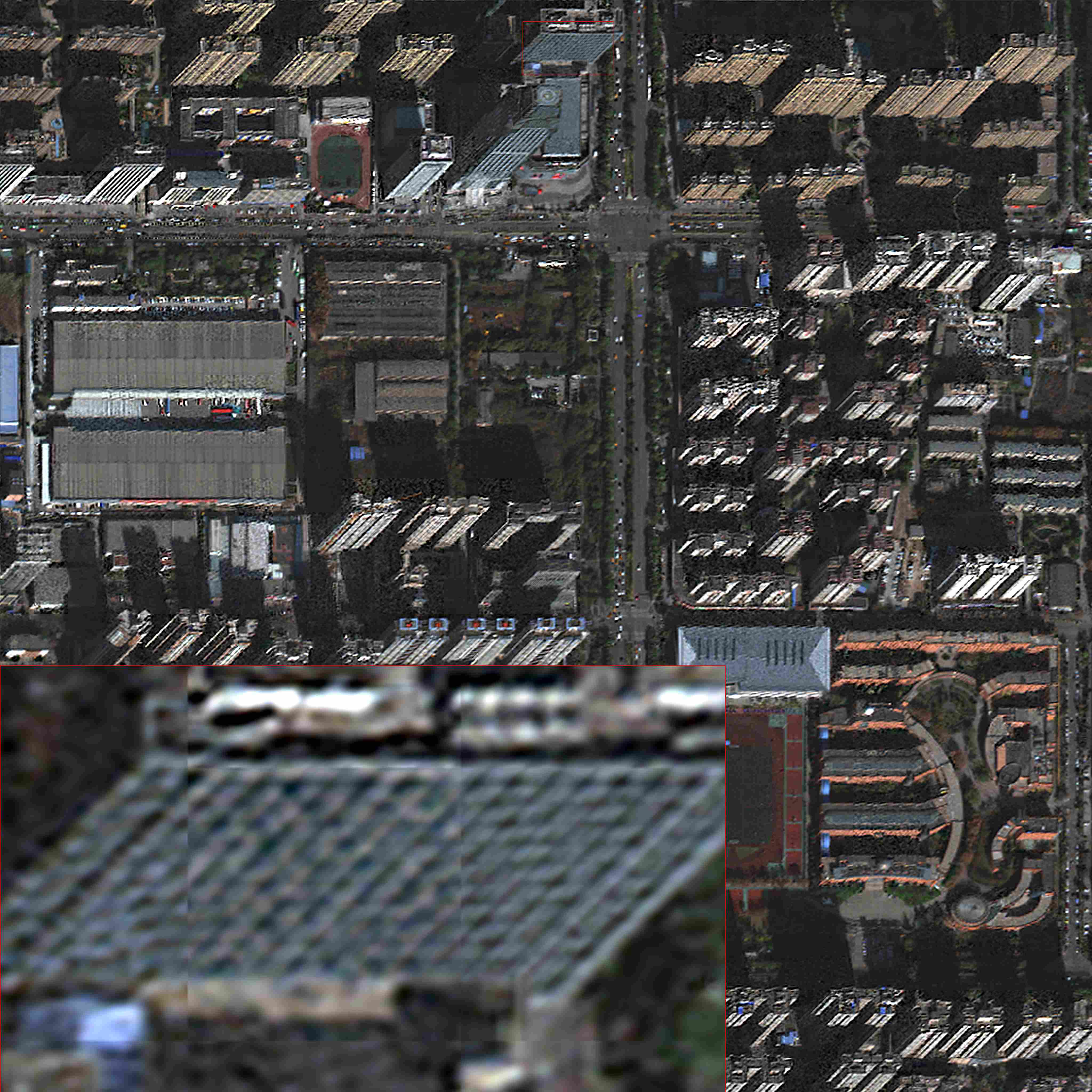} &
				\includegraphics[width=\mywkv]{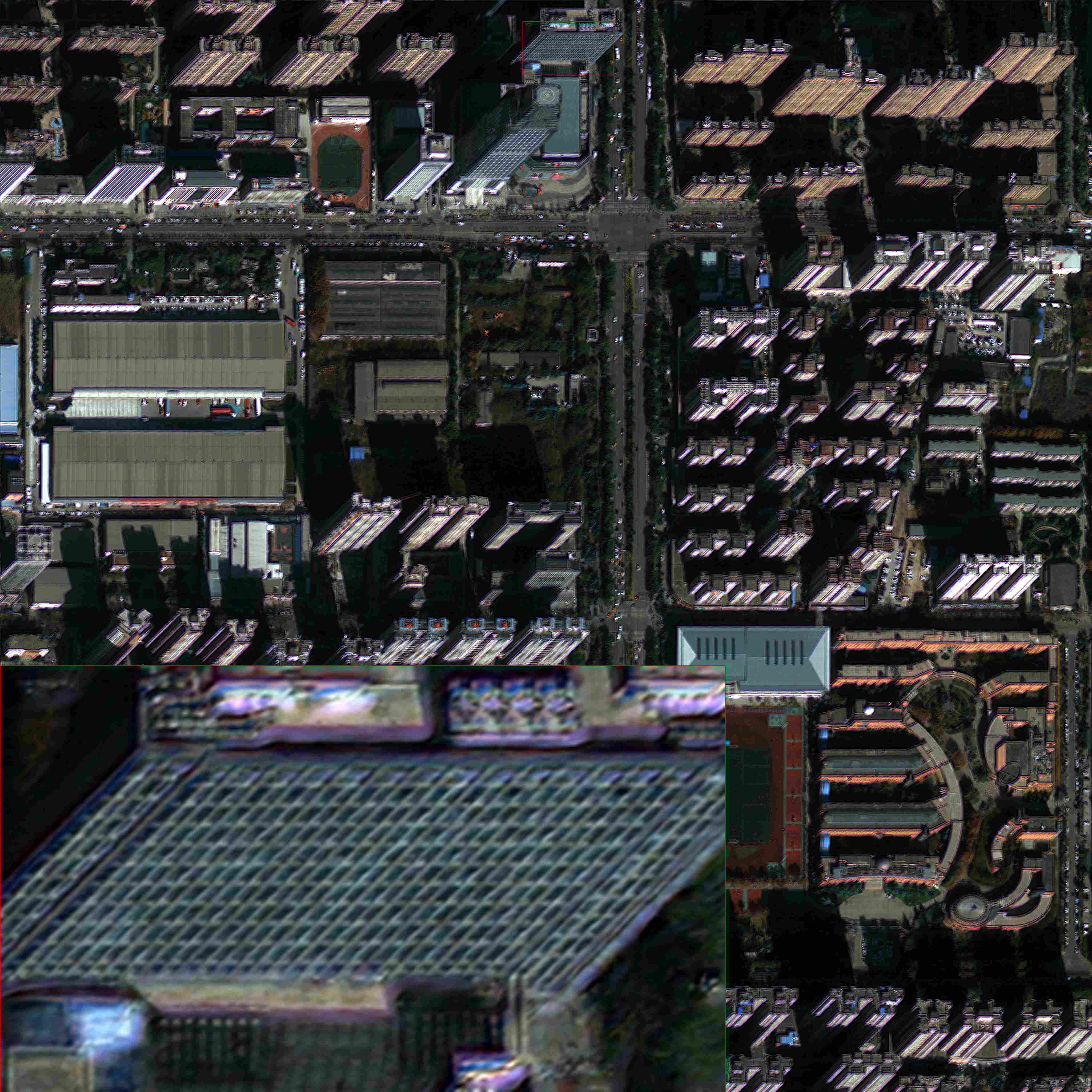} &
				\includegraphics[width=\mywkv]{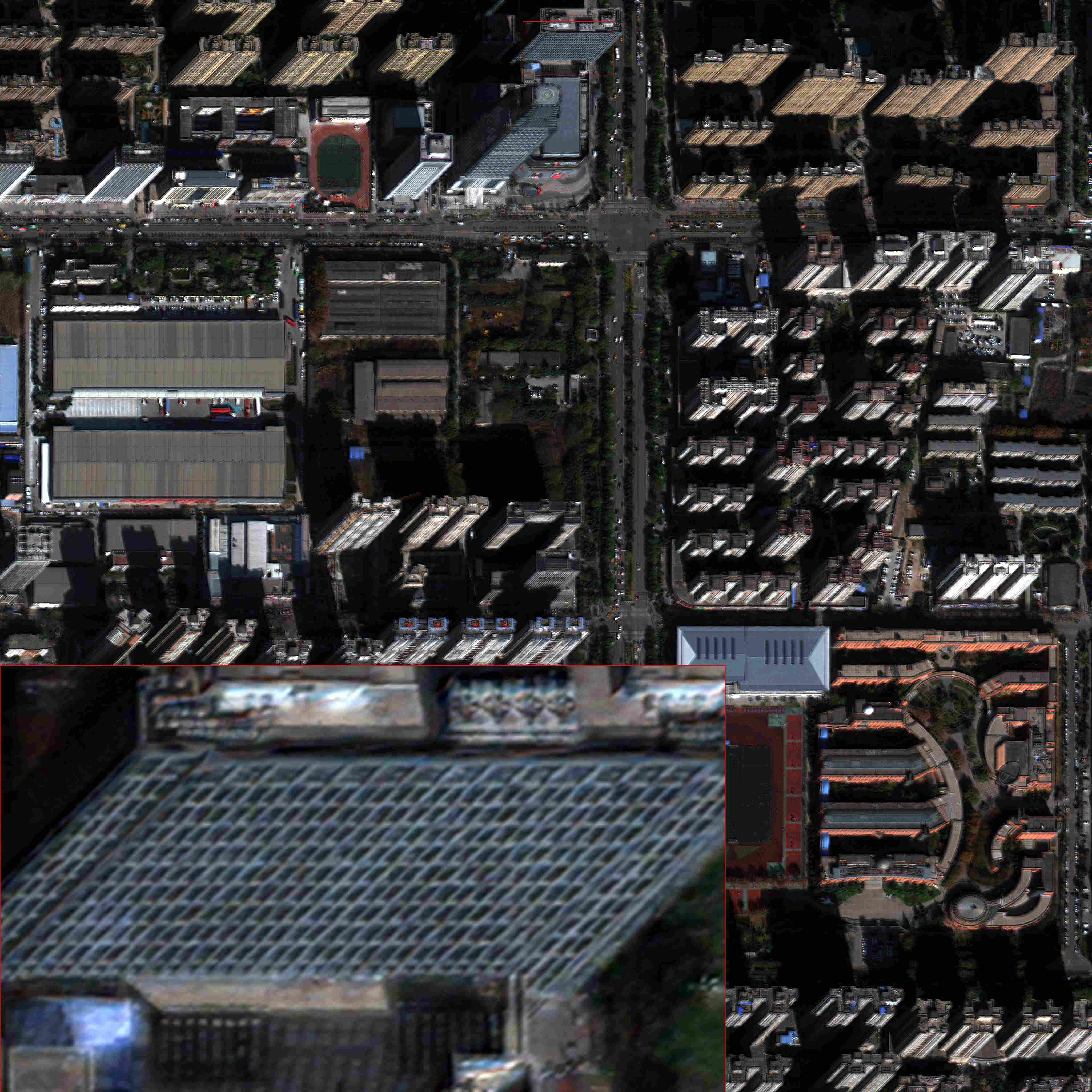} &
				\includegraphics[width=\mywkv]{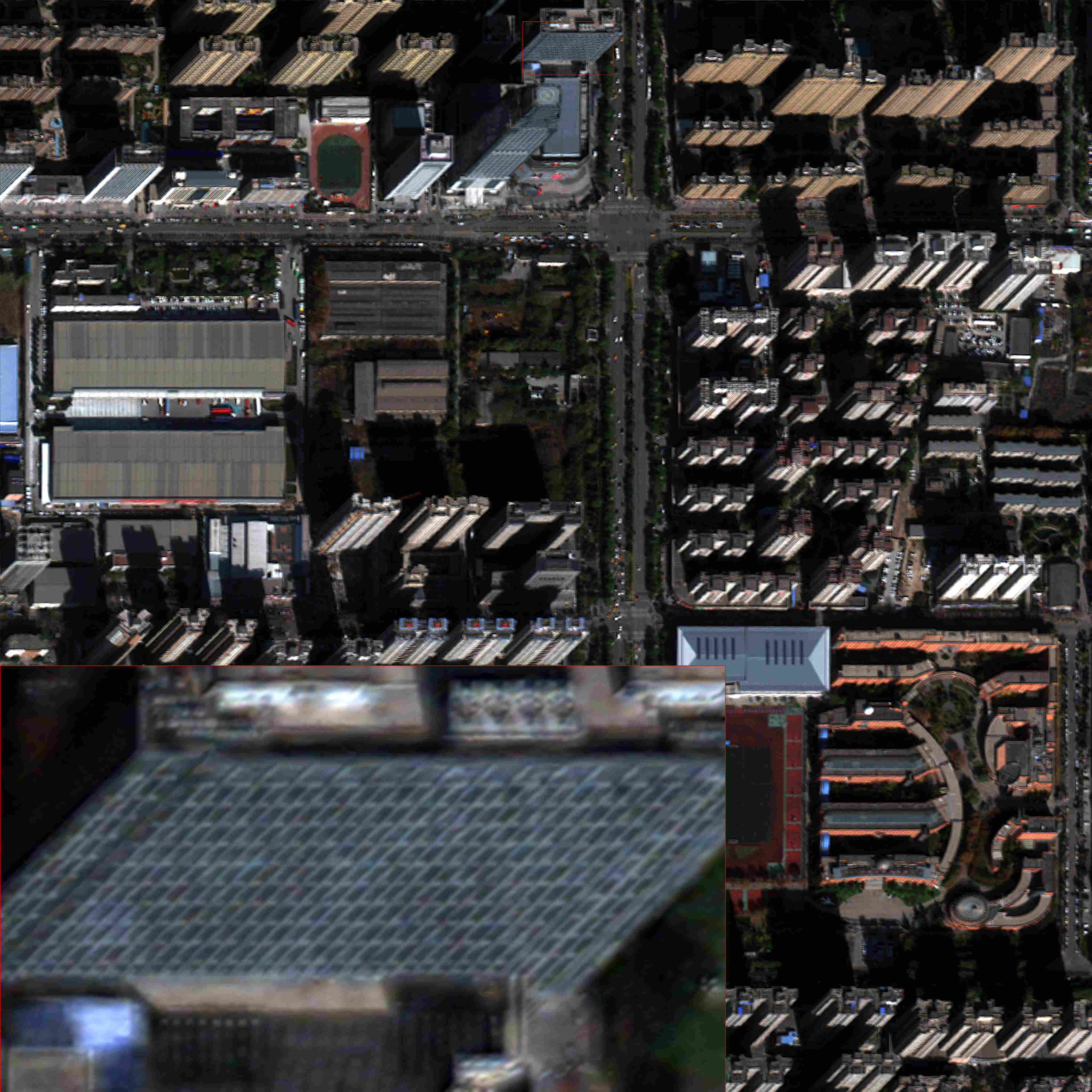} & 
				
				\multirow{6}{*}[\mywkv+0.5pt]{\includegraphics[height=\dimexpr 4\mywkv + 34.5pt\relax, trim=10 0 0 0, clip]{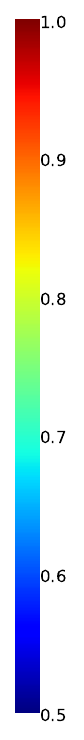}} \\[-1pt]
				
				\includegraphics[width=\mywkv]{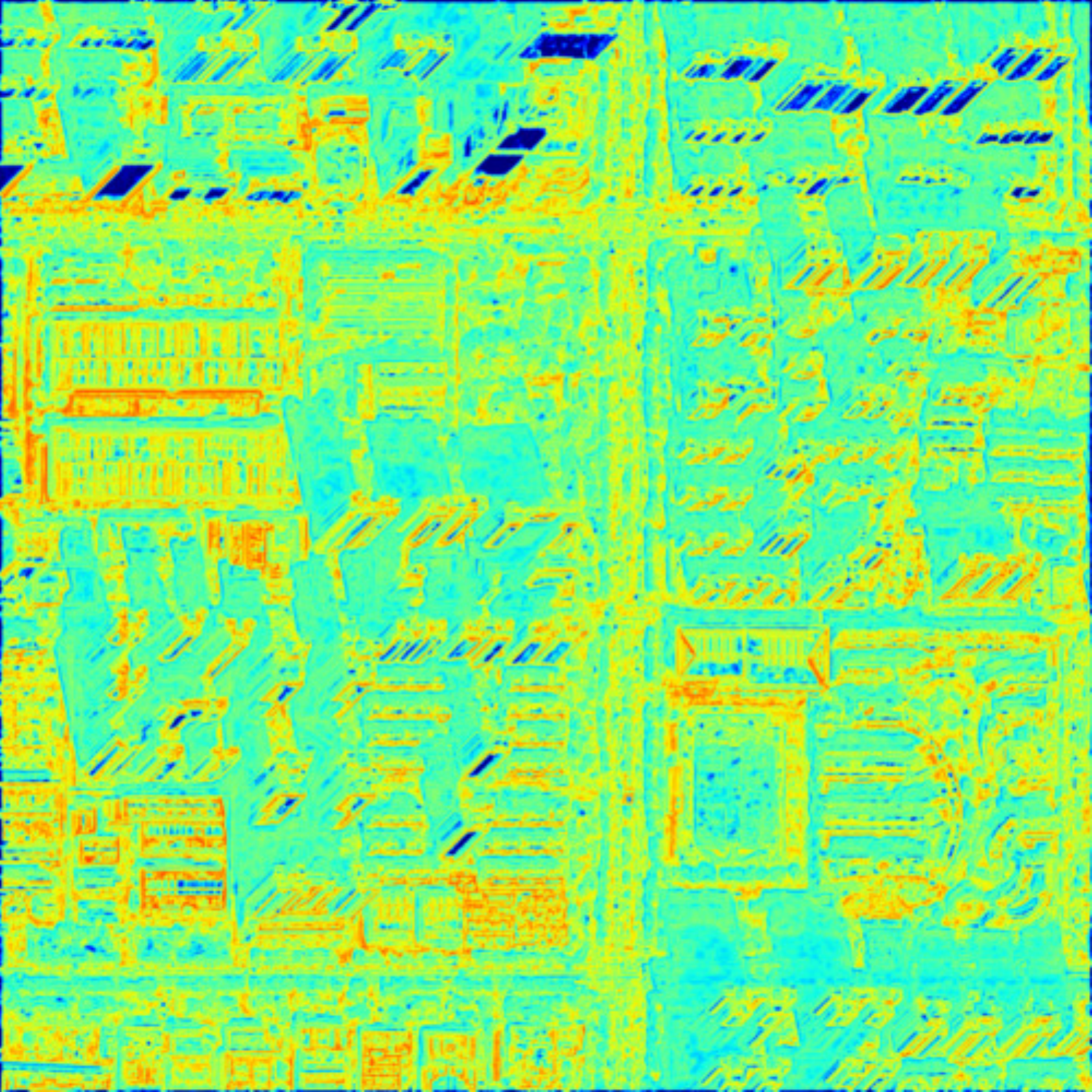} &
				\includegraphics[width=\mywkv]{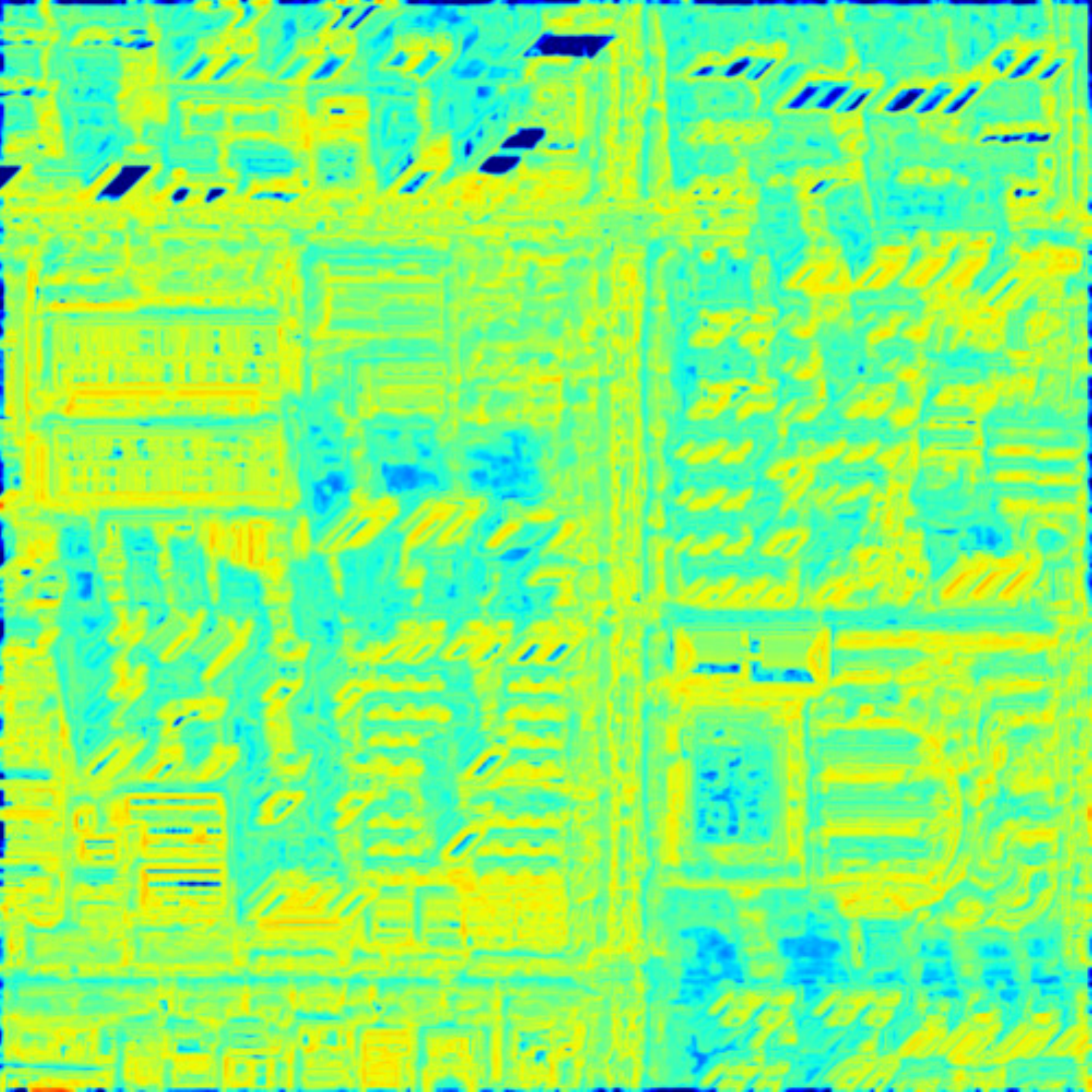} &
				\includegraphics[width=\mywkv]{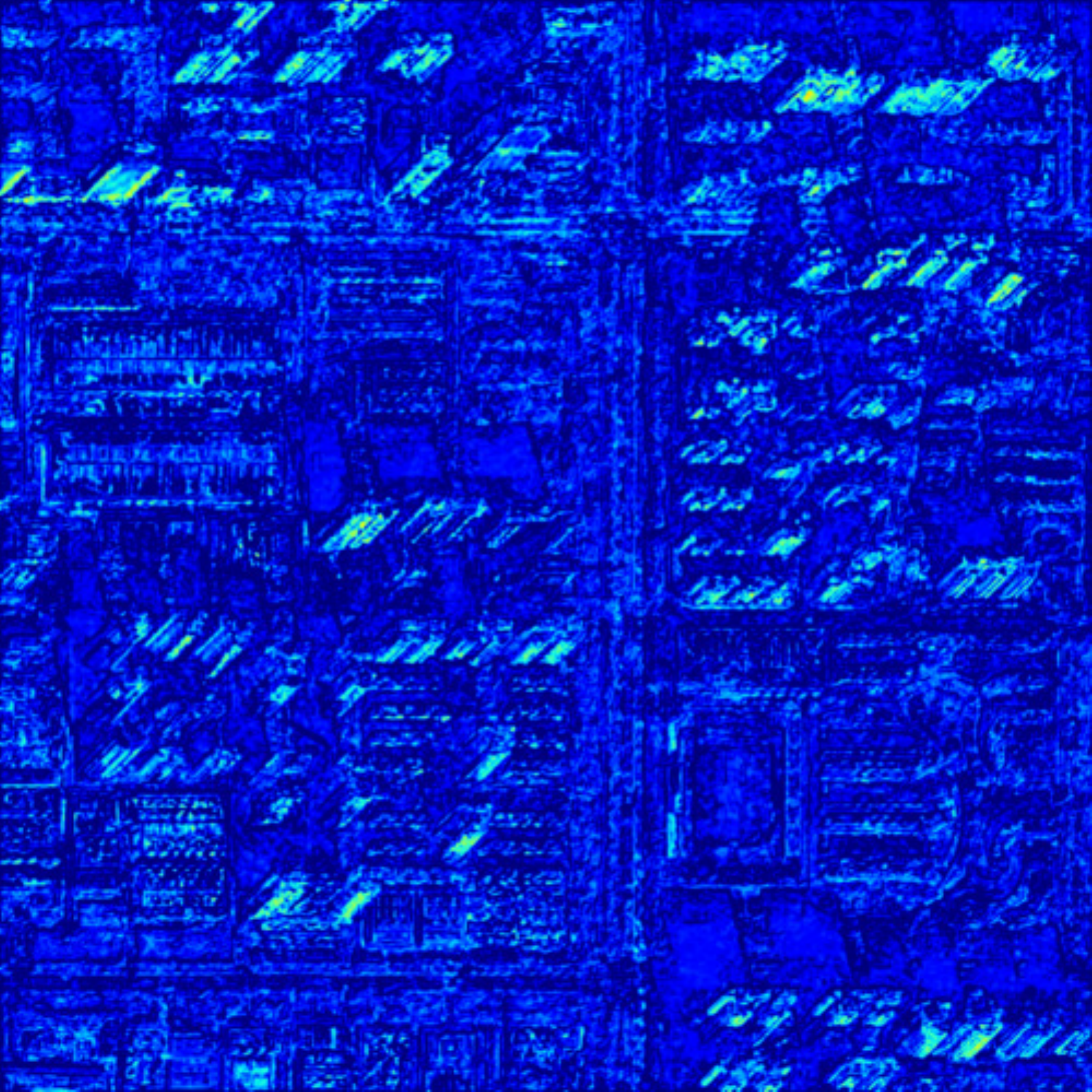} &
				\includegraphics[width=\mywkv]{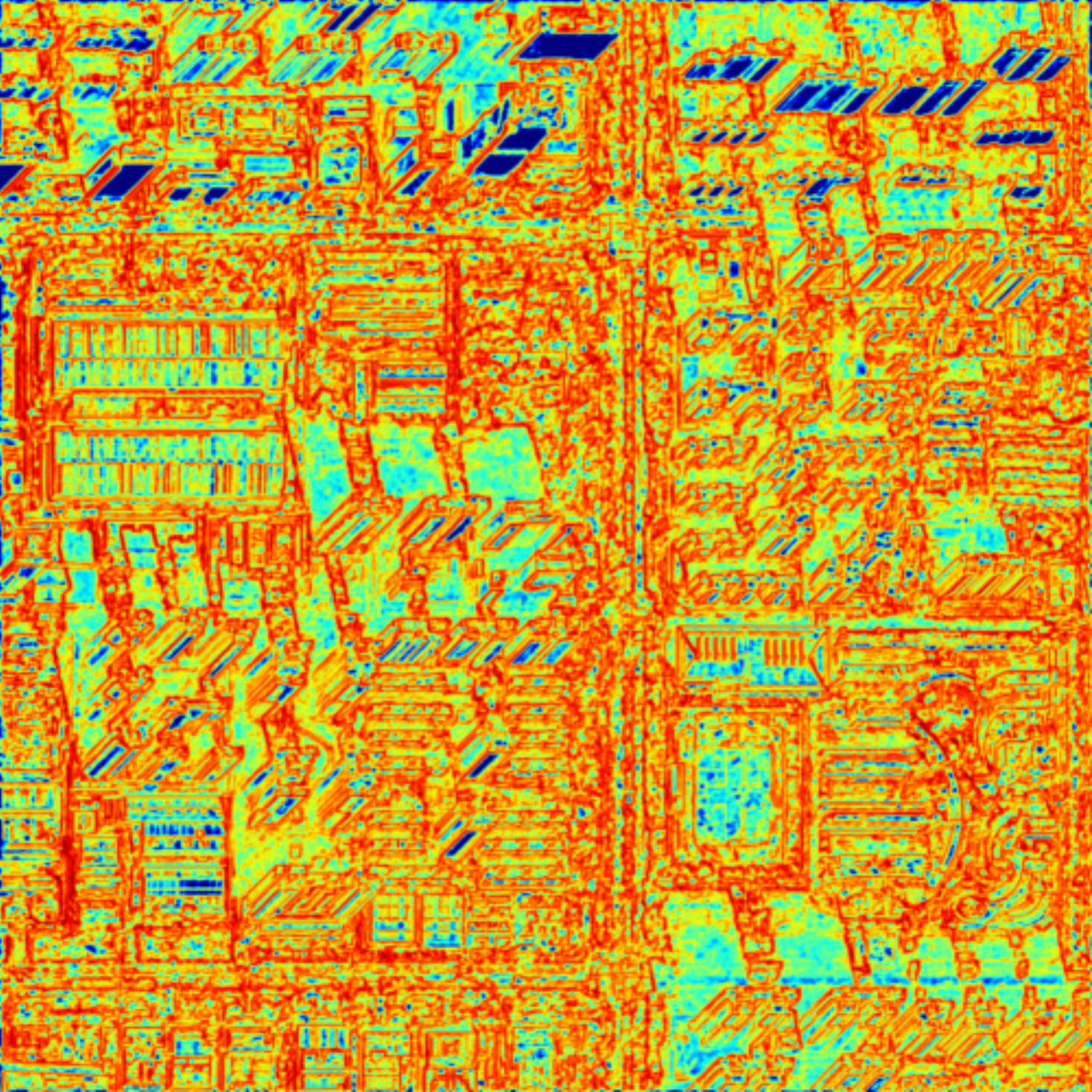} &
				\includegraphics[width=\mywkv]{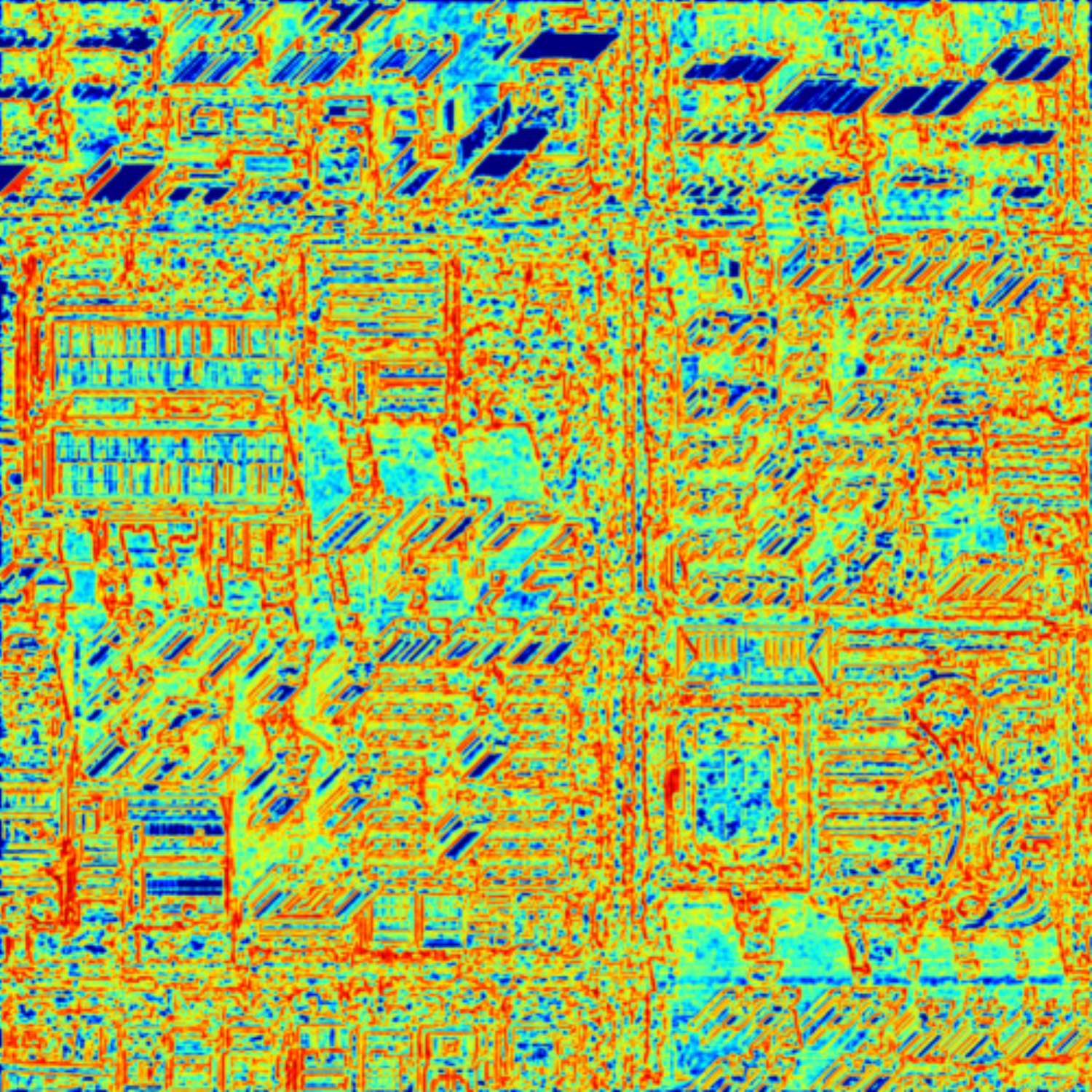} &
				\includegraphics[width=\mywkv]{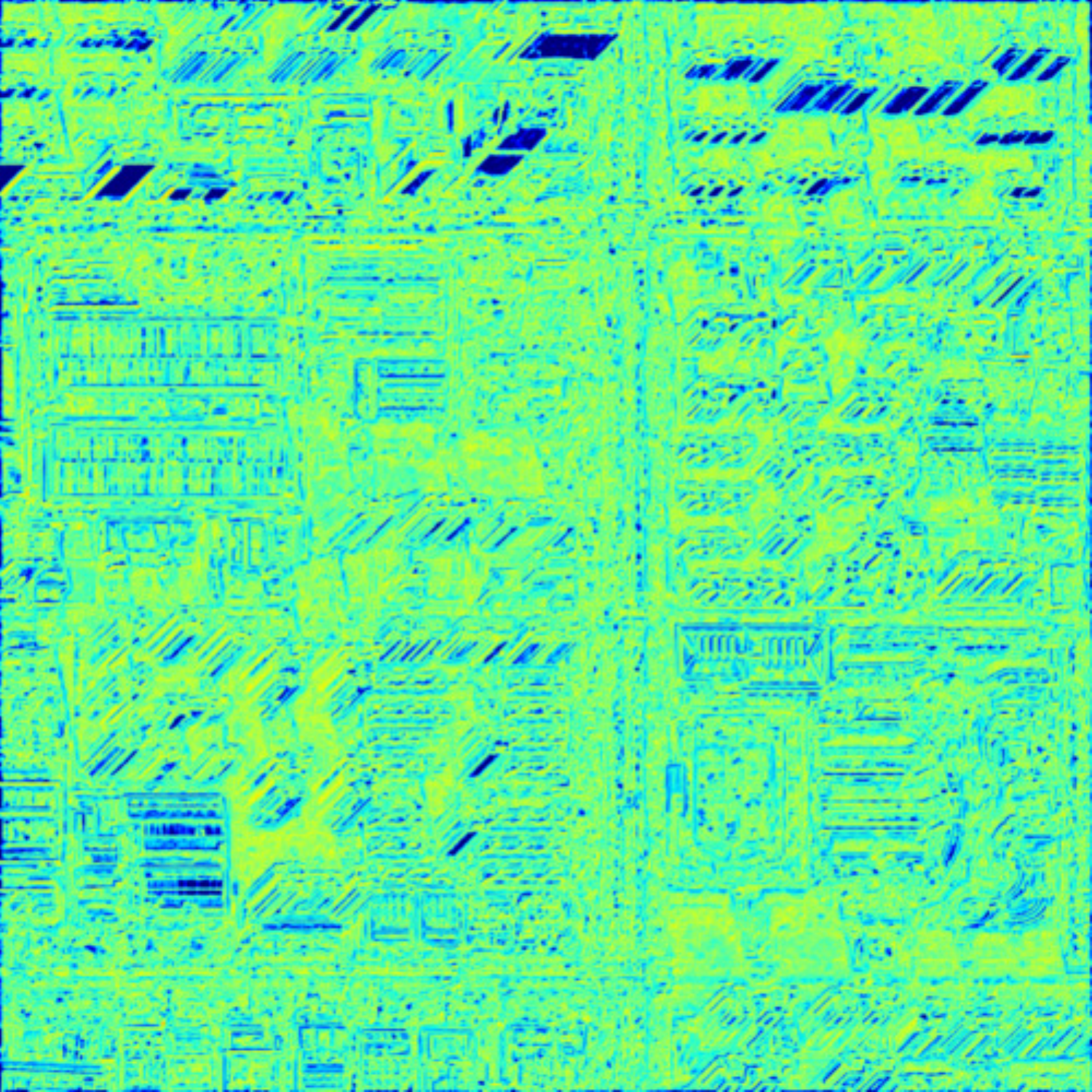} & \\[-2pt]
				
				\small AWLP & \small TV & \small BDSD & \small PNN & \small FusionNet & \small LAGConv & \\[-2pt] 
				
				\includegraphics[width=\mywkv]{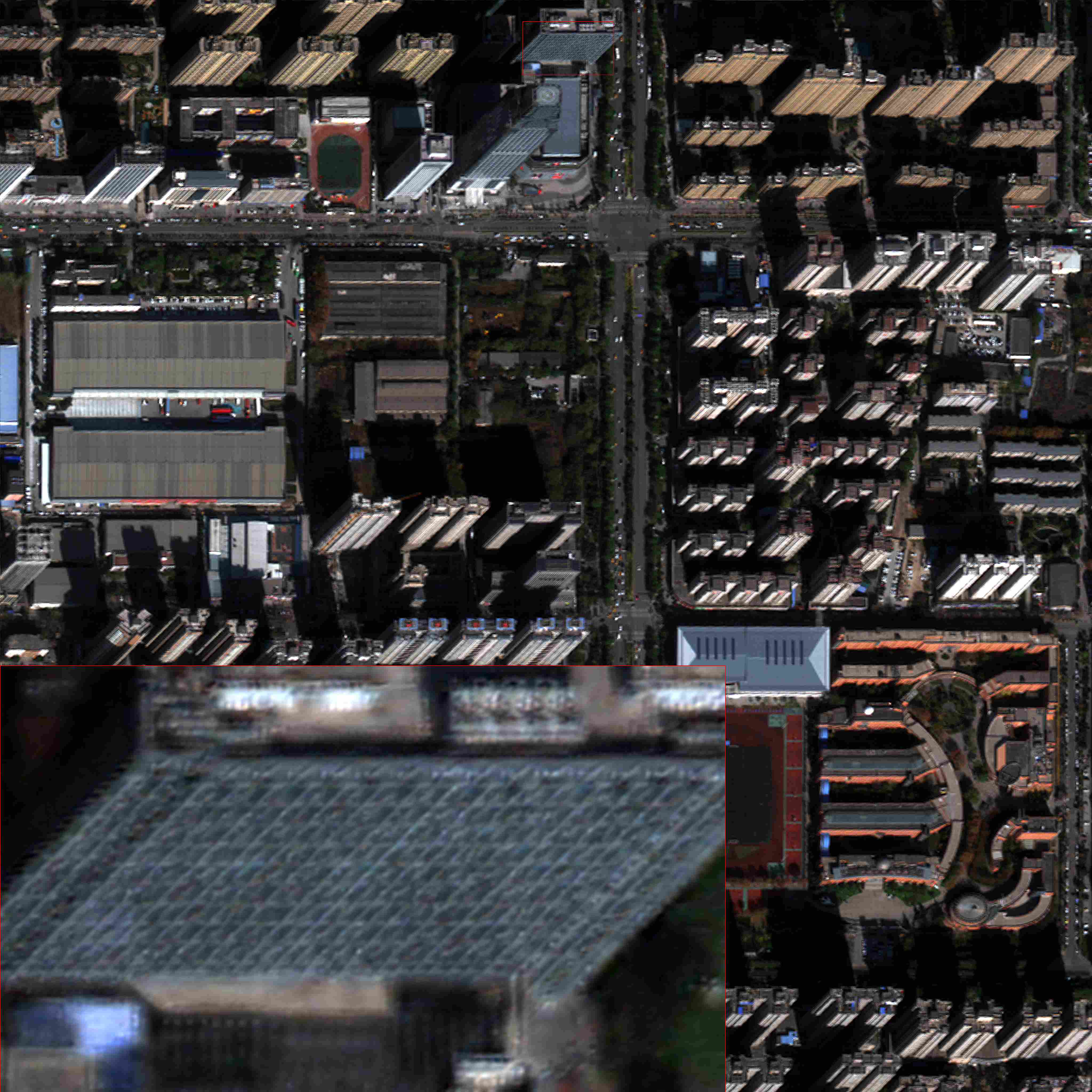} &
				\includegraphics[width=\mywkv]{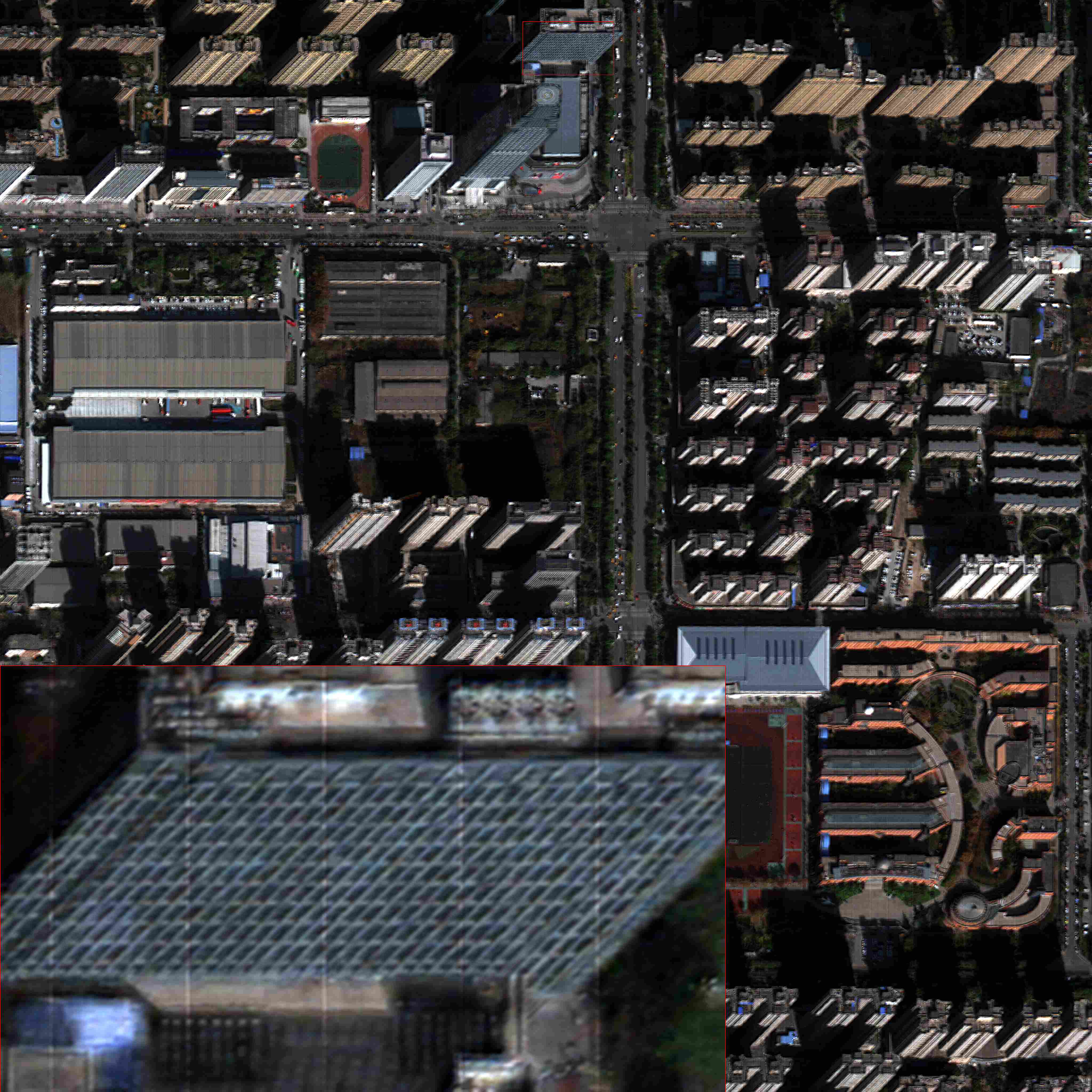} &
				\includegraphics[width=\mywkv]{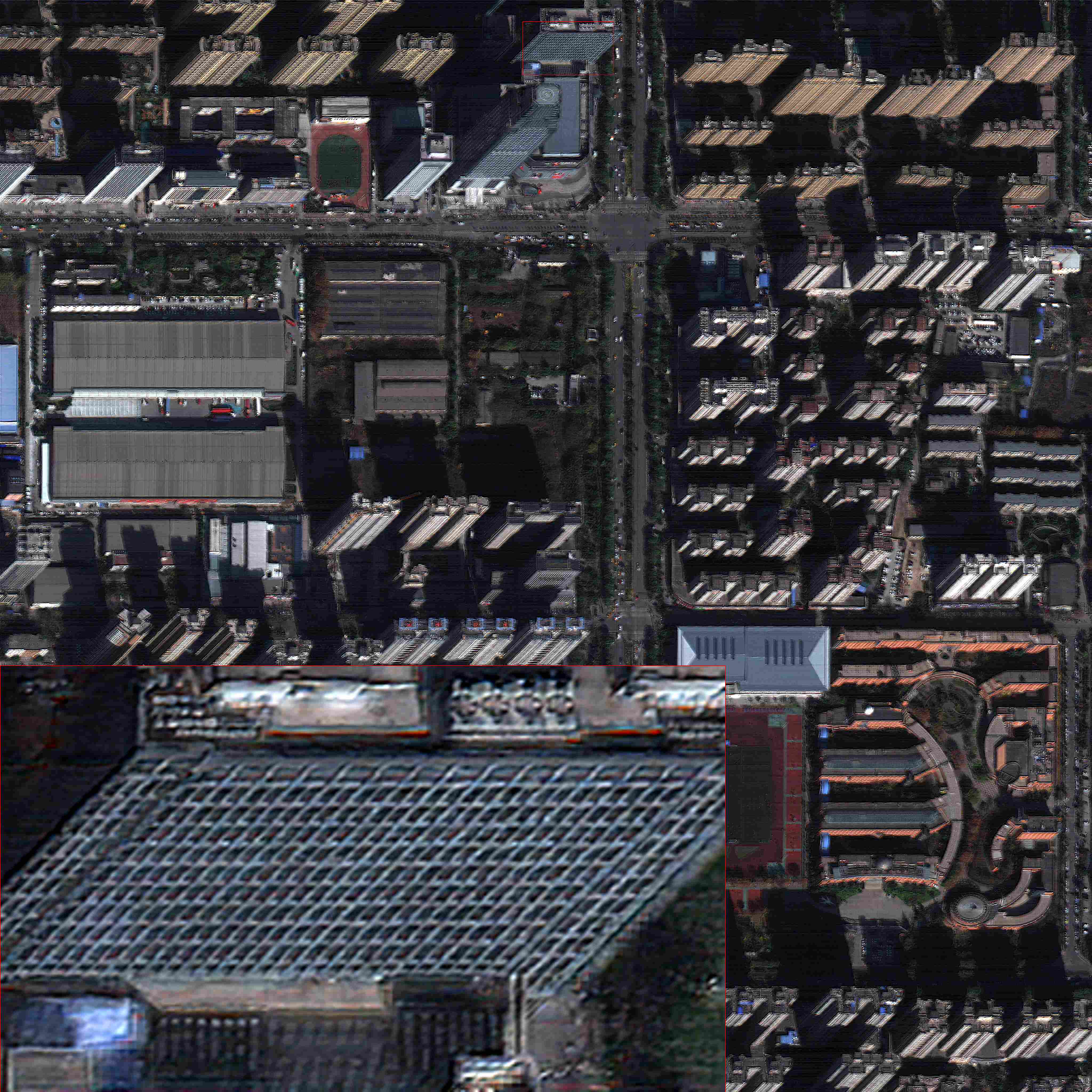} &
				\includegraphics[width=\mywkv]{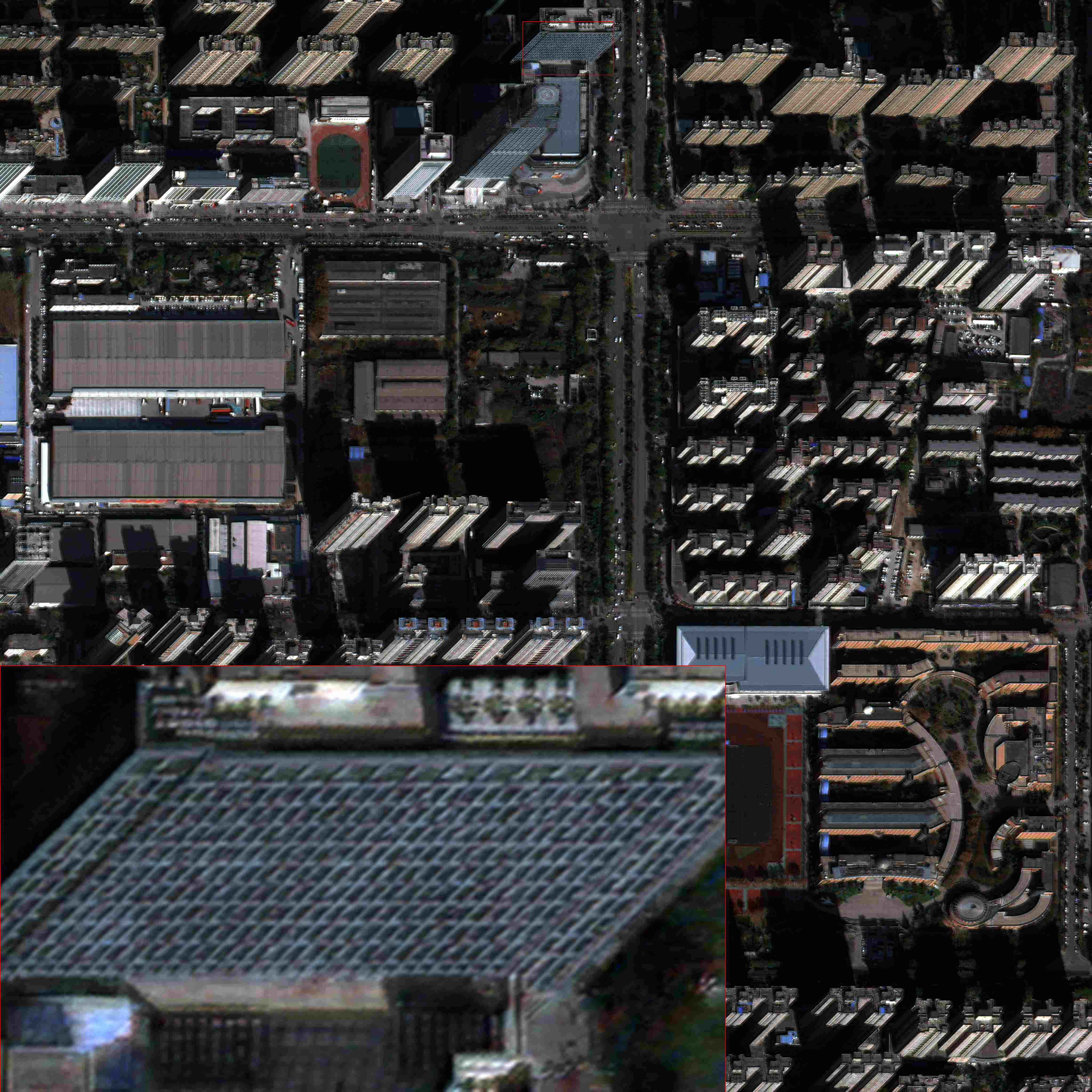} &
				\includegraphics[width=\mywkv]{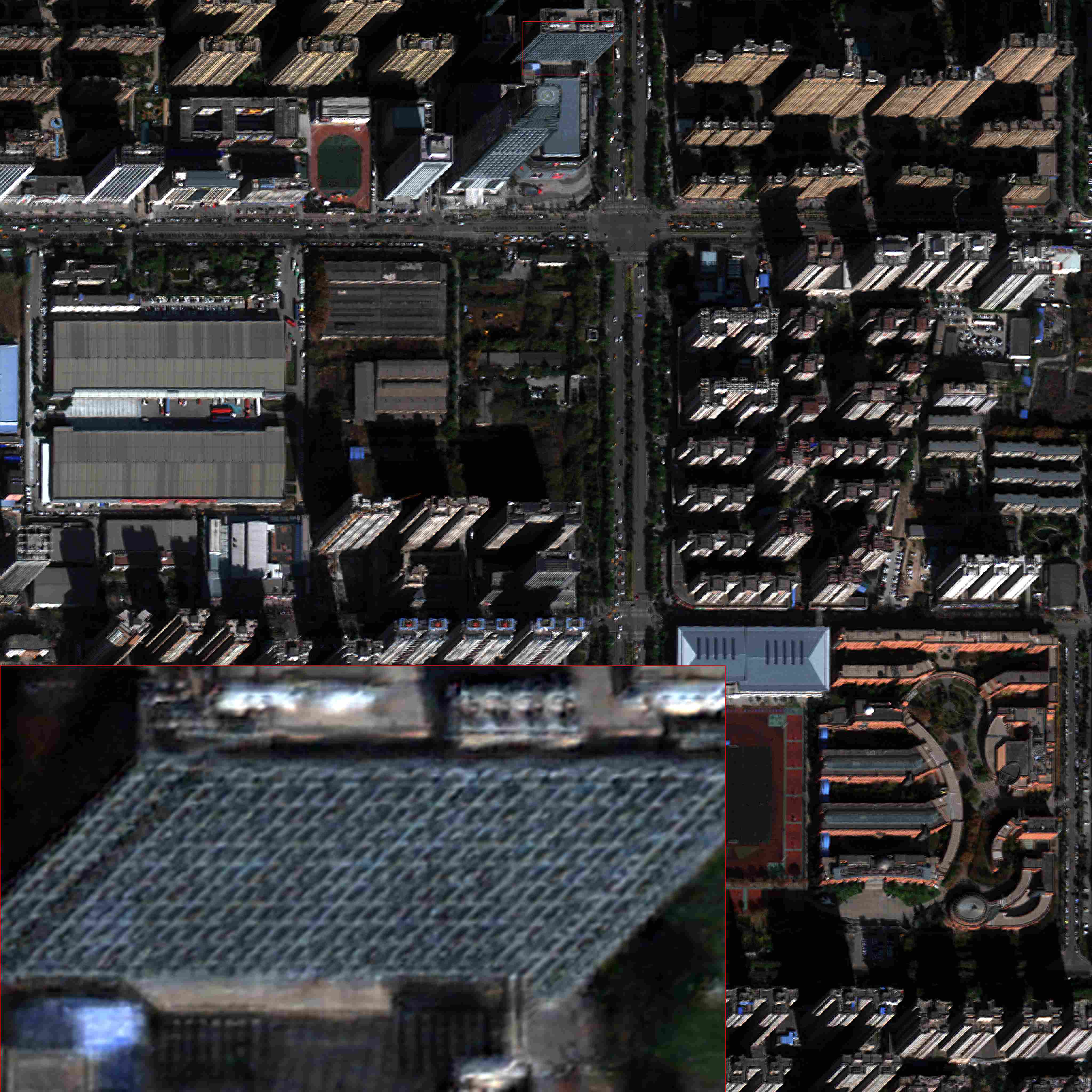} &
				\includegraphics[width=\mywkv]{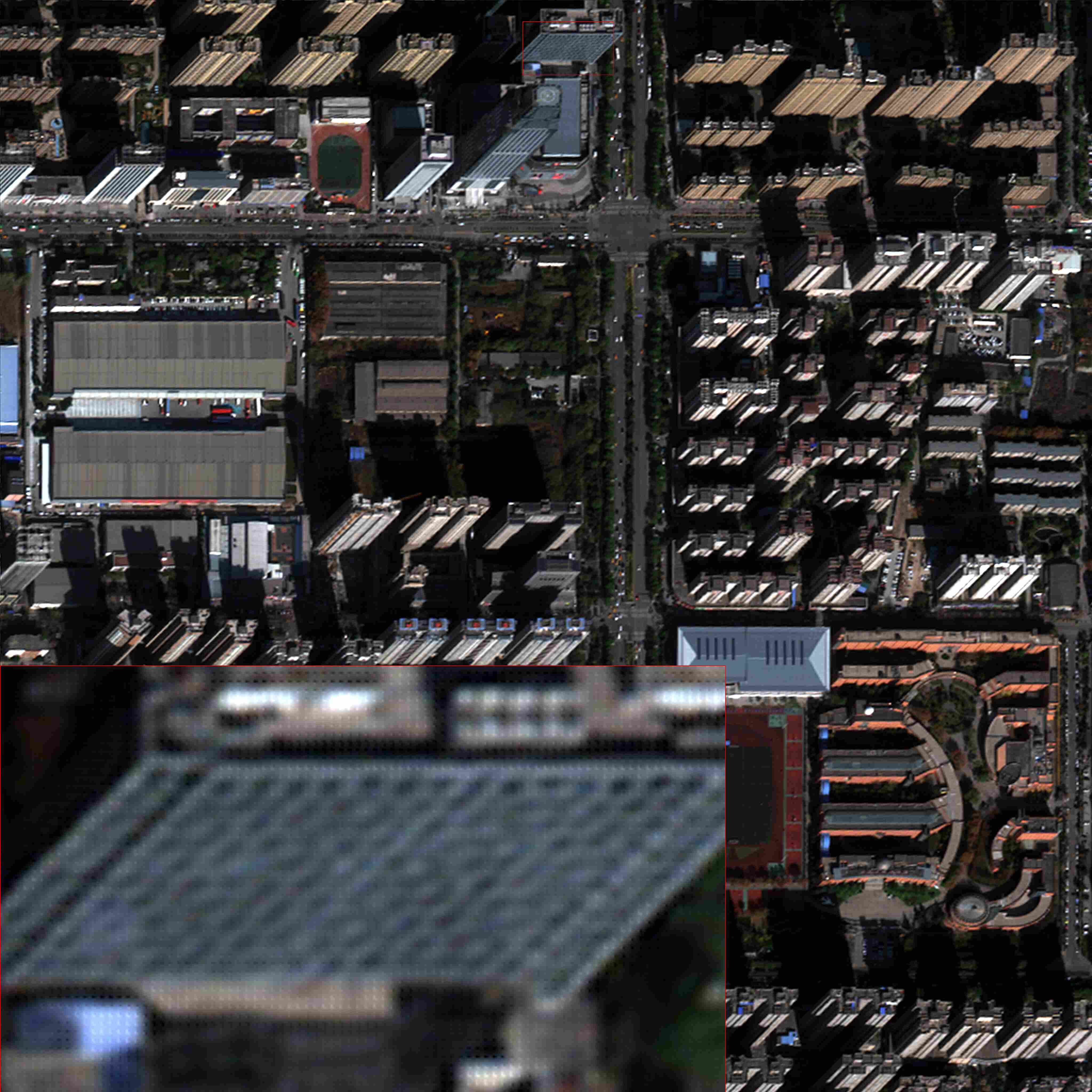} \\[-1pt]
				
				\includegraphics[width=\mywkv]{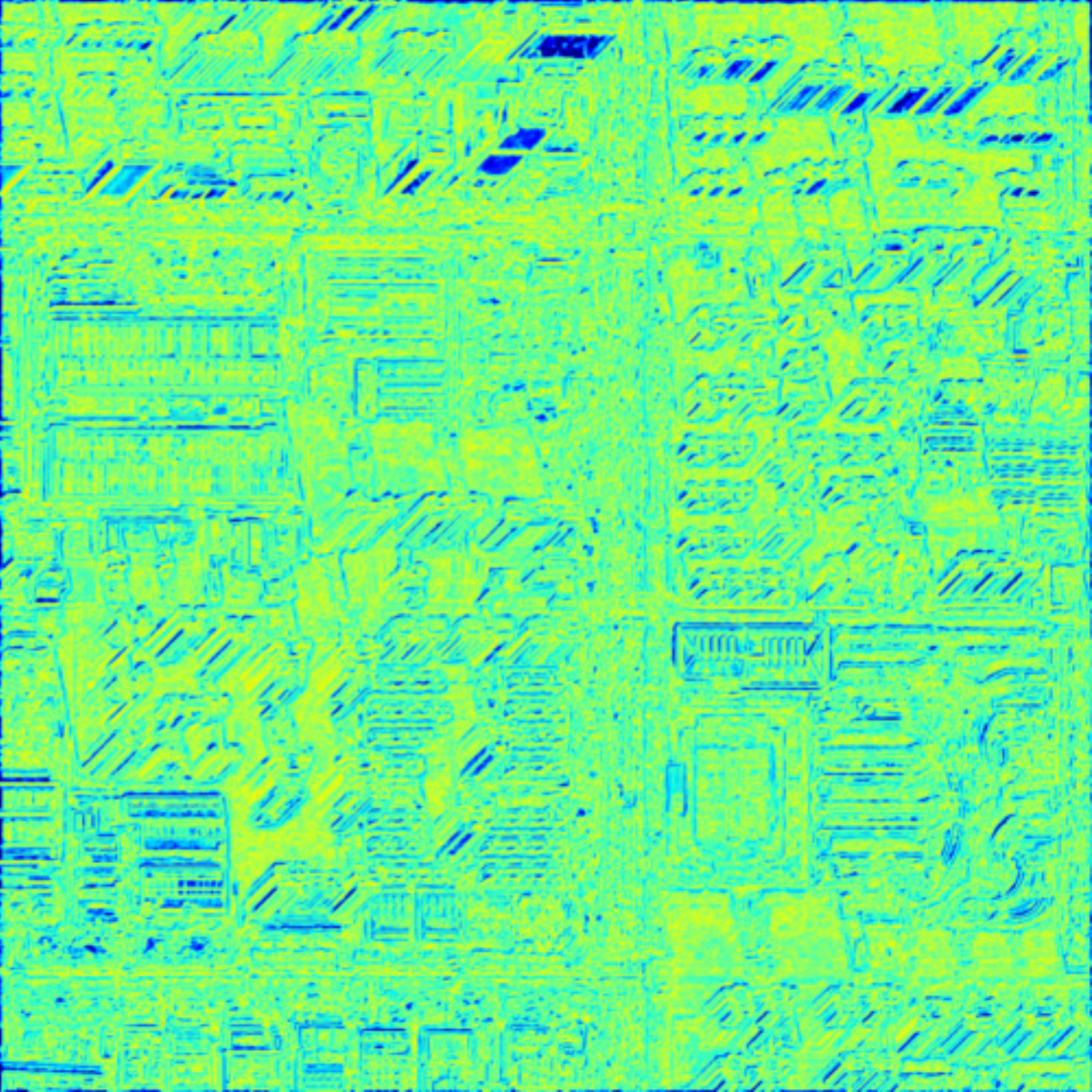} &
				\includegraphics[width=\mywkv]{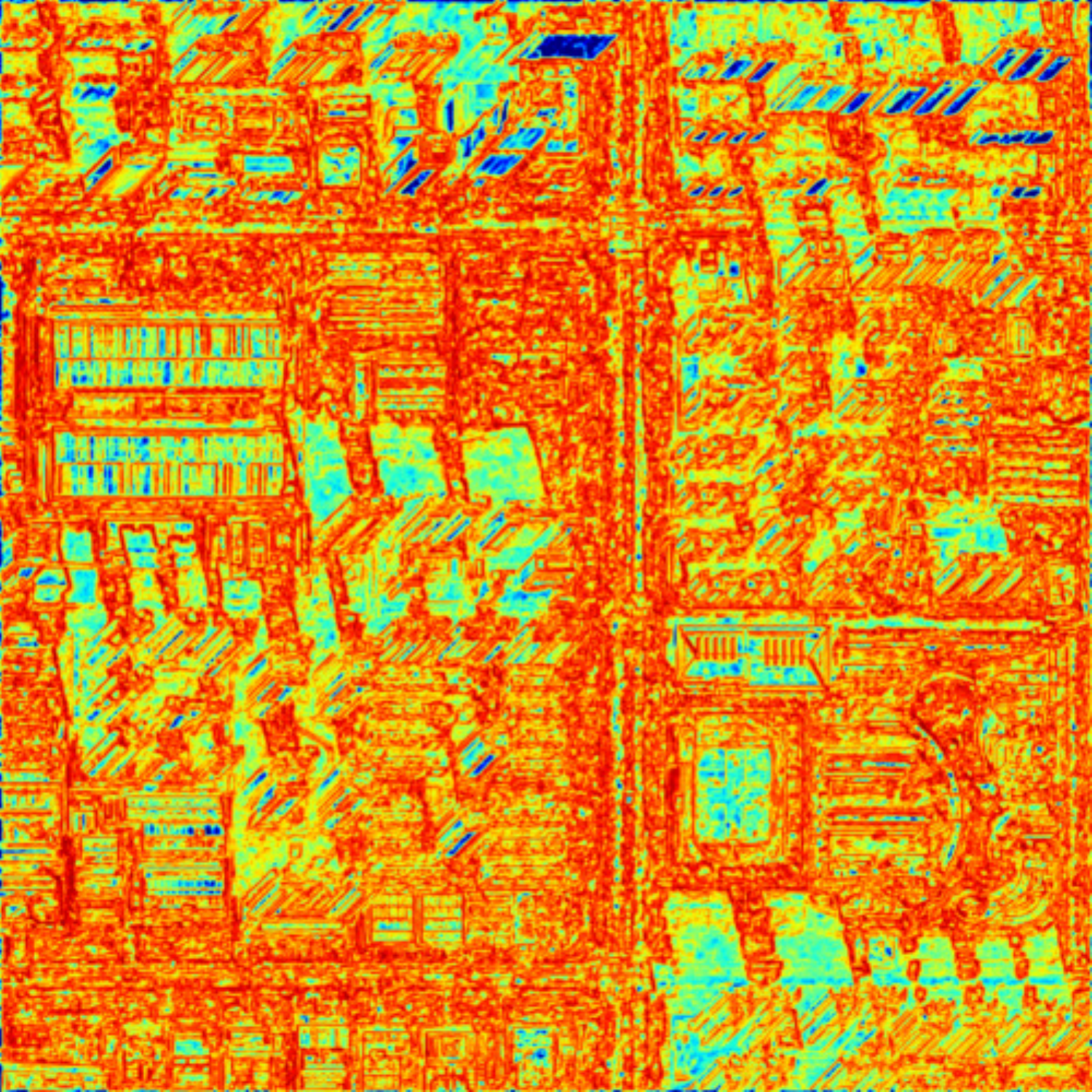} &
				\includegraphics[width=\mywkv]{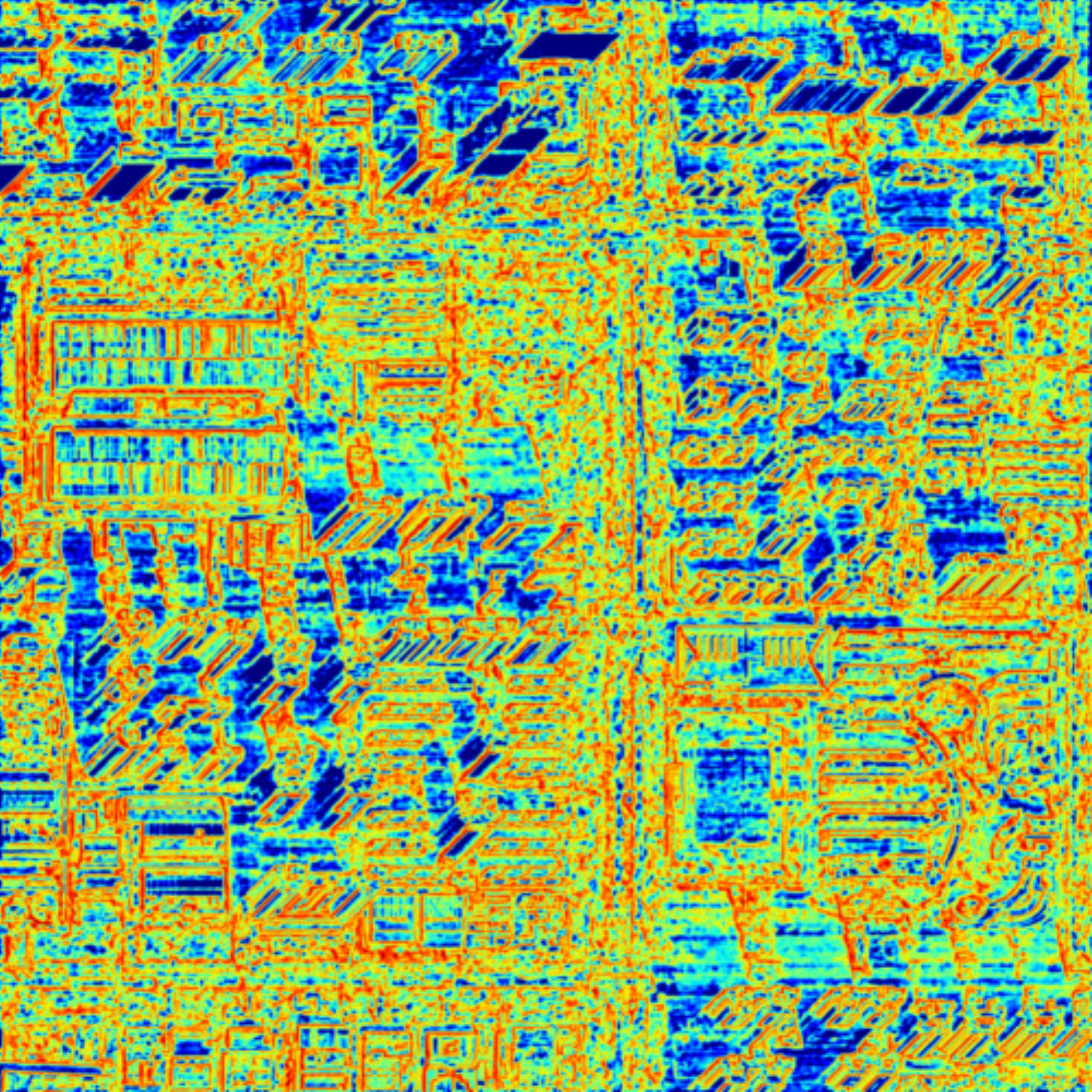} &
				\includegraphics[width=\mywkv]{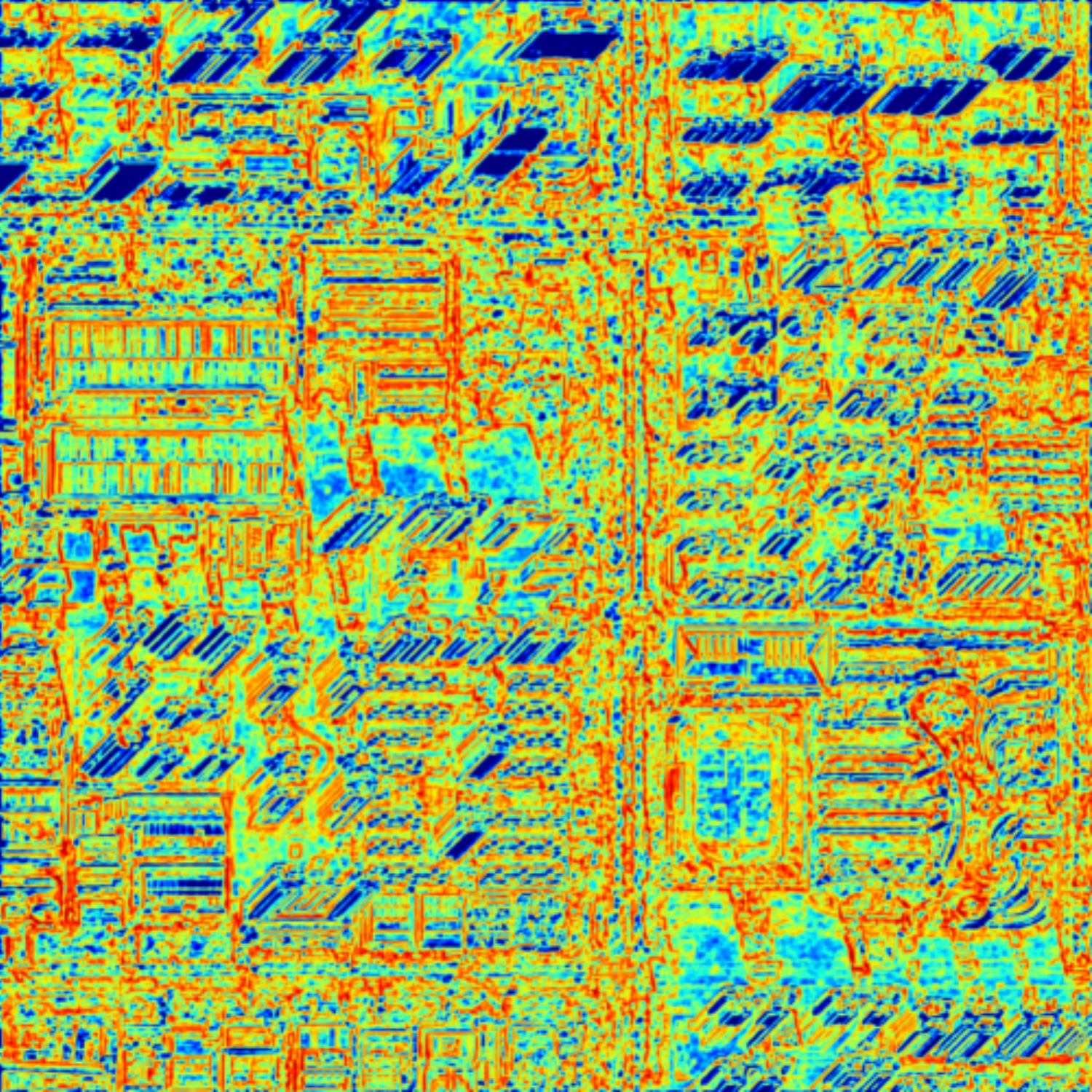} &
				\includegraphics[width=\mywkv]{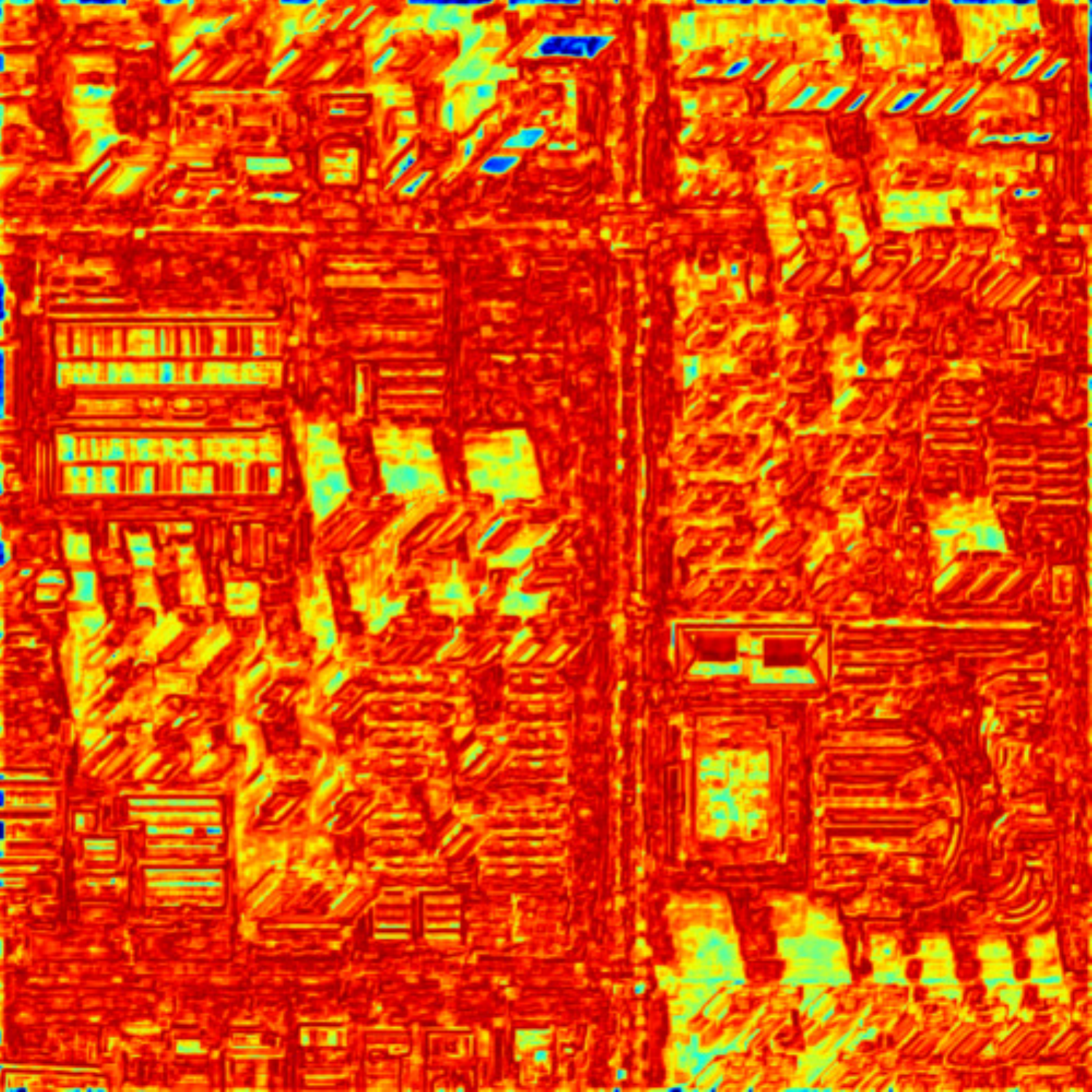} &
				\includegraphics[width=\mywkv]{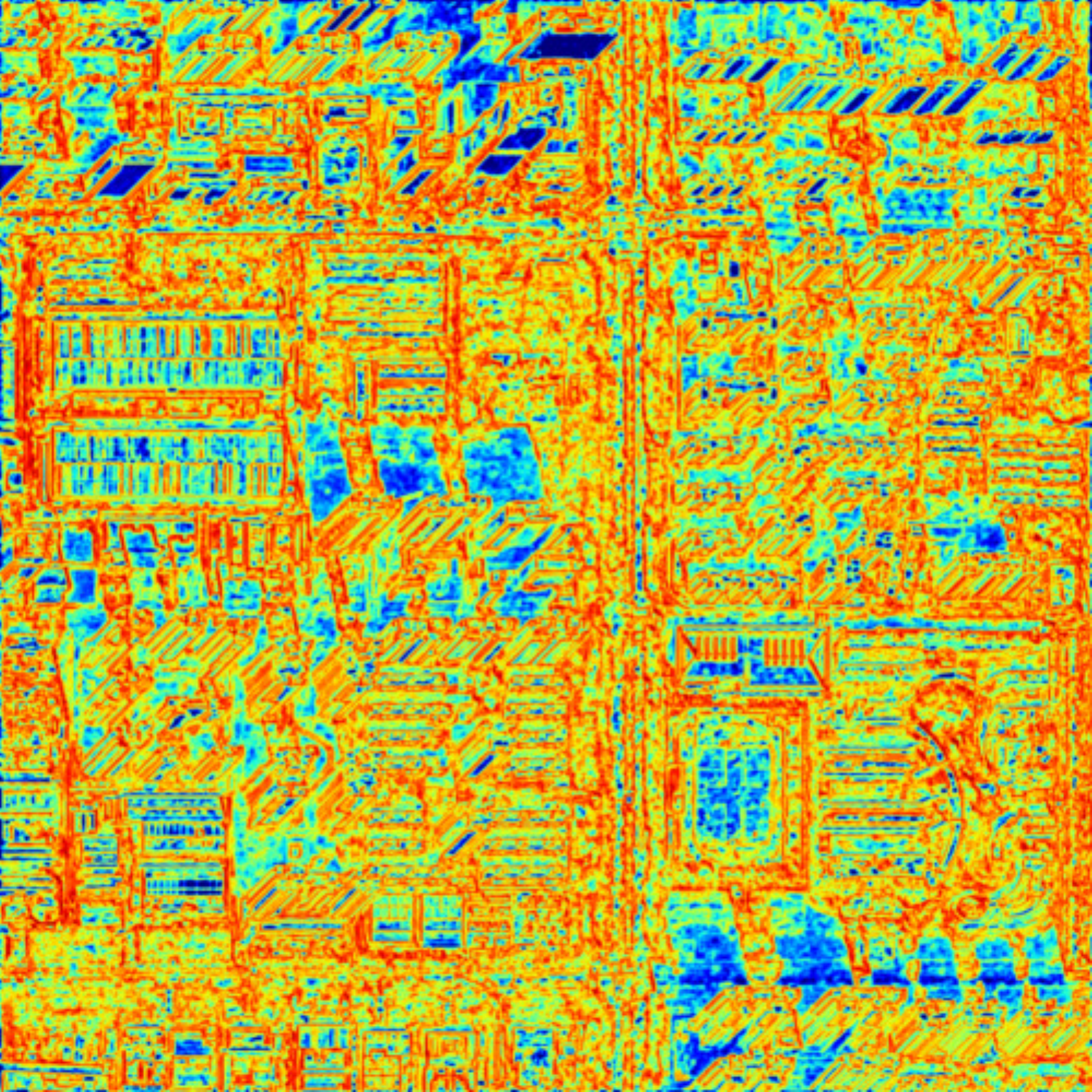} \\[-2pt]
				
				\small PSGAN & \small MSAN & \small PanMamba & \small MARNet & \small GSPan  & \small $\mathrm{GSPan}^*$  \\[-6pt]
			\end{tabular}
			
			\caption{Qualitative comparison on the WV3-4K dataset. The first and third rows show fused images and the second and fourth  rows show HQNR maps.}
			\label{fig:wv3_4k_comparison}
		\end{figure*}

		\begin{figure*}[htbp]
			\centering
	
			\captionsetup[subfloat]{labelformat=empty, skip=2pt}
			\setlength{\tabcolsep}{2pt} 
			
			\begin{tabular}{ccccc}
				
				\includegraphics[width=0.19\textwidth]{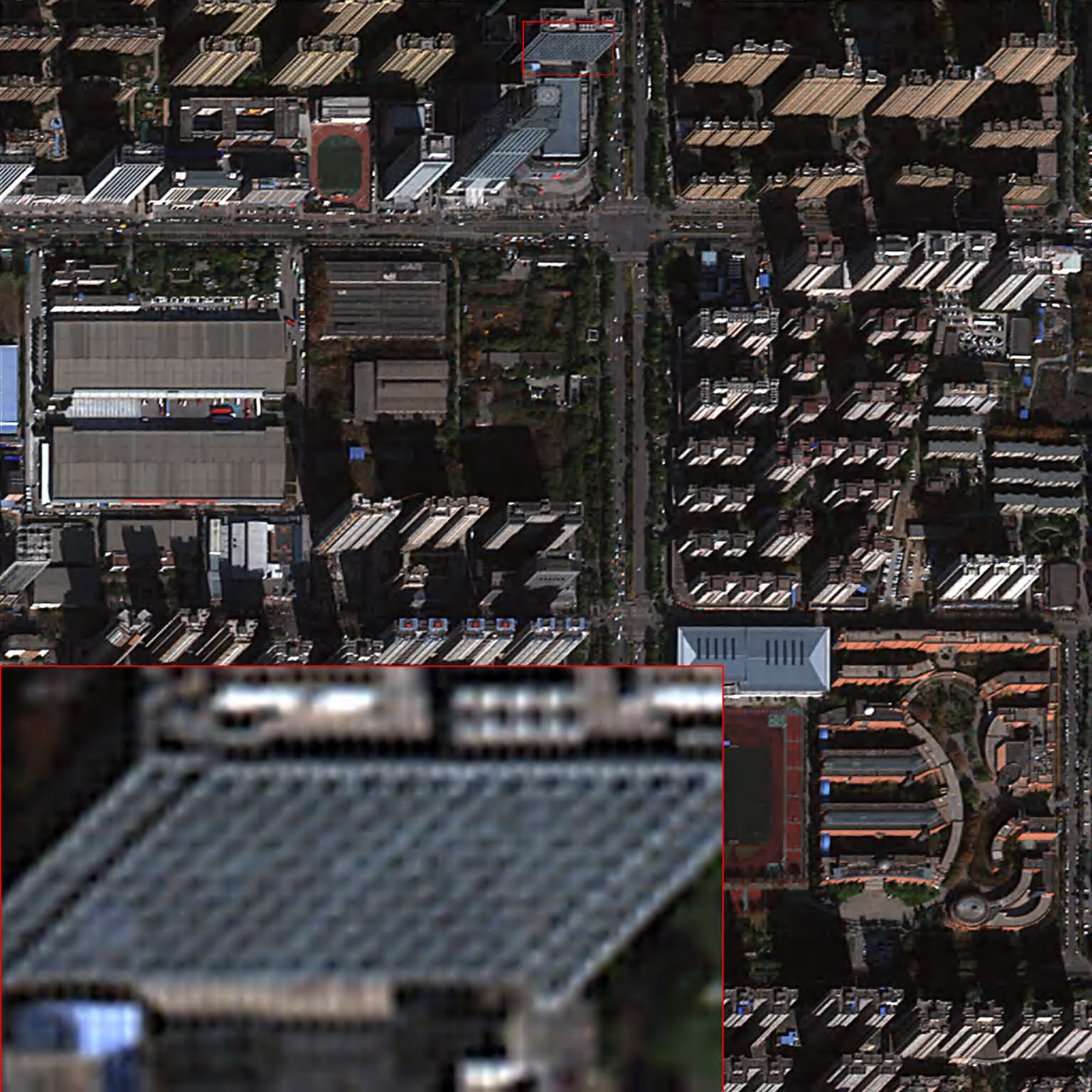} &
				\includegraphics[width=0.19\textwidth]{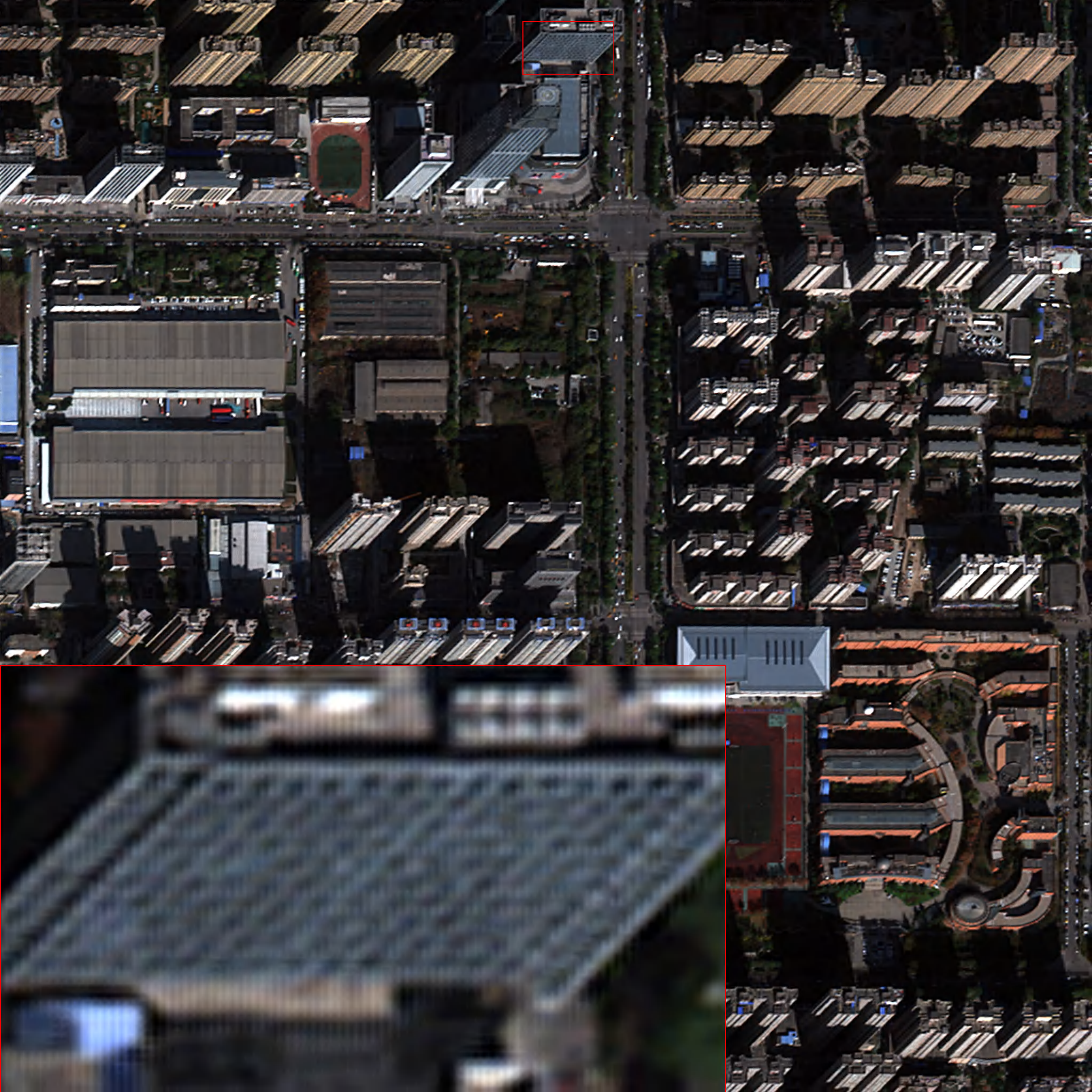} &
				\includegraphics[width=0.19\textwidth]{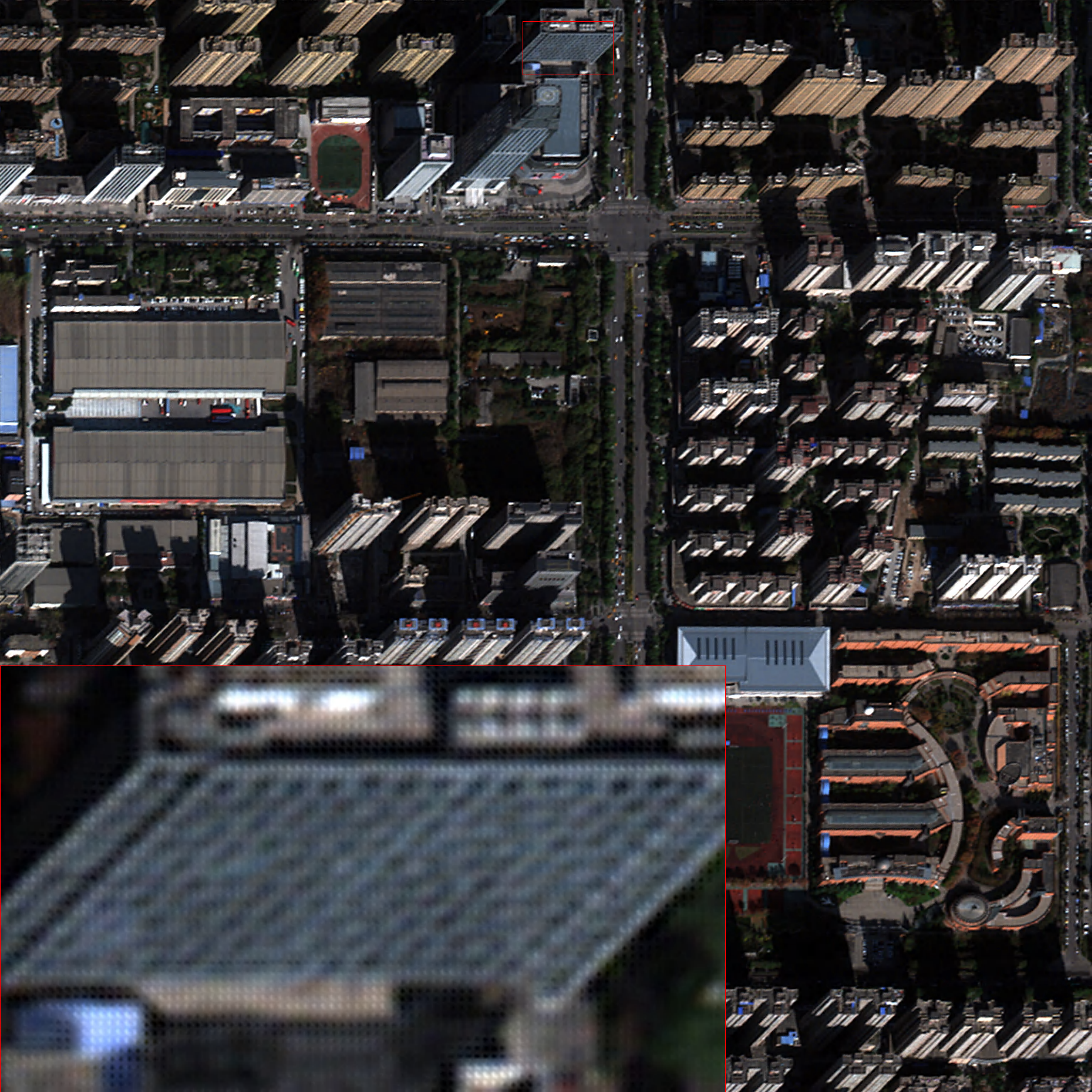} &
				\includegraphics[width=0.19\textwidth]{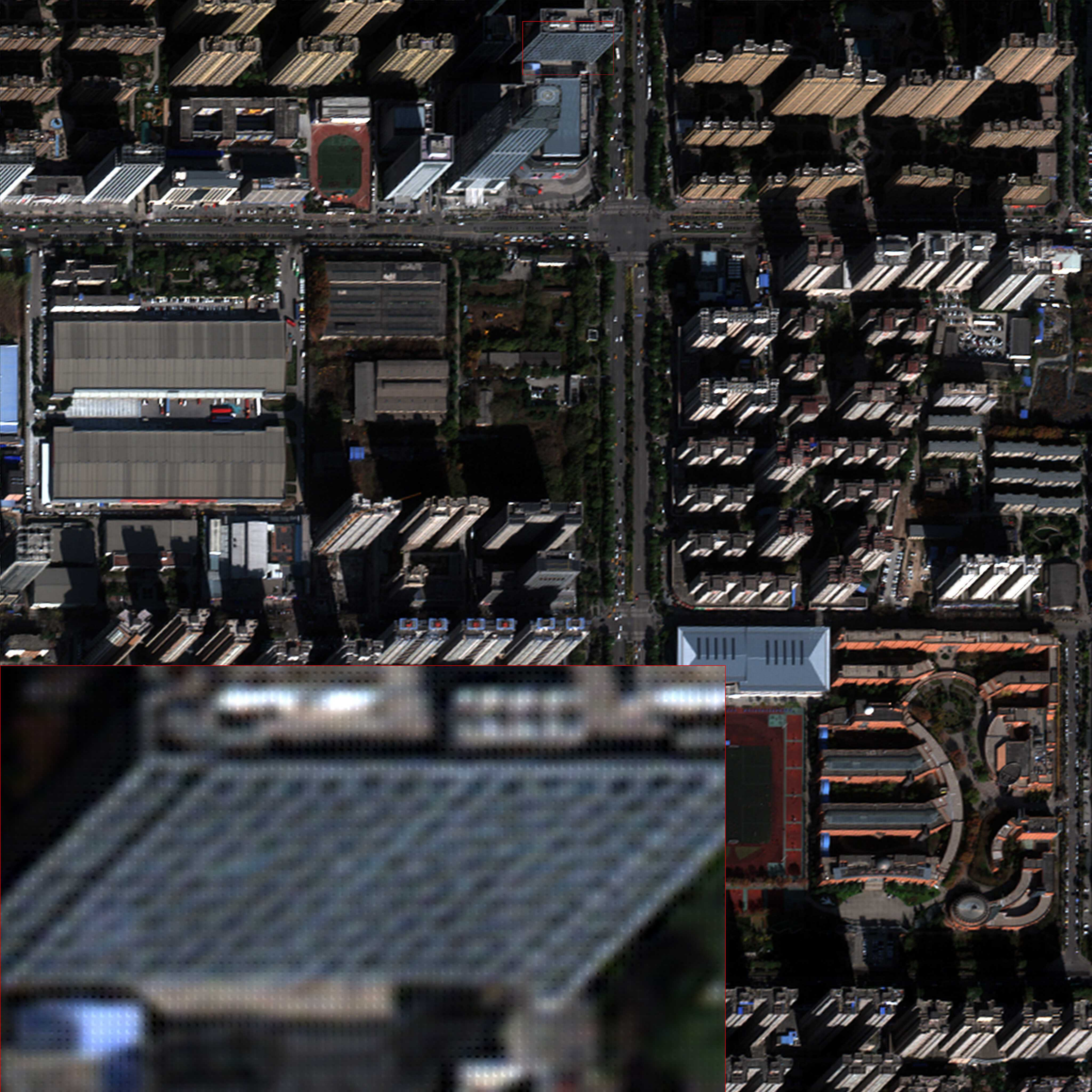} &
				\includegraphics[width=0.19\textwidth]{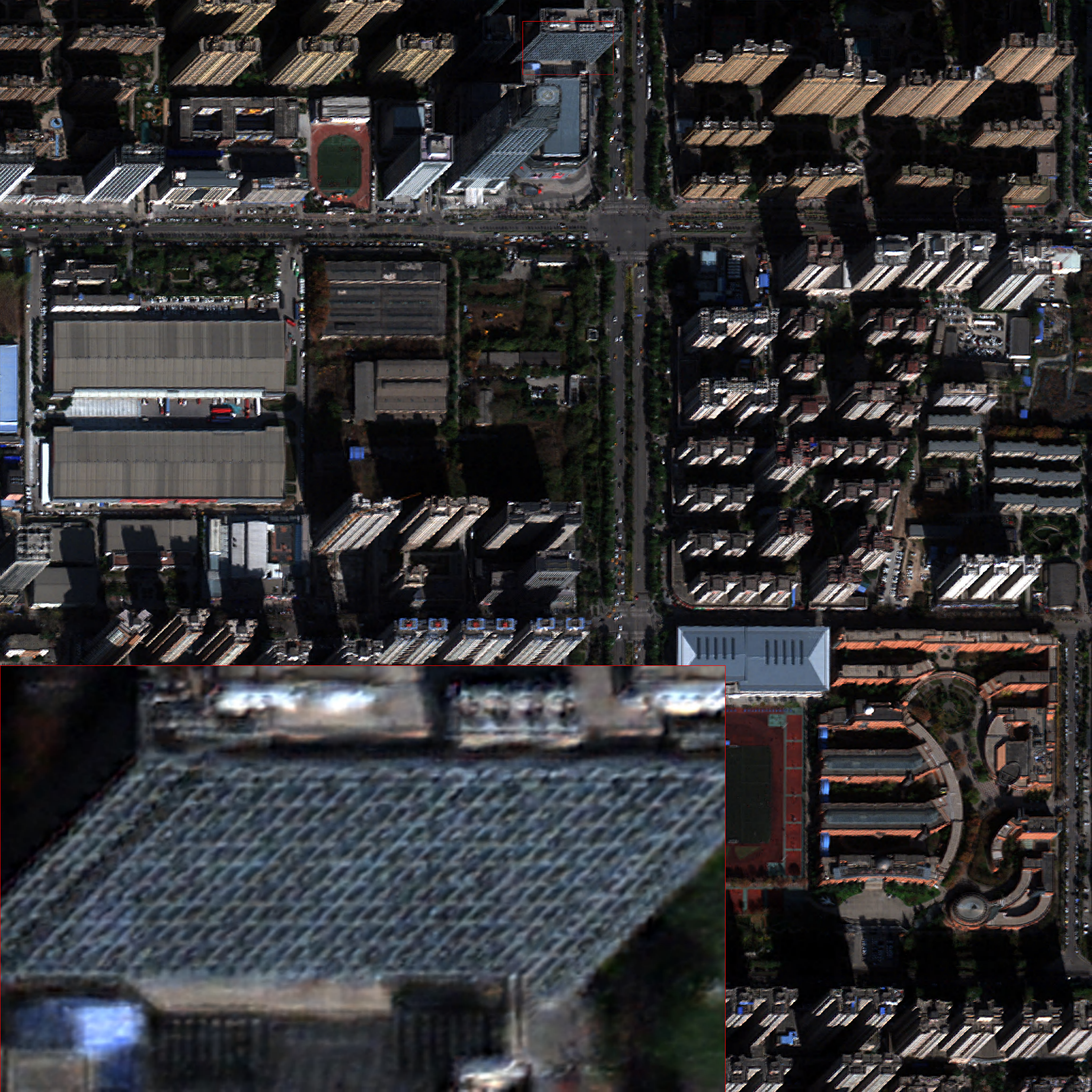} \\

				\small Scale $\times$ 1.5 & \small Scale $\times$ 2 & \small Scale $\times$ 3 & \small Scale $\times$ 4 & \small FR fused result \\
			\end{tabular}
			
			\caption{Arbitrary-scale pansharpening results on WV3-4K. Images from  $\times$ 1.5 to  $\times$ 4 are rendered from RR inputs by changing the target sampling grid, and are compared with the FR fused result obtained from original-scale inputs. Zoomed-in patches show that our method provides stable reconstruction quality and high-fidelity textures across different spatial resolutions.}
			\label{fig:arbitrary_scale}
		\end{figure*}

        \subsection{Arbitrary-Scale Rendering and Scale-Decoupled Asymmetric Inference}
        \label{asr}
		
		\textbf{Arbitrary-scale rendering.}
		The arbitrary-scale rendering ability of GSPan comes from its continuous Gaussian primitive representation. GSPan predicts a set of Gaussian attributes that define a continuous residual field in a normalized coordinate space, rather than directly producing a fused image on a fixed output grid. Given a target sampling grid $\Omega_s$ corresponding to an arbitrary output scale $s$, the fused result $\hat{\mathbf{I}}_s$ can be obtained by rendering the Gaussian residual field on $\Omega_s$ and adding it to the MS base upsampled to the same grid:
		\begin{equation}
		    \hat{\mathbf{I}}_s = \mathcal{U}_s(\mathbf{MS}) + {\mathcal{R_G}}(\Omega_s),
		\end{equation}	
		where $\mathcal{U}_s(\cdot)$ denotes the upsampling operation for the MS observation and ${\mathcal{R_G}}(\Omega_s)$ denotes the rendered Gaussian residual field. Therefore, changing the output resolution only requires changing the target sampling grid during rendering, without modifying the network architecture or retraining the model.
		
		For any given PAN-MS input pair, including the original full-resolution inputs and reduced-resolution inputs, GSPan can render the fused HRMS image at a user-specified target grid.

		Fig.~\ref{fig:arbitrary_scale} visualizes the arbitrary-scale rendering behavior of GSPan. The results from $\times 1.5$ to $\times 4$ are produced from reduced-resolution inputs,  by changing only the target rendering grid. The results show that a single trained model can generate visually stable fused images at different output scales without scale-specific retraining. 
		The comparison between the $\times 4$ result generated from reduced-resolution inputs and the ``FR fused result'' generated from the original-scale inputs further clarifies the role of input information. Although these two results have the same spatial size, the FR fused result contains sharper and richer structures because it is guided by the original full-resolution PAN image.

		\textbf{Scale-Decoupled Asymmetric Inference.}
		The above property further enables the Scale-Decoupled Asymmetric Inference (SDAI) strategy. In standard full-resolution inference, GSPan performs primitive attribute estimation at the target full resolution, and then renders the fused image on the same grid  (Fig.~\ref{fig:sdia}). This quality-oriented mode fully exploits the full-resolution PAN structures but incurs relatively high computational cost for large scenes.
		
		In contrast, SDAI decouples the resolution used for Gaussian primitive attribute estimation from the final rendering resolution. Specifically, it estimates Gaussian primitive parameters at a reduced estimation resolution and then renders the fused image at the target high-resolution grid using the same model weights. Thus, SDAI is an inference mode derived from arbitrary-scale Gaussian rendering. Since the reduced-resolution estimation stage cannot fully perceive fine PAN structures at the target resolution, SDAI is expected to sacrifice part of the spatial fidelity, reflected by a higher $D_s$. In return, it substantially reduces the computational burden and provides an efficiency-oriented option for large-scene processing.
		
		The quantitative results of SDAI are reported in the last two columns of Table~\ref{tab:wv3_4k_fr}, where $\mathrm{GSPan}^{*}$ denotes the proposed SDAI mode. Compared with standard GSPan, $\mathrm{GSPan}^{*}$ achieve approximately $11.1\times$ acceleration with the same model parameters. Meanwhile, $\mathrm{GSPan}^{*}$ obtains the lowest spectral distortion $D_{\lambda}=0.0136$ and an HQNR value of $0.8772$, which is higher than all competing baseline methods, though lower than standard GSPan. The main cost of SDAI lies in spatial fidelity. Its $D_s$ increases to $0.1106$, which is higher than that of standard GSPan.

        Qualitatively, Fig.~\ref{fig:wv3_4k_comparison} further reveals the visual difference between standard GSPan and $\mathrm{GSPan}^*$. The fused image produced by standard GSPan preserves sharper edges and richer fine-scale textures across the large scene, especially in regions with dense structural patterns. In comparison, $\mathrm{GSPan}^*$ maintains a relatively natural spectral appearance and overall scene consistency, but its local structures appear smoother and less detailed. This visual difference is consistent with the quantitative results in Table~\ref{tab:wv3_4k_fr}: standard GSPan achieves lower spatial distortion, whereas $\mathrm{GSPan}^*$ yields lower spectral distortion and much higher inference efficiency. The corresponding HQNR maps also show that standard GSPan contains broader high-quality regions, while the map of $\mathrm{GSPan}^*$ still remains more favorable than those of all competing baseline methods.
		
		 This result is consistent with the design goal of SDAI: it trades part of the fine spatial detail restoration for a substantial gain in inference efficiency. Therefore, standard GSPan is more suitable for quality-oriented large-scene reconstruction, whereas $\mathrm{GSPan}^{*}$ provides a practical efficiency-oriented mode for rapid large-scene fusion, previewing, or resource-constrained applications.

		\begin{figure*}[htbp]
			\centering
	
			\captionsetup[subfloat]{labelformat=empty, skip=2pt}
			\setlength{\tabcolsep}{4pt} 
			
			\begin{tabular}{cccc}
		
				\includegraphics[width=0.22\textwidth]{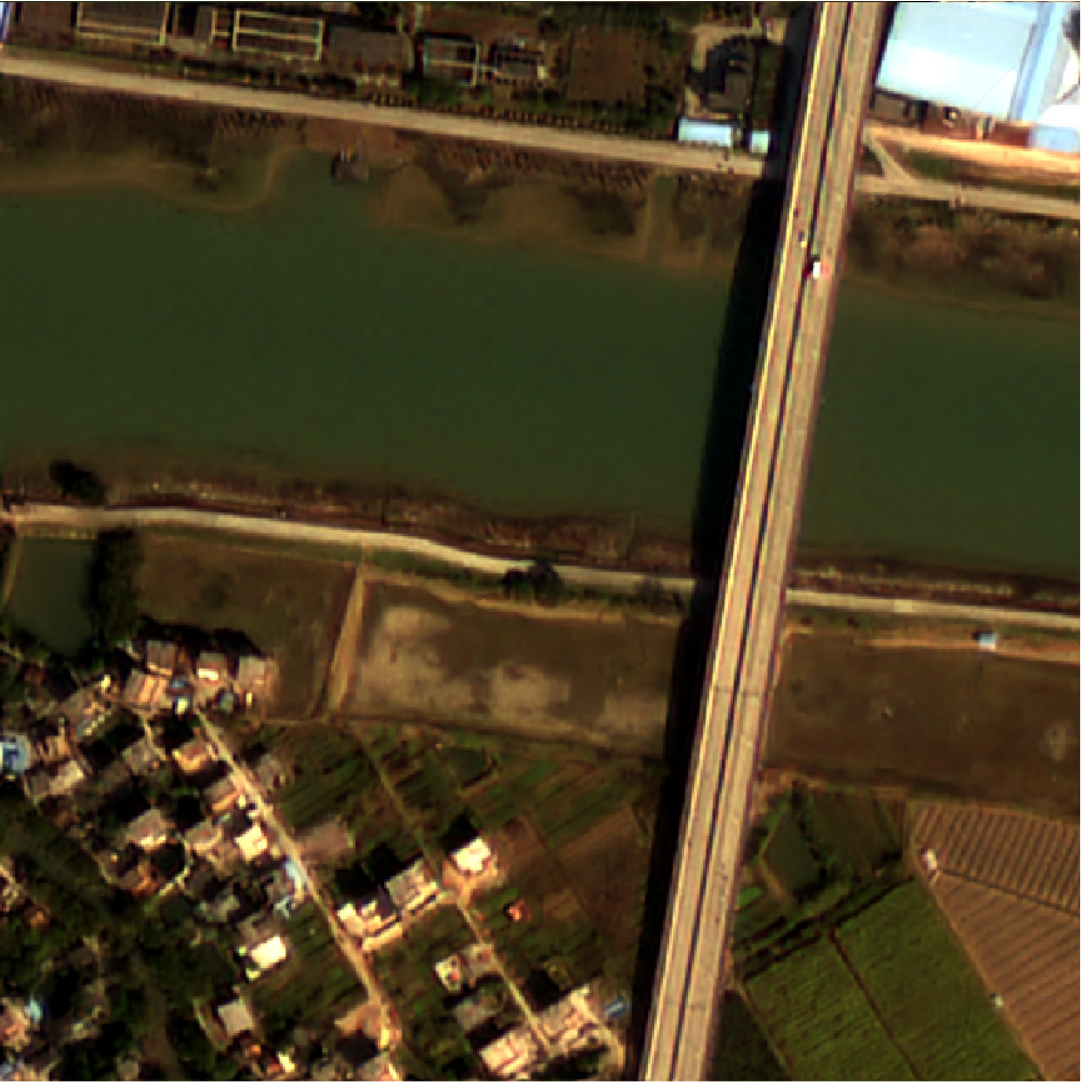} &
				\includegraphics[width=0.22\textwidth]{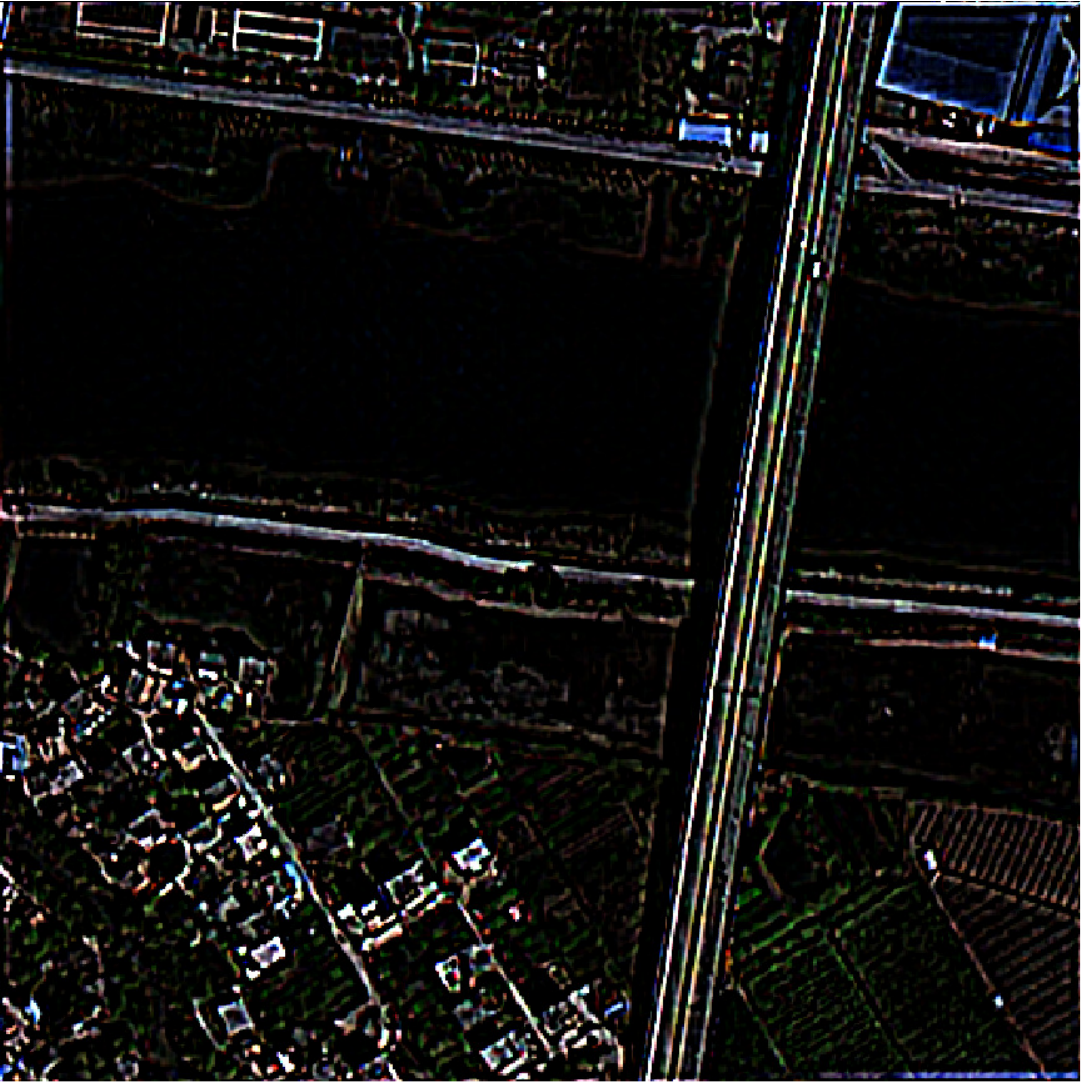} &
				\includegraphics[width=0.22\textwidth]{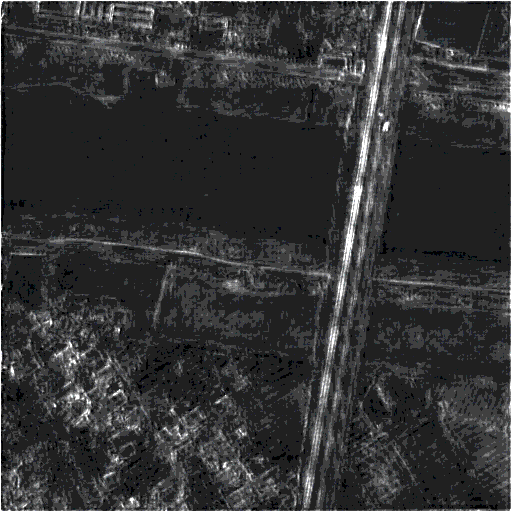} &
				\includegraphics[width=0.22\textwidth]{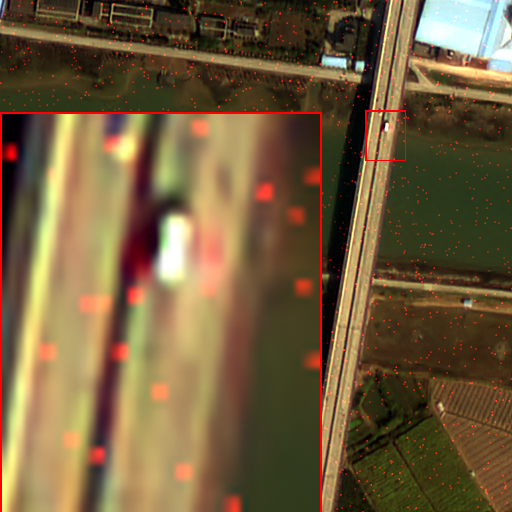}
				\\

				\small (a) Fused Image (FR) & \small (b) Residual Map & \small (c) Gaussian density Map & \small (d) Gaussian Highlight Map \\
			\end{tabular}
			
			\caption{Visual analysis of the learned Gaussian primitives on the GaoFen-2 dataset. (a) Full-resolution fused image; (b) residual map representing the residual detail field; (c) Gaussian density map illustrating the spatial distribution of primitives, where each Gaussian is rendered as a uniform circular spot at its predicted position $\mu$ to reveal content-adaptive clustering; (d) Gaussian highlight map where red dots ($0.2\%$ sampled primitives) demonstrate the anisotropic structural alignment of individual primitives within areas of high spatial complexity.
			}
			\label{fig:gs_viewer_analysis}
		\end{figure*}

		\subsection{Visual Analysis of Gaussian Primitive Distribution}

        To examine how GSPan organizes the learned Gaussian primitives for residual detail reconstruction, we visualize their spatial distribution and geometric characteristics in Fig.~\ref{fig:gs_viewer_analysis}.

        To analyze the spatial arrangement of the primitives independently of their learned anisotropic shapes and appearance-related attributes, we fix the scale, covariance-related parameter, residual coefficient, and spectral coefficient vector to constant values, and render each Gaussian as an isotropic white spot at its predicted center location $\mu$. The resulting Gaussian density map is shown in Fig.~\ref{fig:gs_viewer_analysis}(c). Brighter regions indicate higher primitive density. Compared with the residual map in Fig.~\ref{fig:gs_viewer_analysis}(b), the primitive distribution is clearly non-uniform and content-adaptive. The Gaussians tend to concentrate in regions with pronounced residual responses and dense structural details, such as bridge contours, building boundaries, and field furrows. In contrast, smooth regions show a relatively sparse primitive distribution. This observation suggests that the model learns to allocate more representational capacity to spatially complex areas.

        To further inspect the geometric flexibility of individual primitives, we randomly select $0.2\%$ of the Gaussians and render them in red for visualization, as shown in Fig.~\ref{fig:gs_viewer_analysis}(d). The zoomed-in view shows that the learned primitives are not simply isotropic points. Instead, their anisotropic covariance parameters allow them to adjust spatial extent and orientation according to local image structures. For example, the highlighted primitives along the bridge deck are elongated and aligned with nearby directional edges. This behavior indicates that GSPan can model local anisotropic textures through explicit Gaussian primitives, providing an interpretable representation of residual details field.

		\subsection{Ablation Study}
		
		\subsubsection{Effectiveness of Architectural Designs} 
		
		To validate our architectural choices, we evaluate three configurations as summarized in Table~\ref{tab:abl1}. Comparing the single-stream baseline (Config \uppercase\expandafter{\romannumeral 1}) with our full model reveals that early concatenation of PAN and MS images fails to preserve source-specific features, leading to suboptimal spectral and spatial fidelity. While the dual-stream structure without interaction (Config \uppercase\expandafter{\romannumeral 2}) improves spectral consistency, it results in the highest spatial distortion ($D_s = 0.0448$), as the two streams evolve independently without structural guidance. The proposed model, equipped with the Spatial-Spectral Interactive Attention (SSIA) module, achieves the best performance across all metrics. This validates that SSIA serves as the effective bridge that enables the model to effectively fit spatial details by coupling panchromatic structural guidance with multi-spectral content, confirming the necessity of our hierarchical interaction design.
		\begin{table}[htbp]
			\centering
			\setlength{\tabcolsep}{4pt}
			\caption{Ablation results of architectural designs on the WV3-4K dataset. Best results are highlighted in red.}
			\begin{tabular}{cccccc}
				\toprule
				config & Dual-Steam & SSIA  & $D_\lambda$ & $D_s$  & HQNR \\
				\midrule
				\uppercase\expandafter{\romannumeral 1}    & $\times$      & $\times$      & 0.0473      &  0.0273     & 0.9267  \\
				\uppercase\expandafter{\romannumeral 2}     & $\checkmark$     & $\times$      & 0.0369  & 0.0448  & 0.9200  \\
				GSPan & $\checkmark$     & $\checkmark$     & \textcolor[rgb]{ 1,  0,  0}{0.0450} & \textcolor[rgb]{ 1,  0,  0}{0.0226} & \textcolor[rgb]{ 1,  0,  0}{0.9334} \\
				\bottomrule
			\end{tabular}%
			\label{tab:abl1}%
		\end{table}%
		
		\begin{table}[htbp]
			\centering
			
			\caption{Performance comparison across different Gaussian primitive densities ($m$). Best results are highlighted in red.}
			\begin{tabular}{ccccc}
				\toprule
				$m$     & $D_\lambda$ & $D_s$  & HQNR  & Time(s) \\
				\midrule
				1     & 0.0855 & 0.0354 & 0.8822 & \textcolor[rgb]{ 1,  0,  0}{74.95} \\
				4     & \textcolor[rgb]{ 1,  0,  0}{0.0450 } & \textcolor[rgb]{ 1,  0,  0}{0.0226 } & \textcolor[rgb]{ 1,  0,  0}{0.9334 } & 81.47 \\
				9     & 0.0613 & 0.0372 & 0.9038 & 93.75 \\
				\bottomrule
			\end{tabular}%
			\label{tab:abl2}%
		\end{table}%
		\subsubsection{Impact of Gaussian Primitive Density}
		The primitive density factor $m$, where the total number of Gaussians $N = m \times H \times W$, represents a fundamental trade-off between representation capacity and efficiency. As reported in Table~\ref{tab:abl2}, a low density ($m=1$) is insufficient to capture complex 4K textures, while the best performance in our evaluation is achieved at $m=4$. Interestingly, further increasing the density to $m=9$ does not yield performance gains but instead leads to slight degradation ($D_\lambda$ rises to 0.0613) and increased inference latency. This phenomenon suggests that an excessive number of primitives  increase optimization difficulty. Consequently, we set $m=4$ as the default configuration.
		\subsection{Inference Efficiency Analysis} 
		
		To assess the practical feasibility of GSPan, we analyze its computational efficiency in terms of parameter count and inference latency on the WV3-4K dataset. Results are summarized in Table~\ref{tab:wv3_4k_fr}. 
		
		The efficiency benefit of SDAI is reflected in its inference latency. Standard full-resolution inference (GSPan) requires 76.48 seconds to process a 4K image. However, by performing primitive attributes estimation in a 1K reduced-resolution space and rendering the output directly in a 4K continuous coordinate space, $\mathrm{GSPan}^*$ reduces the latency to just 6.89 seconds—achieving an 11-fold acceleration using the exact same model weights. This speedup shows that our framework provides an efficiency-oriented solution with favorable spectral fidelity for large-scene satellite image processing.

        \section{Conclusion}

        This paper presented GSPan, a continuous Gaussian splatting framework for arbitrary-scale pansharpening. Instead of directly predicting HRMS pixels on a predefined discrete grid, GSPan represents the residual detail field with learnable anisotropic 2D Gaussian primitives and reconstructs the fused image by rendering these primitives as the residual details field over the upsampled MS base. The proposed Dual-Stream Hierarchical Interaction (DSHI) architecture and Spatial-Spectral Interactive Attention (SSIA) module integrate PAN-derived structural cues with MS spectral features, providing effective guidance for Gaussian primitive attribute estimation.

        Benefiting from the continuous primitive representation, GSPan supports rendering on arbitrary target sampling grids without scale-specific retraining. This property further enables the Scale-Decoupled Asymmetric Inference (SDAI) strategy, which estimates Gaussian primitive attributes at a reduced resolution and renders the fused image at the desired output resolution. Experiments on QB, GF2, WV3 and the WV3-4K large-scene dataset demonstrated that standard GSPan achieves strong and consistent fusion performance, while SDAI provides an efficiency-oriented inference mode with substantially reduced latency and favorable spectral fidelity, at the cost of partial spatial detail loss. These results indicate that explicit Gaussian residual modeling offers a flexible and practical representation for pansharpening, especially for scenarios requiring adjustable output resolution or efficient large-scene processing.

        Despite these promising results, several limitations remain. First, although GSPan supports arbitrary-scale rendering, the recoverable spatial details are still bounded by the information content of the input PAN image; rendering on a denser target grid cannot introduce fine structures that are not sufficiently captured during primitive attribute estimation. Second, the proposed SDAI strategy improves inference efficiency by estimating Gaussian primitives at a reduced resolution, but this efficiency gain comes at the cost of partial spatial fidelity compared with standard full-resolution inference. Future work will explore adaptive primitive density allocation, more efficient rasterization, and scale-aware primitive estimation to further improve fusion quality and computational efficiency.

		\bibliographystyle{elsarticle-num}
		\section*{CRediT authorship contribution statement}
		
		\textbf{Fangyi Li}: Conceptualization, Methodology, Investigation, Writing -- original draft,
		Writing -- review \& editing.
		\textbf{Xiaoyuan Yang}: Supervision, Writing -- review \& editing.
		\textbf{Yixiao Li}: Methodology, Writing -- review \& editing.
		\textbf{Zongyang Sui}: Methodology.
		\textbf{Kangqing Shen}: Investigation,  Methodology,  Supervision, Writing -- review \& editing.
		\textbf{Gemine Vivone}: Writing -- review \& editing.
		
		\section*{Declaration of competing interest}
		
		The authors declare that they have no known competing financial interests or personal relationships that could have appeared to influence the work reported in this paper.

		\section*{Data availability}
		
		The QB, GF2, and WV3 datasets used in this study are available from the PanCollection repository at \url{https://github.com/liangjiandeng/PanCollection}. The WV3-4K dataset cannot be publicly released due to redistribution restrictions.
		
		\section*{Acknowledgement}
		This work was supported by the National Natural Science Foundation of China under grant 62371017. This research was supported by the high performance computing (HPC) resources at Beihang University and the Supercomputing Platform of School of Mathematical Sciences at Beihang University.

		\bibliography{cas-refs}

\begin{thebibliography}{10}
\expandafter\ifx\csname url\endcsname\relax
  \def\url#1{\texttt{#1}}\fi
\expandafter\ifx\csname urlprefix\endcsname\relax\def\urlprefix{URL }\fi
\expandafter\ifx\csname href\endcsname\relax
  \def\href#1#2{#2} \def\path#1{#1}\fi

\bibitem{vivone2020new}
G.~Vivone, M.~Dalla~Mura, A.~Garzelli, R.~Restaino, G.~Scarpa, M.~O. Ulfarsson,
  L.~Alparone, J.~Chanussot, A new benchmark based on recent advances in
  multispectral pansharpening: Revisiting pansharpening with classical and
  emerging pansharpening methods, IEEE Geoscience and Remote Sensing Magazine
  9~(1) (2020) 53--81.

\bibitem{li2022deep}
J.~Li, D.~Hong, L.~Gao, J.~Yao, K.~Zheng, B.~Zhang, J.~Chanussot, Deep learning
  in multimodal remote sensing data fusion: A comprehensive review,
  International Journal of Applied Earth Observation and Geoinformation 112
  (2022) 102926.

\bibitem{ZHANG2021323}
H.~Zhang, H.~Xu, X.~Tian, J.~Jiang, J.~Ma, Image fusion meets deep learning: A
  survey and perspective, Information Fusion 76 (2021) 323--336.

\bibitem{Vivone2014Comparison}
G.~Vivone, L.~Alparone, J.~Chanussot, M.~Dalla~Mura, A.~Garzelli, G.~A.
  Licciardi, R.~Restaino, L.~Wald, A critical comparison among pansharpening
  algorithms, IEEE Transactions on Geoscience and Remote Sensing 53~(5) (2015)
  2565--2586.

\bibitem{masi2016pansharpening}
G.~Masi, D.~Cozzolino, L.~Verdoliva, G.~Scarpa, Pansharpening by convolutional
  neural networks, Remote Sensing 8~(7) (2016) 594.

\bibitem{Yang2017PanNet}
J.~Yang, X.~Fu, Y.~Hu, Y.~Huang, X.~Ding, J.~Paisley, Pannet: A deep network
  architecture for pan-sharpening, in: Proceedings of the IEEE international
  conference on computer vision, 2017, pp. 5449--5457.

\bibitem{deng2020detail}
L.-J. Deng, G.~Vivone, C.~Jin, J.~Chanussot, Detail injection-based deep
  convolutional neural networks for pansharpening, IEEE Transactions on
  Geoscience and Remote Sensing 59~(8) (2020) 6995--7010.

\bibitem{liu2020psgan}
Q.~Liu, H.~Zhou, Q.~Xu, X.~Liu, Y.~Wang, Psgan: A generative adversarial
  network for remote sensing image pan-sharpening, IEEE Transactions on
  Geoscience and Remote Sensing 59~(12) (2020) 10227--10242.

\bibitem{pangan_Ma2020}
J.~Ma, W.~Yu, C.~Chen, P.~Liang, X.~Guo, J.~Jiang, Pan-gan: An unsupervised
  pan-sharpening method for remote sensing image fusion, Information Fusion 62
  (2020) 110--120.

\bibitem{zhou2022panformer}
H.~Zhou, Q.~Liu, Y.~Wang, Panformer: A transformer based model for
  pan-sharpening, in: 2022 IEEE international conference on multimedia and expo
  (ICME), IEEE, 2022, pp. 1--6.

\bibitem{lu2025msan}
H.~Lu, Y.~Yang, S.~Huang, R.~Liu, H.~Guo, Msan: Multiscale self-attention
  network for pansharpening, Pattern Recognition 162 (2025) 111441.

\bibitem{meng2023pandiff}
Q.~Meng, W.~Shi, S.~Li, L.~Zhang, Pandiff: A novel pansharpening method based
  on denoising diffusion probabilistic model, IEEE Transactions on Geoscience
  and Remote Sensing 61 (2023) 1--17.

\bibitem{he2025pan}
X.~He, K.~Cao, J.~Zhang, K.~Yan, Y.~Wang, R.~Li, C.~Xie, D.~Hong, M.~Zhou,
  Pan-mamba: Effective pan-sharpening with state space model, Information
  Fusion 115 (2025) 102779.

\bibitem{9638577}
L.~He, J.~Zhu, J.~Li, A.~Plaza, J.~Chanussot, Z.~Yu, Cnn-based hyperspectral
  pansharpening with arbitrary resolution, IEEE Transactions on Geoscience and
  Remote Sensing 60 (2022) 1--21.

\bibitem{10974400}
L.~He, Z.~Fang, J.~Li, H.~Ye, A.~Plaza, Arbitrary-resolution hyperspectral
  pansharpening neural operators, IEEE Transactions on Geoscience and Remote
  Sensing 63 (2025) 1--19.

\bibitem{Ting2025Hyperspectral}
T.~Wang, Z.~Yan, J.~Li, X.~Zhao, C.~Wang, M.~Ng, Hyperspectral and
  multispectral image fusion with arbitrary resolution through self-supervised
  representations, International Journal of Computer Vision 133~(11) (2025)
  7515--7535.

\bibitem{chen2021learning}
Y.~Chen, S.~Liu, X.~Wang, Learning continuous image representation with local
  implicit image function, in: Proceedings of the IEEE/CVF conference on
  computer vision and pattern recognition, 2021, pp. 8628--8638.

\bibitem{cao2023ciaosr}
J.~Cao, Q.~Wang, Y.~Xian, Y.~Li, B.~Ni, Z.~Pi, K.~Zhang, Y.~Zhang, R.~Timofte,
  L.~Van~Gool, Ciaosr: Continuous implicit attention-in-attention network for
  arbitrary-scale image super-resolution, in: Proceedings of the IEEE/CVF
  Conference on Computer Vision and Pattern Recognition, 2023, pp. 1796--1807.

\bibitem{lee2022local}
J.~Lee, K.~H. Jin, Local texture estimator for implicit representation
  function, in: Proceedings of the IEEE/CVF conference on computer vision and
  pattern recognition, 2022, pp. 1929--1938.

\bibitem{liang2024fourier}
Y.-J. Liang, Z.~Cao, S.~Deng, H.-X. Dou, L.-J. Deng, Fourier-enhanced implicit
  neural fusion network for multispectral and hyperspectral image fusion,
  Advances in Neural Information Processing Systems 37 (2024) 63441--63465.

\bibitem{kerbl20233d}
B.~Kerbl, G.~Kopanas, T.~Leimk{\"u}hler, G.~Drettakis, 3d gaussian splatting
  for real-time radiance field rendering, ACM Transactions on Graphics 42~(4)
  (2023).

\bibitem{chen2024survey}
G.~Chen, W.~Wang, A survey on 3d gaussian splatting, ACM Computing Surveys
  58~(12) (2026).

\bibitem{chen2025generalized}
D.~Chen, L.~Chen, Z.~Zhang, L.~Zhang, Generalized and efficient 2d gaussian
  splatting for arbitrary-scale super-resolution, in: Proceedings of the
  IEEE/CVF International Conference on Computer Vision, 2025, pp. 26435--26445.

\bibitem{scarpa2018target}
G.~Scarpa, S.~Vitale, D.~Cozzolino, Target-adaptive cnn-based pansharpening,
  IEEE Transactions on Geoscience and Remote Sensing 56~(9) (2018) 5443--5457.

\bibitem{tfnet_Liu2020}
X.~Liu, Q.~Liu, Y.~Wang, Remote sensing image fusion based on two-stream fusion
  network, Information Fusion 55 (2020) 1--15.

\bibitem{JIN2022158}
C.~Jin, L.-J. Deng, T.-Z. Huang, G.~Vivone, Laplacian pyramid networks: A new
  approach for multispectral pansharpening, Information Fusion 78 (2022)
  158--170.

\bibitem{pereira2026multi}
I.~Pereira-S{\'a}nchez, E.~Sans, J.~Navarro, J.~Duran, Multi-head attention
  residual unfolded network for model-based pansharpening, International
  Journal of Computer Vision 134~(2) (2026) 55.

\bibitem{CHEN2025103002}
Y.~Chen, Z.~Wan, Z.~Chen, M.~Wei, Cslp: A novel pansharpening method based on
  compressed sensing and l-pnn, Information Fusion 118 (2025) 103002.

\bibitem{11478251}
Y.~Yan, Y.~Wang, W.~Tu, J.~Wang, B.~Cai, Q.~Zhuang, X.~Zuo, Y.~Chen, H.~Zhang,
  Z.~Shao, S$^{3}$mamba: Pan-sharpening via spatial–spectral synergistic
  state space model, IEEE Journal of Selected Topics in Applied Earth
  Observations and Remote Sensing 19 (2026) 12820--12834.

\bibitem{RUI2024102325}
X.~Rui, X.~Cao, L.~Pang, Z.~Zhu, Z.~Yue, D.~Meng, Unsupervised hyperspectral
  pansharpening via low-rank diffusion model, Information Fusion 107 (2024)
  102325.

\bibitem{CAO2024102001}
Q.~Cao, L.-J. Deng, W.~Wang, J.~Hou, G.~Vivone, Zero-shot semi-supervised
  learning for pansharpening, Information Fusion 101 (2024) 102001.

\bibitem{WANG2024102003}
H.~Wang, H.~Zhang, X.~Tian, J.~Ma, Zero-sharpen: A universal pansharpening
  method across satellites for reducing scale-variance gap via zero-shot
  variation, Information Fusion 101 (2024) 102003.

\bibitem{SHEN202345}
K.~Shen, X.~Yang, S.~Lolli, G.~Vivone, A continual learning-guided training
  framework for pansharpening, ISPRS Journal of Photogrammetry and Remote
  Sensing 196 (2023) 45--57.

\bibitem{shen2022docsnet}
K.~Shen, X.~Yang, Z.~Li, J.~Jiang, F.~Jiang, H.~Ren, Y.~Li, Docsnet: a
  dual-output and cross-scale strategy for pan-sharpening, International
  Journal of Remote Sensing 43~(5) (2022) 1609--1629.

\bibitem{yang2026g}
Z.~Yang, S.~Yin, J.~Liang, L.-J. Deng, G-zap: A generalizable zero-shot
  framework for arbitrary-scale pansharpening, arXiv preprint arXiv:2603.14412
  (2026).

\bibitem{zhang2024gaussianimage}
X.~Zhang, X.~Ge, T.~Xu, D.~He, Y.~Wang, H.~Qin, G.~Lu, J.~Geng, J.~Zhang,
  Gaussianimage: 1000 fps image representation and compression by 2d gaussian
  splatting, in: European Conference on Computer Vision, Springer, 2024, pp.
  327--345.

\bibitem{peng2025pixel}
L.~Peng, A.~Wu, W.~Li, P.~Xia, X.~Dai, X.~Zhang, X.~Di, H.~Sun, R.~Pei,
  Y.~Wang, et~al., Pixel to gaussian: Ultra-fast continuous super-resolution
  with 2d gaussian modeling, arXiv preprint arXiv:2503.06617 (2025).

\bibitem{hu2025gaussiansr}
J.~Hu, B.~Xia, B.~Chen, W.~Yang, L.~Zhang, Gaussiansr: High fidelity 2d
  gaussian splatting for arbitrary-scale image super-resolution, in:
  Proceedings of the AAAI Conference on Artificial Intelligence, Vol.~39, 2025,
  pp. 3554--3562.

\bibitem{liu2021swin}
Z.~Liu, Y.~Lin, Y.~Cao, H.~Hu, Y.~Wei, Z.~Zhang, S.~Lin, B.~Guo, Swin
  transformer: Hierarchical vision transformer using shifted windows, in:
  Proceedings of the IEEE/CVF international conference on computer vision,
  2021, pp. 10012--10022.

\bibitem{deng2022vivone}
{L.-J. Deng}, {G. Vivone}, {M.E. Paoletti}, {G. Scarpa}, {J. He}, {Y. Zhang},
  {J. Chanussot}, {A. Plaza}, Machine learning in pansharpening: A benchmark,
  from shallow to deep networks, IEEE Geoscience and Remote Sensing Magazine
  10~(3) (2022) 279--315.

\bibitem{wald1997fusion}
L.~Wald, T.~Ranchin, M.~Mangolini, Fusion of satellite images of different
  spatial resolutions: Assessing the quality of resulting images,
  Photogrammetric engineering and remote sensing 63~(6) (1997) 691--699.

\bibitem{alparone2004global}
L.~Alparone, S.~Baronti, A.~Garzelli, F.~Nencini, A global quality measurement
  of pan-sharpened multispectral imagery, IEEE Geoscience and Remote Sensing
  Letters 1~(4) (2004) 313--317.

\bibitem{alparone2007comparison}
L.~Alparone, L.~Wald, J.~Chanussot, C.~Thomas, P.~Gamba, L.~M. Bruce,
  Comparison of pansharpening algorithms: Outcome of the 2006 grs-s data-fusion
  contest, IEEE Transactions on Geoscience and Remote Sensing 45~(10) (2007)
  3012--3021.

\bibitem{garzelli2009hypercomplex}
A.~Garzelli, F.~Nencini, Hypercomplex quality assessment of multi/hyperspectral
  images, IEEE Geoscience and Remote Sensing Letters 6~(4) (2009) 662--665.

\bibitem{9779258}
A.~Arienzo, G.~Vivone, A.~Garzelli, L.~Alparone, J.~Chanussot, Full-resolution
  quality assessment of pansharpening: Theoretical and hands-on approaches,
  IEEE Geoscience and Remote Sensing Magazine 10~(3) (2022) 168--201.

\bibitem{garzelli2007optimal}
A.~Garzelli, F.~Nencini, L.~Capobianco, Optimal mmse pan sharpening of very
  high resolution multispectral images, IEEE Transactions on Geoscience and
  Remote Sensing 46~(1) (2007) 228--236.

\bibitem{otazu2005introduction}
X.~Otazu, M.~Gonz{\'a}lez-Aud{\'\i}cana, O.~Fors, J.~N{\'u}{\~n}ez,
  Introduction of sensor spectral response into image fusion methods.
  application to wavelet-based methods, IEEE Transactions on Geoscience and
  Remote Sensing 43~(10) (2005) 2376--2385.

\bibitem{palsson2013new}
F.~Palsson, J.~R. Sveinsson, M.~O. Ulfarsson, A new pansharpening algorithm
  based on total variation, IEEE Geoscience and Remote Sensing Letters 11~(1)
  (2013) 318--322.

\bibitem{jin2022lagconv}
Z.-R. Jin, T.-J. Zhang, T.-X. Jiang, G.~Vivone, L.-J. Deng, Lagconv:
  Local-context adaptive convolution kernels with global harmonic bias for
  pansharpening, in: Proceedings of the AAAI conference on artificial
  intelligence, Vol.~36, 2022, pp. 1113--1121.

\end{thebibliography}

	\end{sloppypar}
\end{document}